\documentclass[10pt,twocolumn,letterpaper]{article}
\usepackage[group-separator={,}]{siunitx}
\usepackage[pagenumbers]{cvpr} %

\usepackage{multirow}
\usepackage[normalem]{ulem}
\usepackage{booktabs}

\definecolor{cvprblue}{rgb}{0.21,0.49,0.74}
\usepackage[pagebackref,breaklinks,colorlinks,allcolors=cvprblue]{hyperref}
\usepackage[symbol]{footmisc}
\usepackage{colortbl}
\usepackage{tikz}
\usetikzlibrary{calc}
\def\paperID{3706} %
\def\confName{CVPR}
\def\confYear{2025}

\title{Segment Anything, Even Occluded}

\author{
Wei-En Tai$^{1}$$^\ast$ \quad
Yu-Lin Shih$^{1}$$^\ast$ \quad
Cheng Sun$^{2}$ \quad
Yu-Chiang Frank Wang$^{2,3}$ \quad
Hwann-Tzong Chen$^{1,4}$ \\
{\small $^{1}$National Tsing Hua University} \quad
{\small $^{2}$NVIDIA} \quad
{\small $^{3}$National Taiwan University} \quad
{\small $^{4}$Aeolus Robotics}
}
\begin{document}
\maketitle
\footnotetext[1]{These authors contributed equally to this work.}
\begin{abstract}
Amodal instance segmentation, which aims to detect and segment both visible and invisible parts of objects in images, plays a crucial role in various applications including autonomous driving, robotic manipulation, and scene understanding. While existing methods require training both front-end detectors and mask decoders jointly, this approach lacks flexibility and fails to leverage the strengths of pre-existing modal detectors. To address this limitation, we propose SAMEO, a novel framework that adapts the Segment Anything Model (SAM) as a versatile mask decoder capable of interfacing with various front-end detectors to enable mask prediction even for partially occluded objects.
Acknowledging the constraints of limited amodal segmentation datasets, we introduce Amodal-LVIS, a large-scale synthetic dataset comprising 300K images derived from the modal LVIS and LVVIS datasets. This dataset significantly expands the training data available for amodal segmentation research. Our experimental results demonstrate that our approach, when trained on the newly extended dataset, including Amodal-LVIS, achieves remarkable zero-shot performance on both COCOA-cls and D2SA benchmarks, highlighting its potential for generalization to unseen scenarios.
\end{abstract}
    
\begin{figure}
    \centering
    \includegraphics[width=0.115\textwidth]{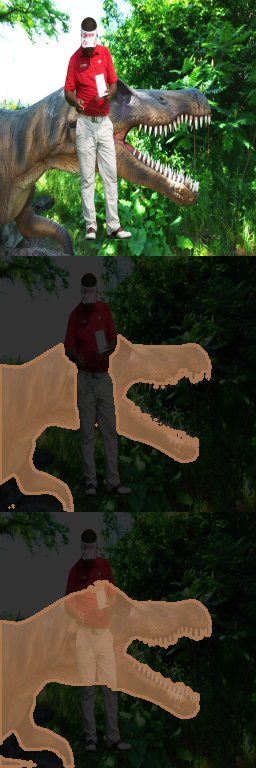}
    \includegraphics[width=0.115\textwidth]{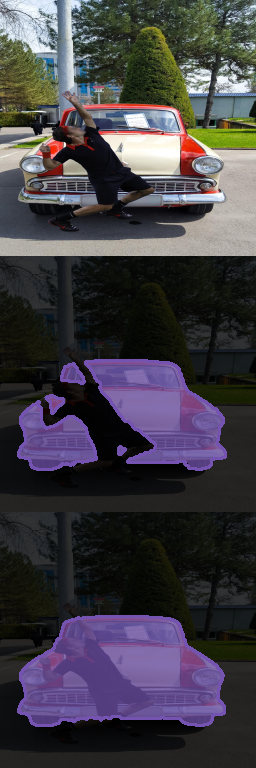}
    \includegraphics[width=0.115\textwidth]{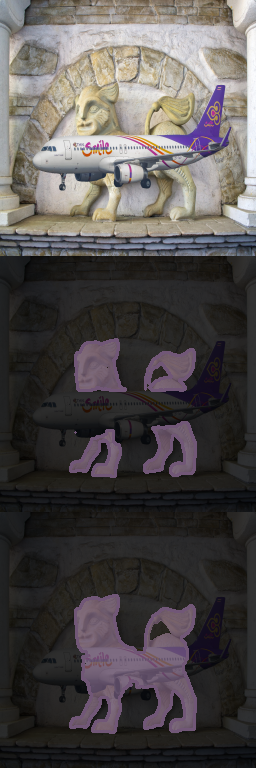}
    \includegraphics[width=0.115\textwidth]{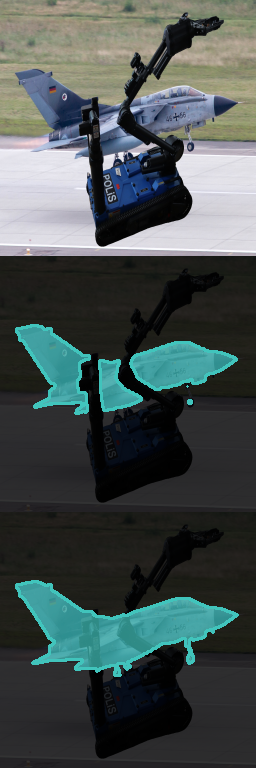}
    \caption{Amodal segmentation examples: The top row shows the original images. The middle row displays EfficientSAM predicted modal masks that only cover the visible parts of objects. The bottom row illustrates amodal masks that reveal the complete object shapes predicted by our method, SAMEO---a Segment Anything Model Even under Occlusion.}
    \label{fig:amodal_instance_segmentation}
\end{figure}

\section{Introduction}
Human visual perception extends beyond what is directly visible in a scene. We can naturally imagine and understand the complete shape of partially occluded objects through a combination of object recognition and prior knowledge about object categories. Even when correctly classifying objects is difficult, we can often infer the complete shape of partially visible objects by analyzing visible parts and reasoning about common occlusion patterns~\cite{
KanizsaLB79,Palmer99,ShipleyK01}. Amodal instance segmentation seeks to replicate this remarkable human capability by detecting and localizing objects in images and predicting their complete shapes, including both visible and occluded portions  (\Cref{fig:amodal_instance_segmentation}).

An effective approach to addressing amodal instance segmentation is to divide the task into two main components: object detection and mask segmentation. In recent years, significant advances have been made in object detection, with state-of-the-art models such as RTMDet~\cite{abs-2212-07784} and ConvNeXt-V2~\cite{WooDHC0KX23} achieving impressive performance. However, current amodal segmentation approaches often require training both the detector and mask decoder jointly, which prevents them from fully utilizing these powerful pre-trained modal detectors. This limitation motivated us to develop a more flexible framework that can leverage existing modal detectors while still maintaining strong amodal segmentation capabilities.
\ULforem
The emergence of foundation models for visual understanding has opened up new possibilities in segmentation tasks. Among these, the Segment Anything Model (SAM)~\cite{KirillovMRMRGXW23} and its efficient variant, EfficientSAM~\cite{XiongVWXXZDWSIK24}, have demonstrated remarkable capabilities in prompt-based modal segmentation. We leverage EfficientSAM's architecture, which features a lightweight encoder for faster inference, and adapt it for amodal segmentation through specialized training. Our approach enables the model to process both \emph{amodal} and \emph{modal} prompts for generating amodal mask predictions while maintaining potential zero-shot capabilities.
\normalem
Besides the improvements in algorithms and architectures, datasets are also crucial for learning-based methods, yet current amodal segmentation datasets encounter several challenges:

\begin{itemize}
    \item \textbf{Limited Scale:} Existing datasets contain relatively few images, hindering the development of robust models.
    \item \textbf{Annotation Quality:} Several datasets relying on automatic generation methods can lead to inconsistent and sometimes incorrect instance annotations when not properly validated.
    \item \textbf{Irrelevant Objects:} A significant portion of annotated objects, such as walls and floors, contribute little to meaningful scene understanding.
\end{itemize}

To address these limitations, we present Amodal-LVIS, a new large-scale dataset derived from LVIS~\cite{GuptaDG19} and LVVIS~\cite{WangJTHYXWG23}. Our dataset contains 300K carefully curated images, where each image contains one instance annotation. These annotations form paired examples between synthetic occluded instances and their original unoccluded versions. Furthermore, we have processed and refined existing datasets to create a comprehensive training collection comprising approximately 1M images and 2M instance annotations.

Experimental results show that our method with the EfficientSAM architecture, when trained on our combined dataset, achieves remarkable zero-shot performance that surpasses previous supervised amodal segmentation methods. These results validate our approach of leveraging an efficient existing architecture with high-quality, large-scale training data for amodal segmentation tasks.

Our main contributions can be summarized as follows:
\begin{enumerate}
    \item \textbf{Flexible Amodal Framework:} The proposed method, SAMEO, adapts EfficientSAM for amodal instance segmentation that works with both modal and amodal detector prompts through specialized training.
    \item \textbf{Large-scale Dataset:} A new Amodal-LVIS dataset containing 300K images, forming paired examples between synthetic occluded instances and their original unoccluded versions.
    \item \textbf{Dataset Collection:} A comprehensive training collection of 1M images and 2M instances created by combining and refining existing amodal datasets with Amodal-LVIS.
    \item \textbf{Zero-shot Performance:} State-of-the-art zero-shot results on both COCOA-cls and D2SA benchmarks, surpassing previous supervised methods.
\end{enumerate}

\begin{figure}
    \centering
    \includegraphics[width=0.475\textwidth]{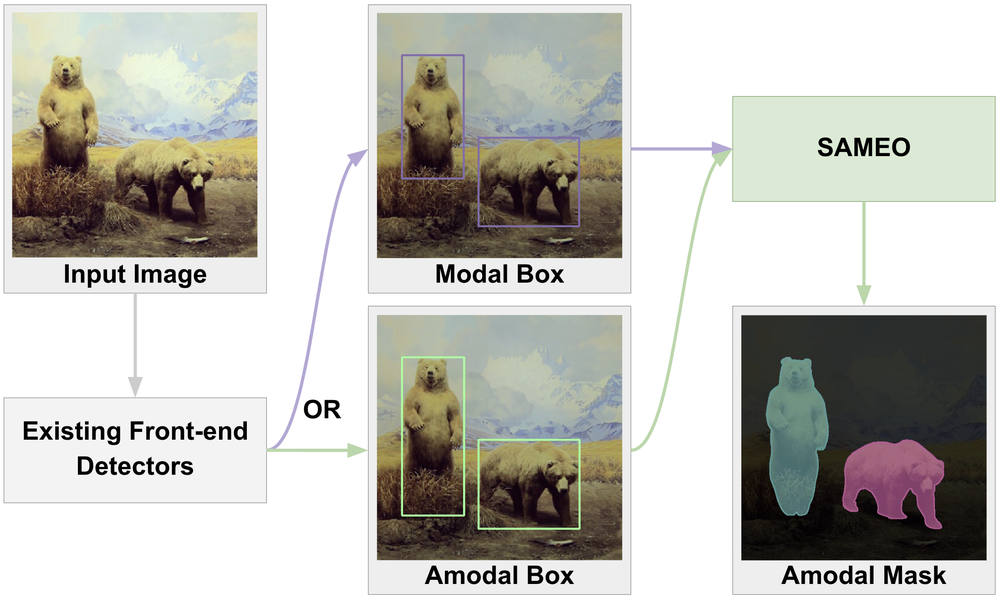}
    \caption{Overview of our amodal segmentation pipeline. Given an input image, existing object detectors first generate either modal boxes (showing visible regions) or amodal boxes (showing complete object extent). Our SAMEO then processes these detections to produce amodal masks that recover the full shape of objects, including occluded parts.}
    \label{fig:inference_pipeline}
\end{figure}

\section{Related Work}
\subsection{Instance Segmentation}
Instance segmentation is a fundamental computer vision task that simultaneously addresses object detection and segmentation, aiming to both locate objects in a scene and generate precise mask predictions for each detected instance. Initially focusing on visible parts of objects (modal instance segmentation), this field continues to evolve with the emergence of deep learning architectures. State-of-the-art methods demonstrate improvements through transformer-based feature extraction~\cite{WangT23}, modernized convolutions~\cite{WooDHC0KX23,PreteGR21}, and optimized speeds~\cite{abs-2212-07784}. Further detection models built upon the DETR architecture~\cite{CarionMSUKZ20} have achieved additional advances through specialized query selection mechanisms and training schemes~\cite{0097LL000NS23,ZongS023}.

Building upon modal instance segmentation, amodal instance segmentation extends the task to predict complete object shapes, including occluded regions. This extension is first formalized by Li and Malik~\cite{LiM16}, leading to various architectural innovations~\cite{GaoQWXHZF23,YangRXCCPB19,YaoHWXHLWFZ22}. Notable approaches include ORCNN~\cite{FollmannKHKB19}, ASN~\cite{QiJ0SJ19}, which enhance Mask R-CNN~\cite{HeGDG17} with occlusion reasoning capabilities, and BCNet~\cite{KeTT21} with its bilateral layers for handling object overlaps. Currently, AISFormer~\cite{Tran0YFKL22} represents the state-of-the-art in amodal instance segmentation by introducing transformers to effectively model long-range dependencies.

\subsection{Segment Anything Model}
The Segment Anything Model (SAM) ~\cite{KirillovMRMRGXW23} represents a significant advancement in foundational computer vision models, capable of segmenting any visual object based on various prompts, including points or boxes. Trained on a dataset of 11M images, SAM demonstrates outstanding zero-shot generalization capabilities across diverse object categories and domains. 

The original SAM model, despite its strong performance, faces practical limitations due to high computational demands, including significant memory requirements and slow inference speed. EfficientSAM~\cite{XiongVWXXZDWSIK24} addresses these challenges by using a Masked Autoencoder (MAE)~\cite{HeCXLDG22} pre-training method to learn the feature embeddings from SAM's original ViT-H encoder, resulting in faster inference speed and reduced model size while maintaining comparable segmentation performance.

\subsection{Amodal Datasets}
Several datasets have been introduced for amodal segmentation. COCOA~\cite{ZhuTMD17} is the first amodal dataset, providing semantic-level amodal annotations for COCO images. D2SA/COCOA-cls~\cite{FollmannKHKB19} extends this with instance-level annotations. DYCE~\cite{EhsaniMF18} offers synthetic indoor scenes with accurate ground truth. The KINS dataset~\cite{QiJ0SJ19} focuses on traffic scenarios with 14K images of vehicles and pedestrians. More recently, MUVA~\cite{LiYTBZJ023} introduces a multi-view shopping scenario dataset, while MP3D-Amodal~\cite{ZhanZXZ24} provides real-world indoor scenes from Matterport3D. WALT~\cite{ReddyTN22} uniquely utilizes time-lapse imagery to obtain amodal ground truth, and KITTI-360-APS~\cite{MohanV22} extends KITTI-360~\cite{LiaoXG23} with amodal panoptic annotations.
Furthermore, datasets from related amodal completion works, such as pix2gestalt~\cite{OzgurogluLS0DTV24}, have contributed to the development of the field.

\section{Our Approach}
\label{sec:method}
\subsection{Enabling Amodal Mask Prediction}
\paragraph{Preliminaries: EfficientSAM.}
\ULforem
Segment Anything Model (SAM)~\cite{KirillovMRMRGXW23} is a foundation model for image segmentation that can generate high-quality object masks based on any prompt. The original SAM architecture consists of three main components: \emph{i})~an image encoder that transforms the input image into image embeddings, \emph{ii})~a lightweight transformer-based prompt encoder that converts prompts (points, boxes) into unified embeddings, and \emph{iii})~a mask decoder that utilizes a transformer architecture with two cross-attention layers to process both image and prompt embeddings for generating the final segmentation mask.
\normalem
In our approach, we mainly use EfficientSAM~\cite{XiongVWXXZDWSIK24}, a compact adaptation of the original SAM model. EfficientSAM replaces SAM's image encoder with a lightweight ViT variant~\cite{DosovitskiyB0WZ21} while maintaining the original prompt encoder and mask decoder.

\paragraph{Model Architecture.}
We propose SAMEO for amodal instance segmentation, retaining a lightweight image encoder $\mathcal{E}$ as in the original architecture of EfficientSAM, a transformer-based prompt encoder $\mathcal{P}$, and a mask decoder $\mathcal{D}$ with dual cross-attention layers. Given an input image $I$ and a bounding box prompt $B$, the proposed SAMEO pipeline predicts the amodal mask $\hat{M}$ and the estimated IoU $\hat{\rho}$ as follows:

\begin{equation}
\hat{M}, \hat{\rho} = \mathcal{D}(\mathcal{E}(I), \mathcal{P}(B)) \,.
\end{equation}

\paragraph{Training Strategy.}

During training, we exclusively fine-tune EfficientSAM's mask decoder while keeping the original weights of the image encoder and prompt encoder unchanged. The model receives two inputs: an image and a bounding box prompt derived from ground-truth annotations. The box prompts are randomly selected from modal and amodal ground-truth boxes with equal probability. The training objective combines Dice loss~\cite{MilletariNA16}, Focal loss~\cite{LinGGHD17}, and L1 loss for IoU estimation:
\begin{equation}
\mathcal{L} = \mathcal{L}_{\text{Dice}} + \mathcal{L}_{\text{Focal}} + \lambda\mathcal{L}_{\text{IoU}} \,,
\end{equation}
where
\begin{equation}
\mathcal{L}_{\text{Dice}} = 1 - \frac{2|\hat{M} \cap M_{\text{gt}}|}{|\hat{M}| + |M_{\text{gt}}|} \,,
\end{equation}
\begin{equation}
\mathcal{L}_{\text{Focal}} = -(1-p_t)^\gamma \log(p_t) \,,
\end{equation}
\begin{equation}
\mathcal{L}_{\text{IoU}} = \left|\hat{\rho} - \text{IoU}(\hat{M}, M_{\text{gt}}) \right| \,.
\end{equation}
Here, $M_{\text{gt}}$ represents the ground truth amodal mask, $p_t$ is the predicted probability for the target class, $\gamma$ is the focusing parameter set to 2 in our experiments, and $\lambda$ is empirically set to 0.05.

\paragraph{Inference Pipeline.}
For inference, SAMEO can be flexibly integrated with various object detectors, including both amodal detectors (\eg, AISFormer~\cite{Tran0YFKL22}) and conventional modal detectors (\eg, RTMDet~\cite{abs-2212-07784}). The detection outputs serve as box prompts for our model, which then generates corresponding amodal masks (\Cref{fig:inference_pipeline}). This modular design allows our model to enhance existing detection systems with amodal segmentation capabilities while achieving state-of-the-art performance.

\begin{table*}
\centering
\begin{tabular}{l@{\hspace{2em}}c@{\hspace{2.5em}}S[table-format=6.0,group-separator={,},group-minimum-digits=4]@{\hspace{2.5em}}S[table-format=6.0,group-separator={,},group-minimum-digits=4]@{\hspace{2em}}c@{\hspace{2em}}c}
    \toprule
    \textbf{Dataset} & \textbf{Type} &  \textbf{\# Instances} & \textbf{\# Images} & \textbf{POI (\%)} & \textbf{Average ROR (\%)}\\
    \midrule
    COCOA (\textit{no stuff})~\cite{ZhuTMD17} & Real & 32926 & 5073 & 54.9 & 30.2 \\
    COCOA-cls ~\cite{FollmannKHKB19} & Real & 10562 & 3499 & 49 & 21.8 \\
    KINS~\cite{QiJ0SJ19} & Real & 188085 & 14993 & 54.3 & 42.2 \\
    KITTI-360-APS~\cite{MohanV22} & Real & 89938 & 12496 & 50 & 33.9 \\
    D2SA~\cite{FollmannKHKB19} & Synthetic & 28720 & 5600 & 53.6 & 23.7 \\
    MUVA~\cite{LiYTBZJ023} & Synthetic & 198573 & 26406 & 76.7 & 32.1 \\
    WALT$^{*}$~\cite{ReddyTN22} & Synthetic & 485369 & 40000 & 32 & 36.5 \\
    DYCE$^{*}$~\cite{EhsaniMF18} & Synthetic & 66453 & 5229 & 77.8 & 29 \\
    MP3D-amodal$^{*}$~\cite{ZhanZXZ24} & Synthetic & 2968 & 2549 & 100 & 41.1 \\
    pix2gestalt~\cite{OzgurogluLS0DTV24} & Synthetic & 849667  & 849667 & 100 & 35.9 \\
    \midrule
    Amodal-LVIS & Synthetic & 399398 & 301493 & 50 & 34.5 \\
    \midrule
    Total & -- & 2352659 & 1267424 & -- & -- \\
    \bottomrule
\end{tabular}
\caption{Amodal dataset collection. Datasets marked with * are generated or refined by ourselves. POI is calculated as the percentage of occluded instances (where modal mask is not equal to amodal mask) over all instances. Meanwhile, average ROR is the average ratio of occluded regions, considering only instances with occlusions.}
\label{tab:datasets}
\end{table*}

\begin{figure}
    \centering
    \begin{tabular}{@{}c@{\hspace{0.2cm}}c@{\hspace{0.2cm}}c@{}}
        \small(a) DYCE & \small(b) MP3D-amodal & \small(c) pix2gestalt \\
        \includegraphics[height=0.1\textheight]{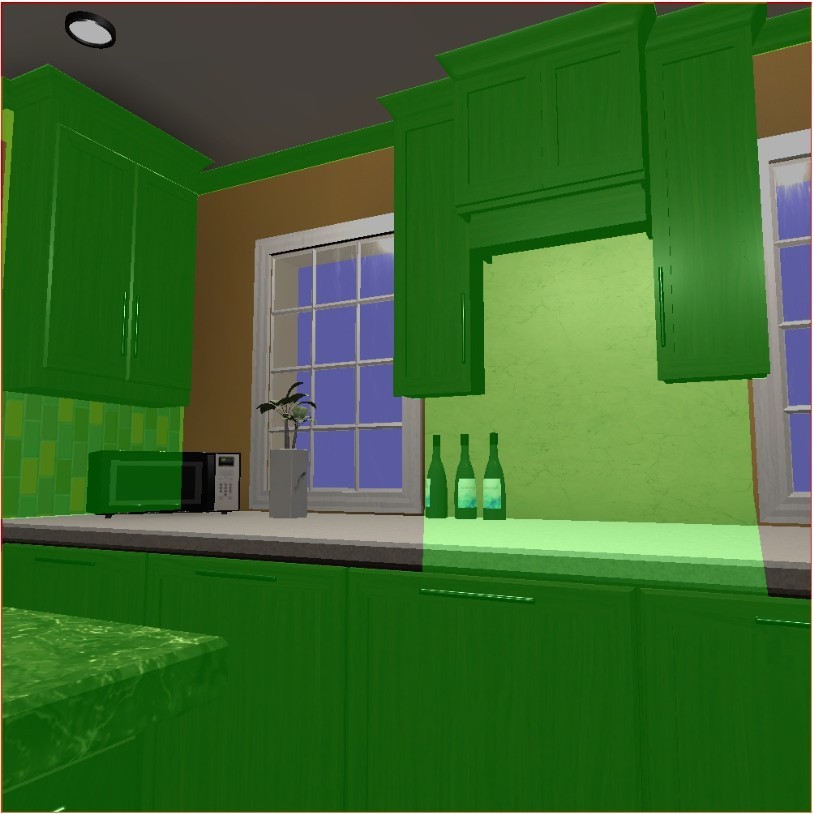} &
        \includegraphics[height=0.1\textheight]{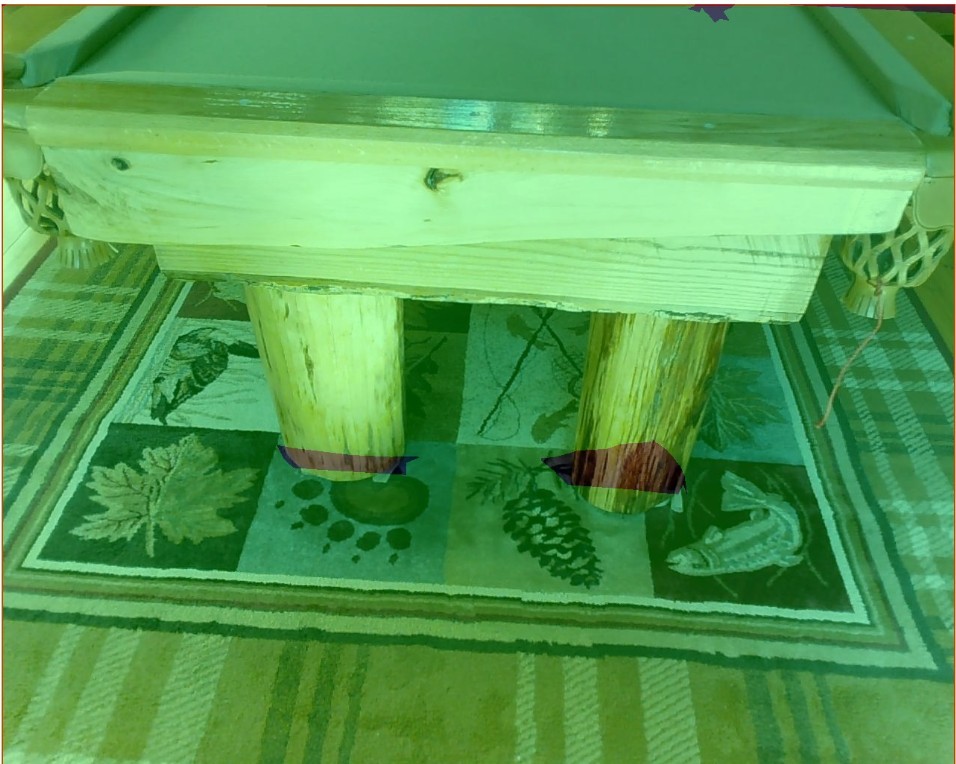} &
        \includegraphics[height=0.1\textheight]{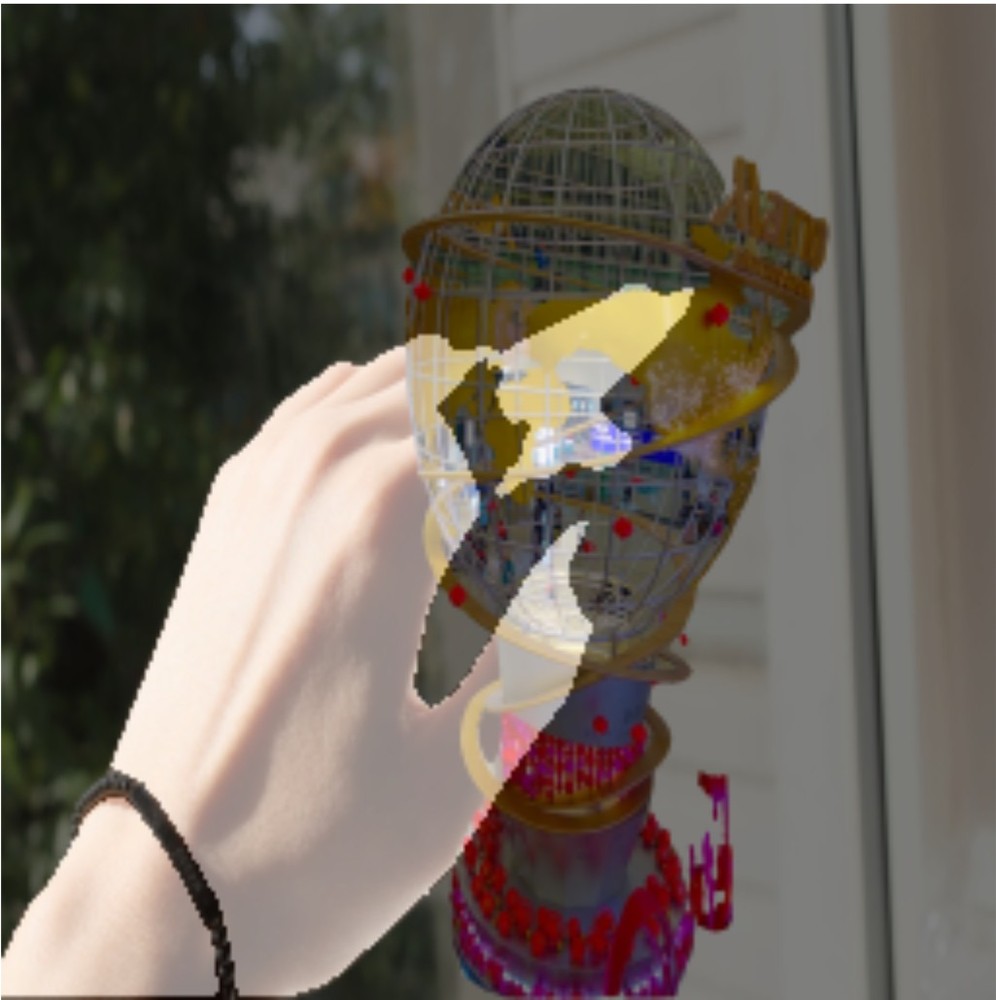}
    \end{tabular}
    \caption{Examples of limitations in existing amodal datasets: (a) DYCE and (b) MP3D-amodal show meaningless architectural elements rendered from 3D meshes that dominate the image space, while (c) pix2gestalt contains potentially incomplete amodal masks due to restrictive generation criteria.}
    \label{fig:amodal_dataset_limitation}
\end{figure}

\subsection{Amodal Dataset Collection}
\paragraph{Limitations of Existing Amodal Datasets.}
Existing amodal datasets have inherent limitations in both manual and synthetic annotation mechanisms. Human-annotated datasets, while closely representing real-world scenarios, are costly to produce and prone to errors in occluded region estimation. Synthetic datasets, though efficient to generate, lack reliable verification mechanisms for object completeness and may not accurately reflect natural occlusion patterns (\Cref{fig:amodal_dataset_limitation}).

\paragraph{Dataset Collection and Quality Control.}
To leverage the advantages of both mechanisms, we have collected and filtered datasets of both annotation types. Our cleaning process addresses specific issues in each dataset to ensure data quality while maintaining realistic occlusion representations (\Cref{tab:datasets}).

For synthetic datasets DYCE and MP3D-amodal, generated using 3D furniture meshes, we identify and address two main quality issues: meaningless architectural elements (walls, floors, ceilings) occupying the majority of image space and objects with minimal visible areas. We implement filters to remove cases where visible parts are less than 10\% of the whole object, objects occupying more than 90\% of the image area, and architectural element annotations.

The WALT dataset leverages road surveillance time-lapse footage for synthetic data generation. It obtains bounding boxes for cars and people using a pre-trained detector and then identifies complete objects by analyzing these bounding box intersections. These discovered complete objects are then composited back into the same scenes to generate synthetic training data.  However, their layer-by-layer placement can create unrealistic occlusions. We address this by implementing an occlusion threshold filter to ensure natural occlusion patterns.

For other datasets with class annotation, such as COCOA, the availability of semantic labels enabled straightforward quality control. We filter out ``stuff'' class annotations across these datasets to focus on meaningful objects that align with amodal instance segmentation goals.

\begin{figure}
    \centering
    \begin{tabular}{@{}c@{\hspace{0.1cm}}c@{\hspace{0.1cm}}c@{}}
        \small Original Image & \small Occluder & \small Synthesized Image \\
        \includegraphics[height=0.1\textheight]{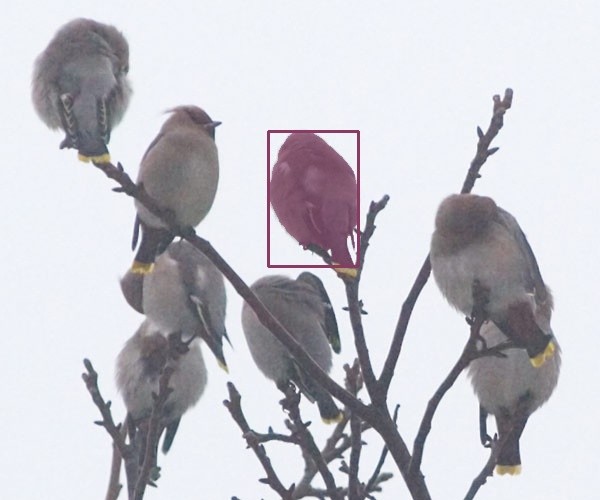} &
        \includegraphics[height=0.1\textheight]{fig/000000000143_2_occ_1.jpg} &
        \includegraphics[height=0.1\textheight]{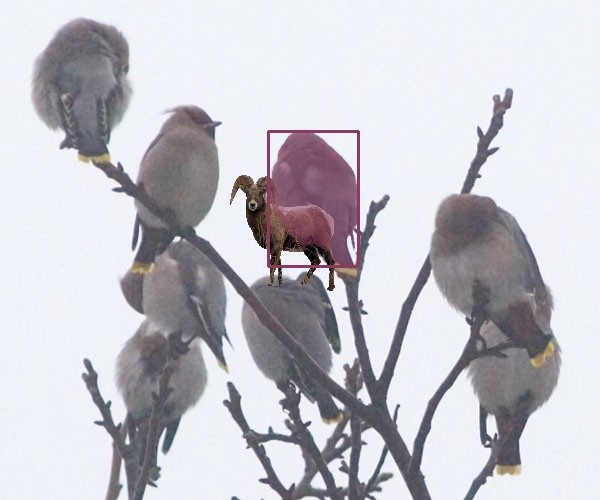}
    \end{tabular}
    \caption{Amodal-LVIS dataset generation process. From left to right: original image with unoccluded objects, a selected occluder object, and the synthesized image with occlusion. Our dataset includes both the original and the synthesized image for each instance to prevent occlusion bias during training.}
    \label{fig:amodal_lvis}
\end{figure}

\subsection{Amodal-LVIS}
We propose a synthetic dataset for amodal mask segmentation through precise object occlusion generation, incorporating complete object collection, synthetic occlusion generation, and a dual annotation mechanism to prevent model bias. Combined with existing datasets, our collection totals 1M images and 2M instance annotations.
\paragraph{Complete Object Collection.}
To obtain complete objects for synthetic occlusion, we utilize SAMEO, which is pre-trained on the previously mentioned amodal datasets, to generate pseudo labels for instances within LVIS and LVVIS datasets. Our model predicts amodal masks for each instance, which are then compared with the ground-truth visible mask annotations. This comparison helps us identify complete, unoccluded objects.
\paragraph{Synthetic Occlusion Generation.}
The occlusion generation process involves pairing randomly selected complete objects from our collected pool. To ensure that the occlusions look realistic, we normalize the paired objects to similar sizes while maintaining their aspect ratios. Object positioning and occlusion rates are controlled using bounding box annotations, which allow for precise management of how objects occlude one another.

\paragraph{Dual Annotation Mechanism.}
Our experiments in \Cref{sec:experiments} show that training solely on occluded masks leads to model confusion, resulting in an over-prediction of occluded objects even when the prompts are intended to target foreground instances. To resolve this issue, we include both occluded and original unoccluded versions of instances in our dataset (\Cref{fig:amodal_lvis}). This dual annotation mechanism prevents occlusion bias while providing comprehensive training examples for both states.

\section{Experiments}
\label{sec:experiments}
\subsection{Settings}
\paragraph{Implementation Details.}
Our model is trained on NVIDIA Tesla V100/A100 GPUs for 1,440/2,340/22,500 iterations on COCOA-cls/D2SA/MUVA respectively. For zero-shot SAMEO, we increase the batch size to 32 and train for 40,000 iterations. We use the Adam optimizer with a learning rate of $1\times10^{-4}$ without any learning rate scheduler. For each instance during training, we randomly select either the amodal or modal ground truth bounding box as the prompt with equal probability.

\paragraph{Datasets and Baselines.}
For training, we use our dataset collection and the proposed Amodal-LVIS dataset. As for evaluation, COCOA-cls, D2SA and MUVA datasets are utilized. We primarily compare our approach against AISFormer, the current state-of-the-art model in amodal instance segmentation. Unlike conventional instance segmentation models that incorporate both object detection and mask decoding components, SAMEO functions solely as a mask decoder. This allows our model to flexibly integrate with existing amodal and modal instance segmentation models, using their object box predictions as prompts to generate refined amodal masks. For a comprehensive evaluation, we compare our results against the original mask predictions from these front-end models.

To evaluate our zero-shot performance, we extend the comparisons beyond AISFormer to include both modal and amodal front-end models equipped with the original EfficientSAM as their mask decoder. This comparison demonstrates our successful adaptation of EfficientSAM for amodal mask segmentation while maintaining zero-shot capabilities.
\paragraph{Evaluation Metrics.}
We evaluate our method using two standard metrics: mean Average Precision (AP) and mean Average Recall (AR). Since our model is class-agnostic, we compute both metrics without considering class labels. For a fair comparison, we reproduce baseline methods and evaluate them using the same class-agnostic AP and AR. For methods that use SAMEO as mask decoder, we refine the confidence score $\hat{\rho}_\mathrm{front}$ of front-end models using the estimated IoU $\hat{\rho}_\mathrm{ours}$ predicted by SAMEO when calculating these metrics. The refined confidence score $\hat{\rho}_\mathrm{ref}$ of these cases is computed as follows:
\begin{equation}
\hat{\rho}_\mathrm{ref} = \hat{\rho}_\mathrm{front} \times \hat{\rho}_\mathrm{ours} \,.
\end{equation}

\begin{table*}
\centering

\begin{tabular}{l@{\hskip 0.25in}c@{\hskip 0.1in}c@{\hskip 0.1in}c@{\hskip 0.1in}c@{\hskip 0.25in}c@{\hskip 0.1in}c@{\hskip 0.1in}c@{\hskip 0.1in}c@{\hskip 0.25in}c@{\hskip 0.1in}c@{\hskip 0.1in}c@{\hskip 0.1in}c}
\toprule
\multirow{2}{*}{\textbf{Model}} & \multicolumn{4}{c@{\hskip 0.25in}}{\textbf{COCOA-cls}} & \multicolumn{4}{c@{\hskip 0.25in}}{\textbf{D2SA}} & \multicolumn{4}{c}{\textbf{MUVA}} \\
& \textbf{AP} & \textbf{AP$_{50}$} & \textbf{AP$_{75}$} & \textbf{AR} & \textbf{AP} & \textbf{AP$_{50}$} & \textbf{AP$_{75}$} & \textbf{AR} &\textbf{AP} & \textbf{AP$_{50}$} & \textbf{AP$_{75}$} & \textbf{AR} \\
\midrule[\heavyrulewidth]
AISFormer~\cite{Tran0YFKL22} & 40.6 & 70.5 & 42.5 & 55.2 & 66.3 & 89.9 & 72.8 & 76.1 & 69.4 & 90.3 & 75.6 & 80.7 \\
RTMDet$^{*}$~\cite{abs-2212-07784} & 49.8 & 71.2 & 54.7 & 69.6 & 59.7 & 81.3 & 63.4 & 78.2 & 46.0 & 68.2 & 46.4 & 57.8 \\
ConvNeXt-V2$^{*}$~\cite{WooDHC0KX23} & 46.7 & 67.5 & 52.0 & 70.9 & 66.1 & 89.1 & 69.8 & 75.3 & 47.9 & 69.0 & 46.6 & 61.7 \\
ViTDet$^{*}$~\cite{WangT23} & 47.4 & 69.4 & 52.5 & 68.8 & 63.0 & 86.4 & 65.8 & 75.4 & 45.3 & 66.9 & 44.7 & 60.7\\
\midrule
AISFormer$+$SAMEO & 54.3 & 74.0 & 59.7 & 69.3 & 79.8 & 92.7 & 84.8 & 84.2 & 76.2 & 90.8 & 80.9 & \textbf{85.2} \\
RTMDet$^{*}$$+$SAMEO & \textbf{55.3} & \textbf{75.2} & \textbf{60.8} & \textbf{74.3} & 72.7 & 85.8 & 77.5 & 84.2 & 75.8 & 89.2 & 79.9 & 83.1 \\
ConvNeXt-V2$^{*}$$+$SAMEO & 54.1 & 73.1 & 59.3 & 74.0 & \textbf{80.8} & \textbf{94.0} & \textbf{85.1} & \textbf{87.1}& \textbf{79.2} & \textbf{93.1} & \textbf{82.6} & 81.3 \\
ViTDet$^{*}$$+$SAMEO & 54.1 & 73.3 & 59.2 & 72.3 & 78.6 & 92.3 & 82.6 & 84.7 & 74.1 & 89.0 & 78.2 & 82.3\\

\bottomrule
\end{tabular}

\caption{Quantitative comparison on various datasets. Models marked with * are modal instance segmentation methods that detect objects and segment their visible masks without handling occlusions, with performance metrics calculated using their modal mask predictions. SAMEO takes bounding box predictions from the front-end models as prompts to generate amodal instance masks. Evaluation metrics include Average Precision (AP) at different IoU thresholds and Average Recall (AR). Bold numbers indicate the best performance.}
\label{tab:nonzeroshot}
\end{table*}

\begin{table*}
\centering
\begin{tabular}{l@{\hskip 0.3in}cccc@{\hskip 0.3in}cccc}
\toprule
\multirow{2}{*}{\textbf{Model}} & \multicolumn{4}{c@{\hskip 0.3in}}{\textbf{COCOA-cls}} & \multicolumn{4}{c}{\textbf{D2SA}} \\
& \textbf{AP} & \textbf{AP$_{50}$} & \textbf{AP$_{75}$} & \textbf{AR} & \textbf{AP} & \textbf{AP$_{50}$} & \textbf{AP$_{75}$} & \textbf{AR} \\
\midrule[\heavyrulewidth]
AISFormer & 40.6 & 70.5 & 42.5 & 55.2 & 66.3 & 89.9 & 72.8 & 76.1 \\
AISFormer$+$EfficientSAM$^\dagger$ & 47.6 & 70.0 & 51.7 & 64.2 & 69.6 & 89.2 & 72.3 & 77.7 \\
RTMDet$^{*}$$+$EfficientSAM$^\dagger$ & 48.7 & 71.1 & 53.2 & 65.9 & 63.0 & 82.7 & 64.8 & 74.3 \\
RetinaNet$^{*}$~\cite{PreteGR21}$+$EfficientSAM$^\dagger$ & 44.8 & 67.1 & 48.6 & 68.4 & 60.6 & 77.5 & 62.9 & 79.3 \\
\midrule
DINO$^{*}$~\cite{0097LL000NS23}$+$SAMEO$^\dagger$ & 50.2 & 70.4 & 55.7 & \textbf{73.7} & 69.8 & 86.2 & 72.6 & 75.9 \\ 
RetinaNet$^{*}$$+$SAMEO$^\dagger$ & 51.0 & 72.2 & 56.6 & 72.6 & 62.5 & 78.0 & 64.8 & 80.7 \\
AISFormer$+$SAMEO$^\dagger$ & 52.8 & 73.4 & 57.9 & 67.9 & 74.1 & 90.2 & 78.3 & 79.9 \\
ViTDet$^{*}$$+$SAMEO$^\dagger$ & 53.2 & 73.0 & 58.6 & 71.2 & 72.1 & 89.0 & 74.9 & 80.7 \\
ConvNeXt-V2$^{*}$$+$SAMEO$^\dagger$ & 53.4 & 73.0 & 59.0 & 73.0 & 74.3 & \textbf{91.6} & 78.0 & \textbf{83.1} \\ 
CO-DETR$^{*}$~\cite{ZongS023}$+$SAMEO$^\dagger$ & 54.0 & 74.8 & 60.2 & 73.5 & \textbf{75.0} & 91.0 & \textbf{78.5} & 82.2\\
RTMDet$^{*}$$+$SAMEO$^\dagger$ & \textbf{54.4} & \textbf{75.0} & \textbf{60.2} & 73.4 & 68.4 & 84.5 & 71.0 & 77.7 \\
\bottomrule
\end{tabular}
\caption{Zero-shot performance on COCOA-cls and D2SA datasets. The results show SAMEO not only significantly outperforms AISFormer but also successfully adapts EfficientSAM's modal segmentation capability to amodal segmentation, demonstrating consistent performance improvements when paired with various front-end detectors. † indicates zero-shot evaluation without training on the test dataset. * denotes modal object detectors that provide modal bounding boxes as prompts. Bold numbers indicate the best performance.}
\label{tab:ap_zeroshot}
\end{table*}

\subsection{Results}
\begin{figure*}[t]
    \centering
    \begin{minipage}{0.02\textwidth}
        \vspace{0.5cm}
        \rotatebox{90}{\textbf{Ours}} \\[0.8cm]
        \rotatebox{90}{\textbf{AISFormer}}
    \end{minipage}%
    \begin{minipage}{0.98\textwidth}
        \centering
        \resizebox{!}{4.31cm}{\includegraphics{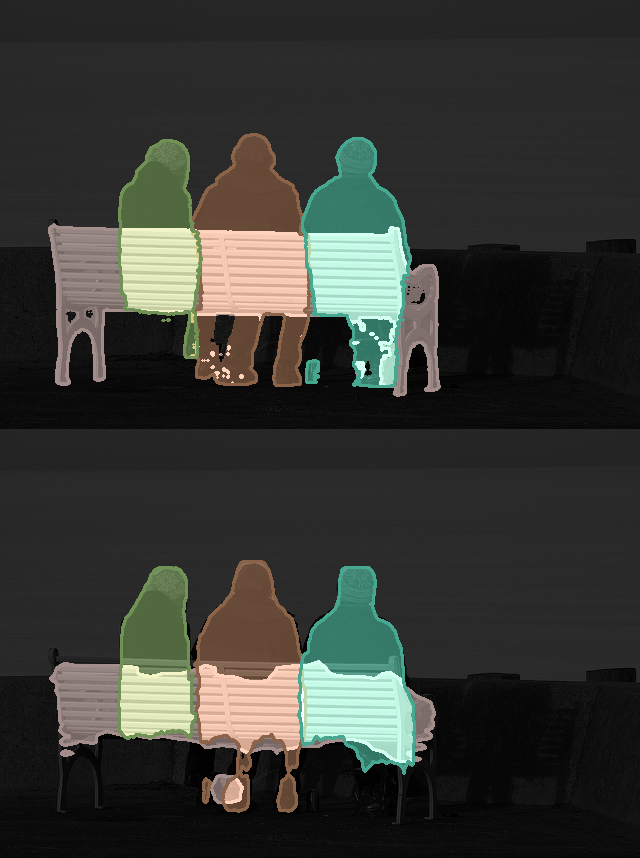}}
        \resizebox{!}{4.31cm}{\includegraphics{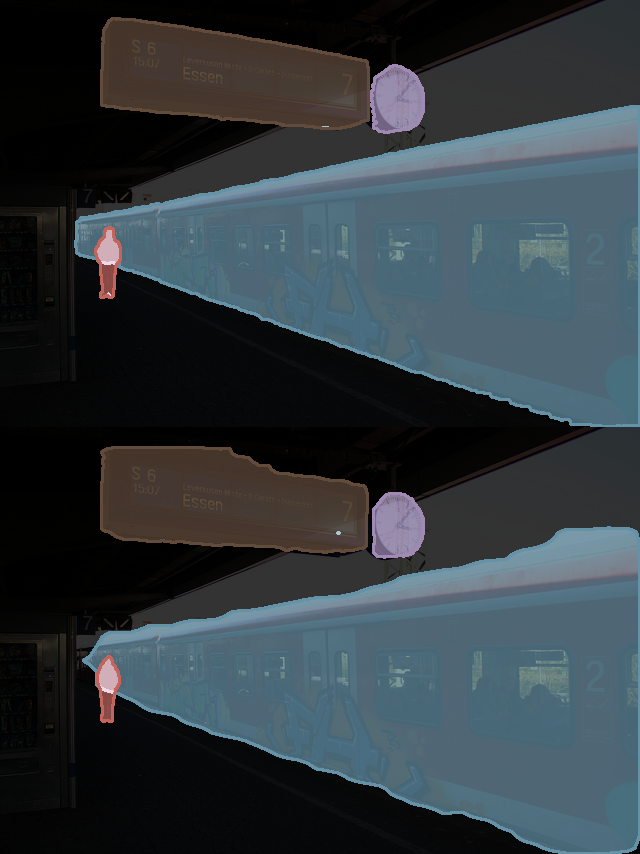}}
        \resizebox{!}{4.31cm}{\includegraphics{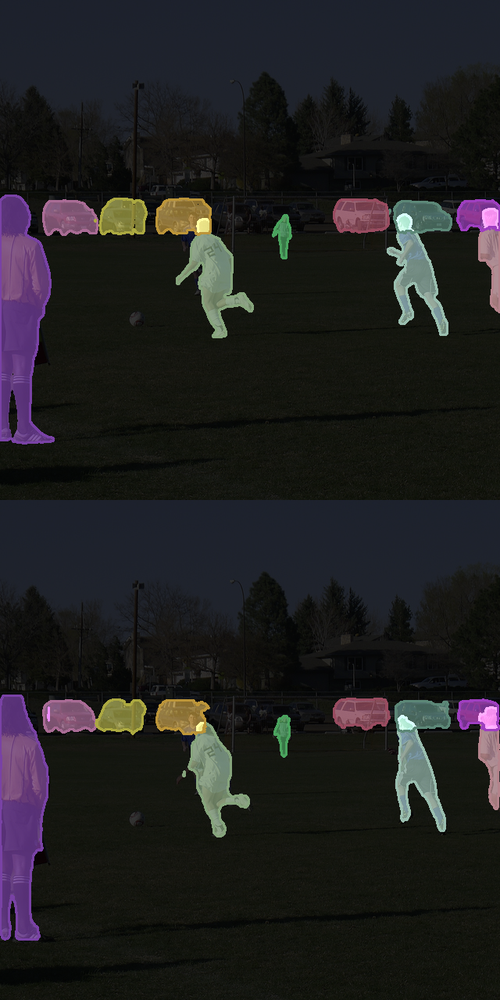}}
        \resizebox{!}{4.31cm}{\includegraphics{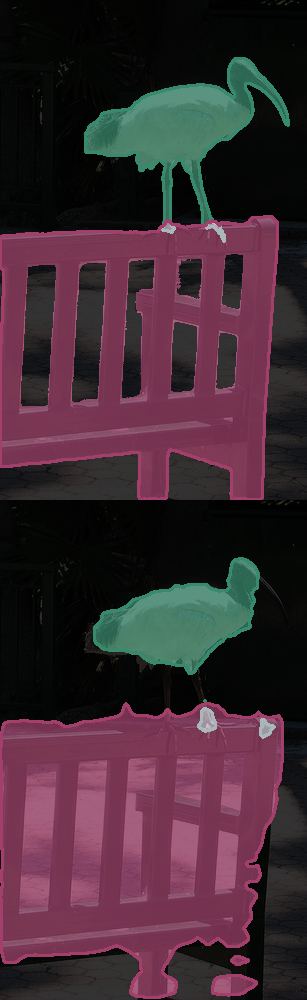}}
        \resizebox{!}{4.31cm}{\includegraphics{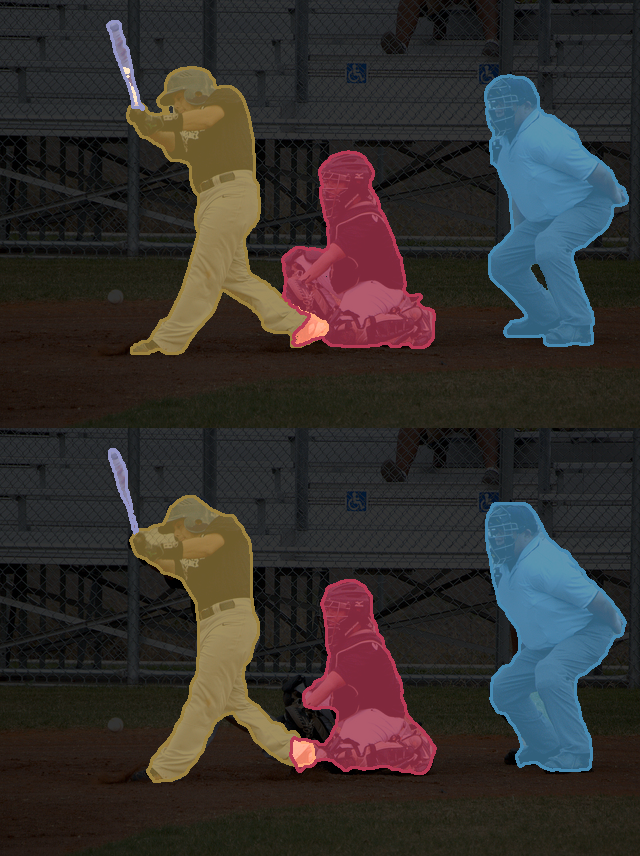}}
        \resizebox{!}{4.31cm}{\includegraphics{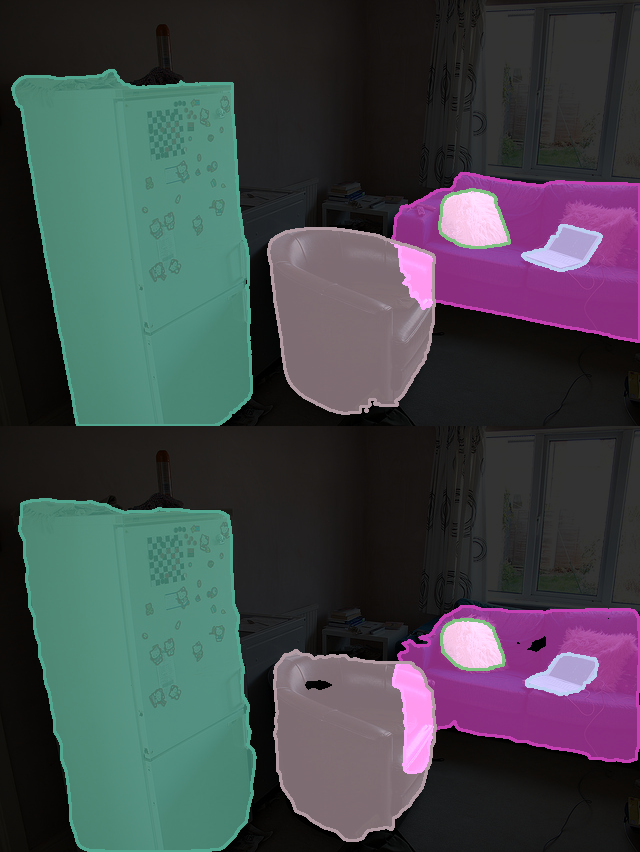}}
    \end{minipage}
    
    \vspace{1pt}
    \begin{minipage}{0.02\textwidth}
        \vspace{0.5cm}
        \rotatebox{90}{\textbf{Ours}} \\[0.6cm]
        \rotatebox{90}{\textbf{AISFormer}}
    \end{minipage}%
    \begin{minipage}{0.98\textwidth}
        \centering
        \resizebox{!}{4cm}{\includegraphics{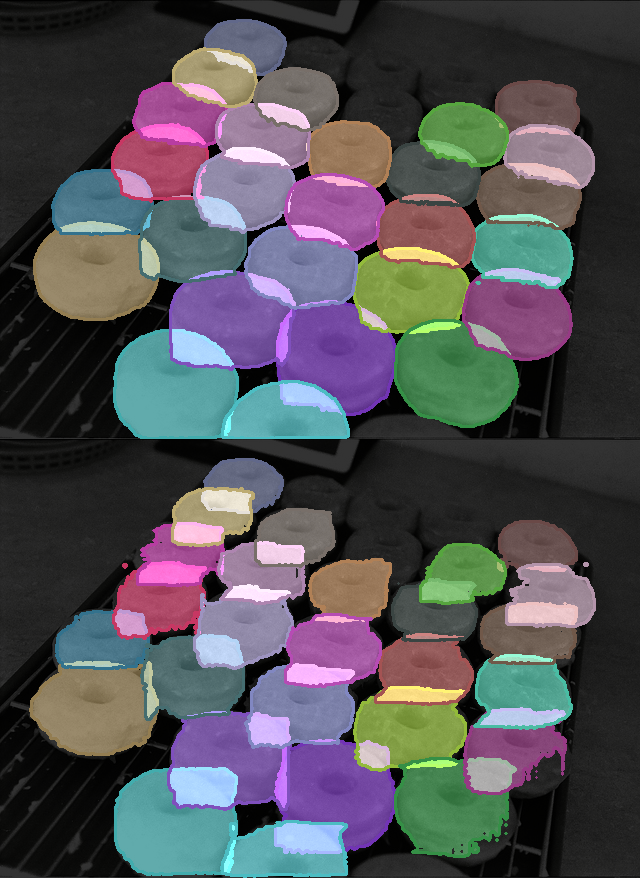}}
        \resizebox{!}{4cm}{\includegraphics{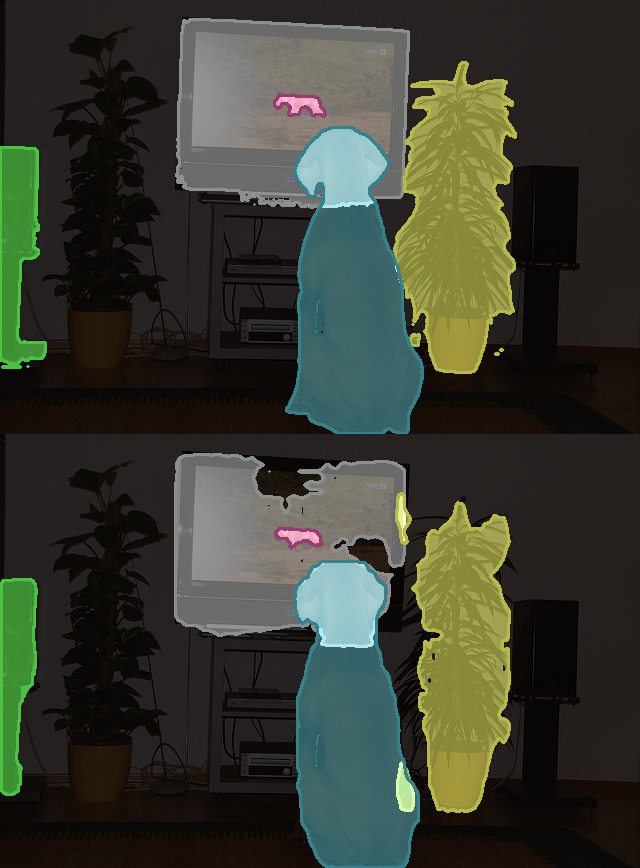}}
        \resizebox{!}{4cm}{\includegraphics{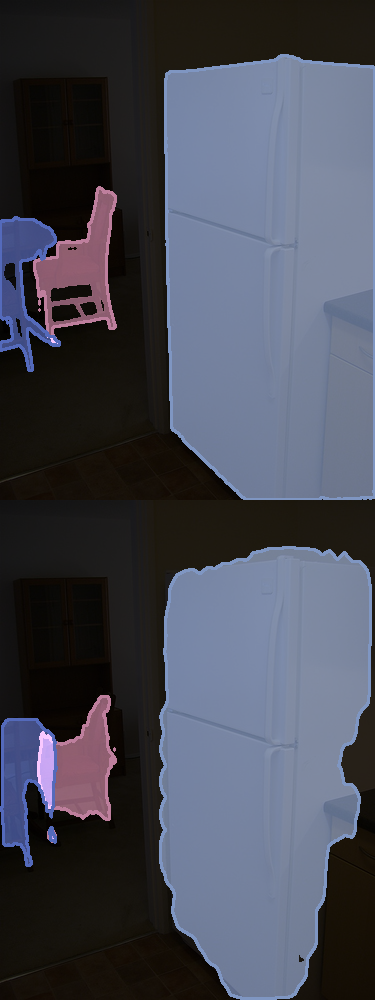}}
        \resizebox{!}{4cm}{\includegraphics{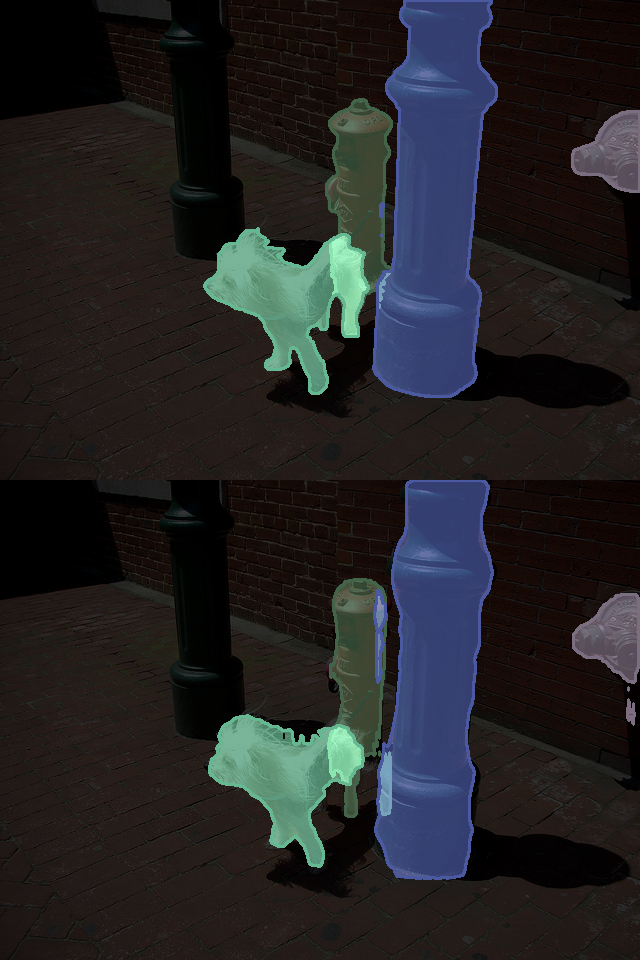}}
        \resizebox{!}{4cm}{\includegraphics{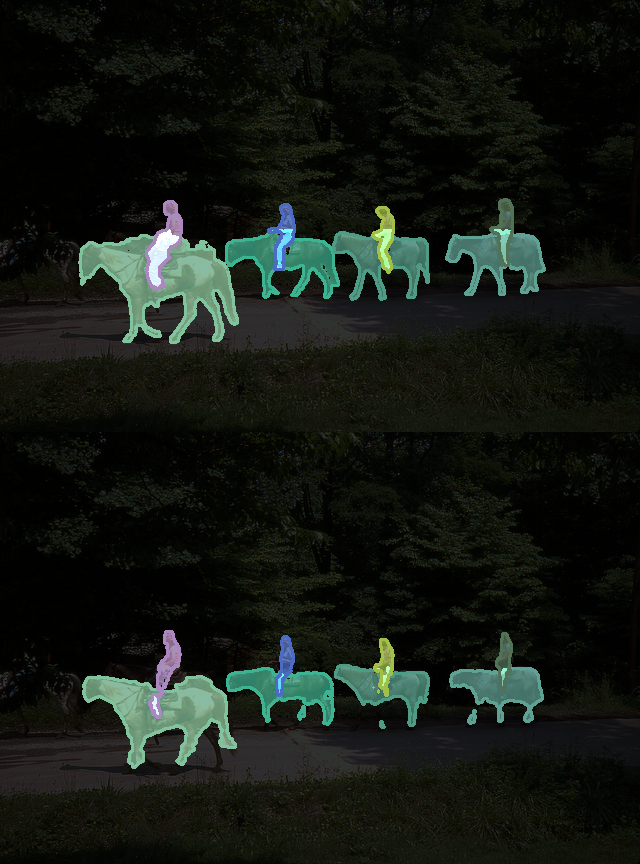}}
        \resizebox{!}{4cm}{\includegraphics{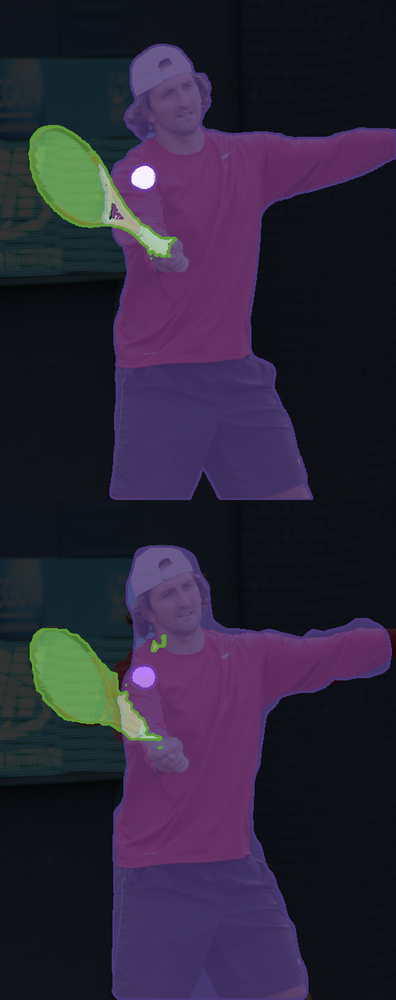}}
        \resizebox{!}{4cm}{\includegraphics{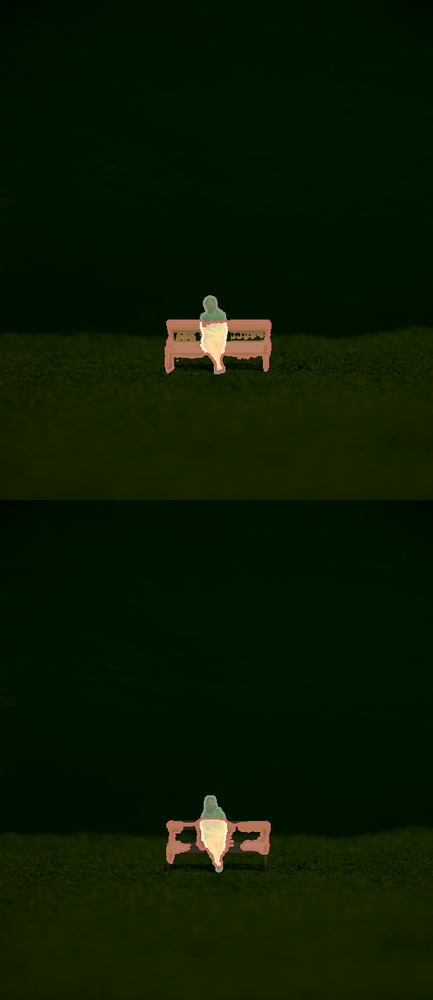}}
    \end{minipage}
    
    \vspace{1pt}
    \begin{minipage}{0.02\textwidth}
        \vspace{0.4cm}
        \rotatebox{90}{\textbf{Ours}} \\[0.8cm]
        \rotatebox{90}{\textbf{AISFormer}}
    \end{minipage}%
    \begin{minipage}{0.98\textwidth}
        \centering
        \resizebox{!}{4.335cm}{\includegraphics{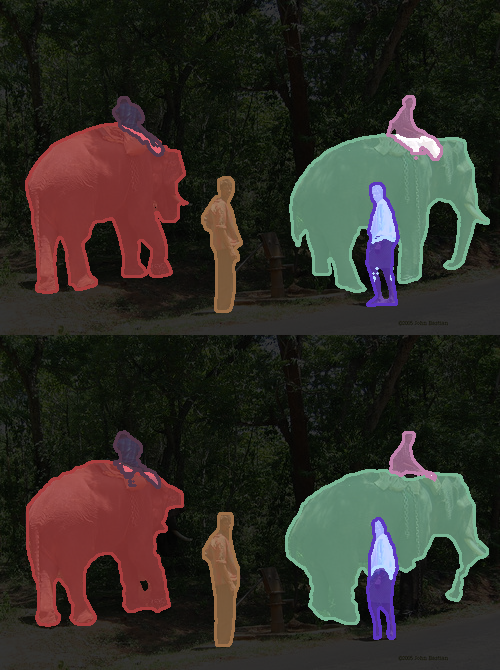}}
        \resizebox{!}{4.335cm}{\includegraphics{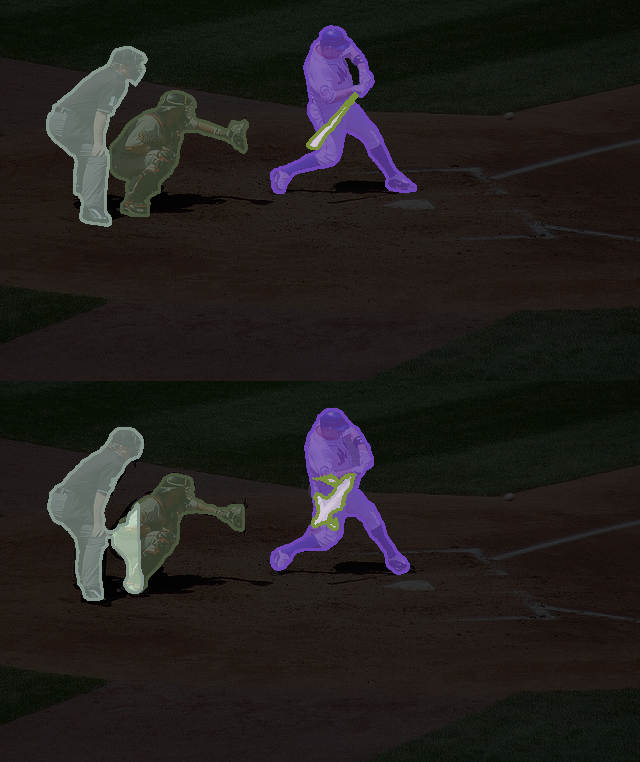}}
        \resizebox{!}{4.335cm}{\includegraphics{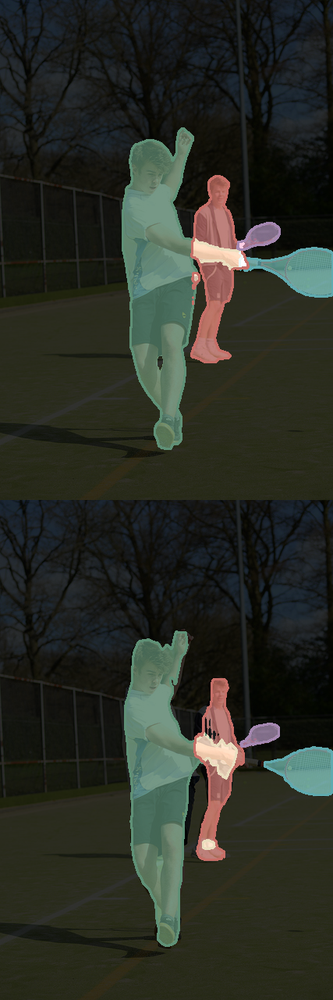}}
        \resizebox{!}{4.335cm}{\includegraphics{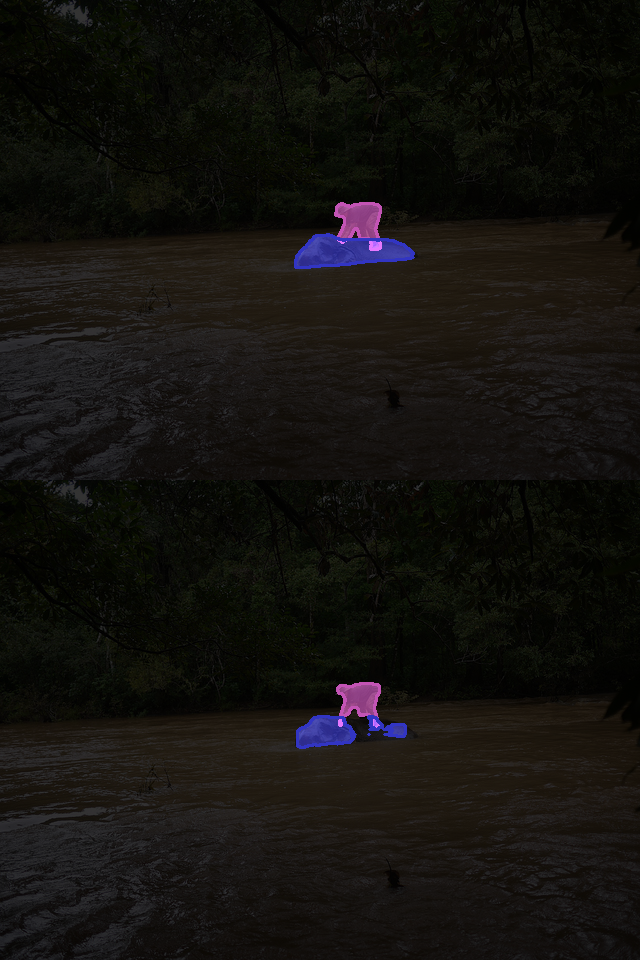}}
        \resizebox{!}{4.335cm}{\includegraphics{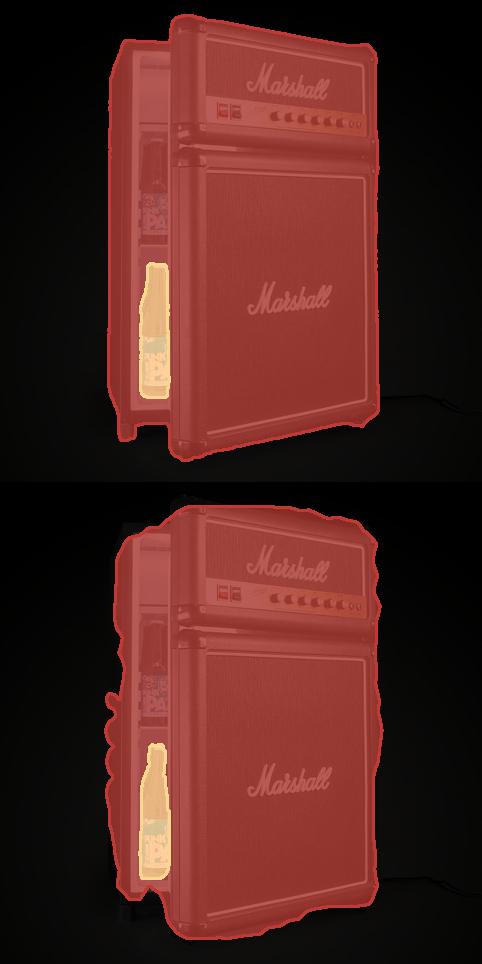}}
        \resizebox{!}{4.335cm}{\includegraphics{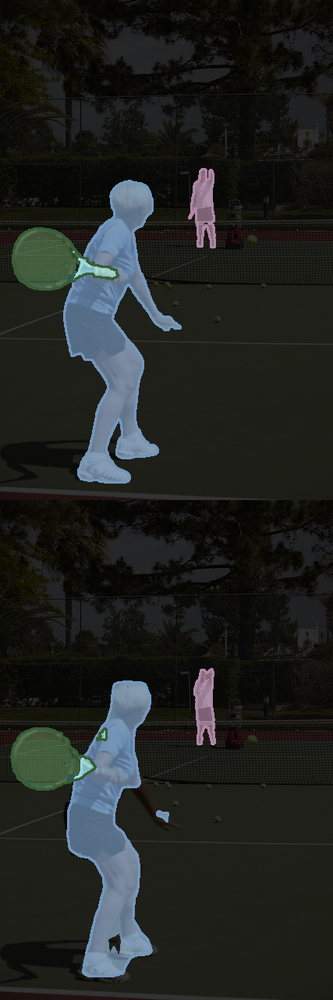}}
        \resizebox{!}{4.335cm}{\includegraphics{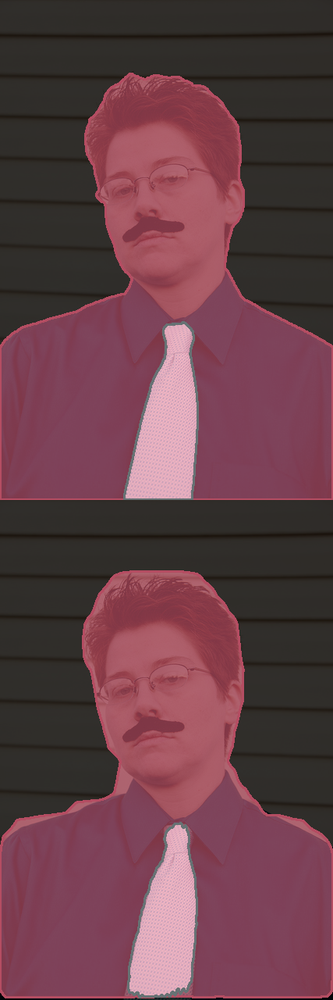}}
    \end{minipage}
    \vspace{0.5pt}

    \begin{minipage}{0.02\textwidth}
        \vspace{0.4cm}
        \rotatebox{90}{\textbf{Ours}} \\[0.6cm]
        \rotatebox{90}{\textbf{AISFormer}}
    \end{minipage}%
    \begin{minipage}{0.98\textwidth}
        \centering
        \resizebox{!}{3.7cm}{\includegraphics{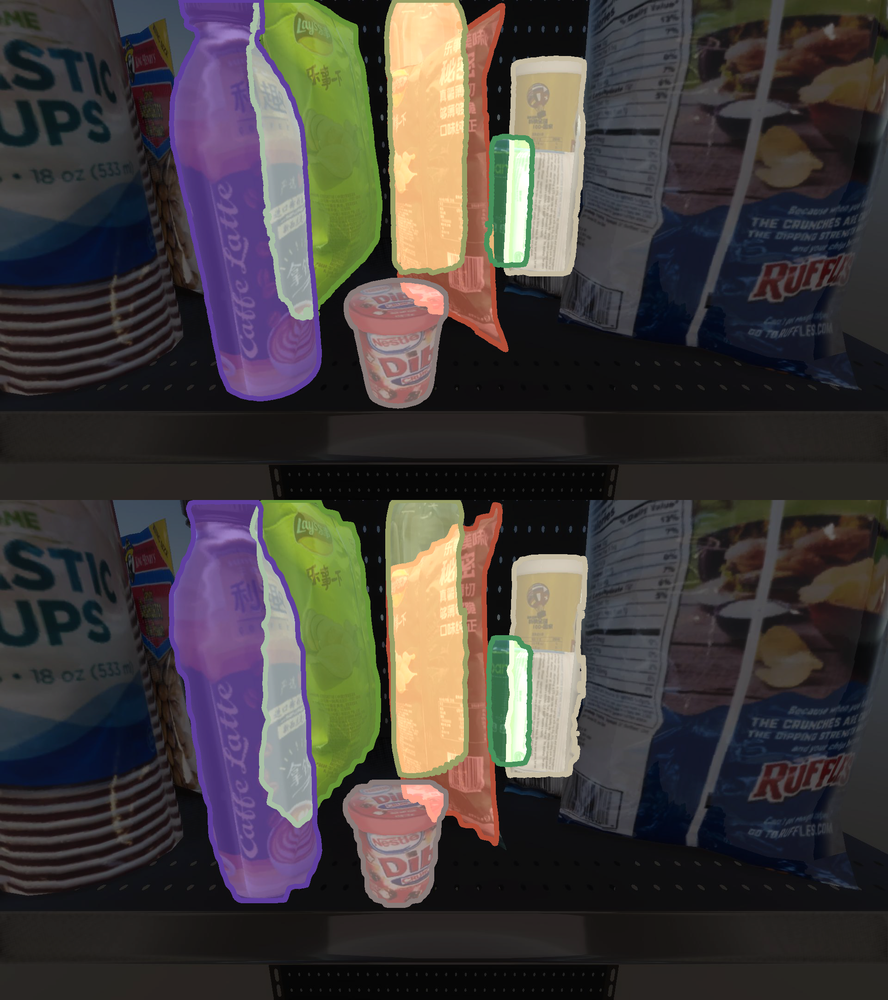}}
        \resizebox{!}{3.7cm}{\includegraphics{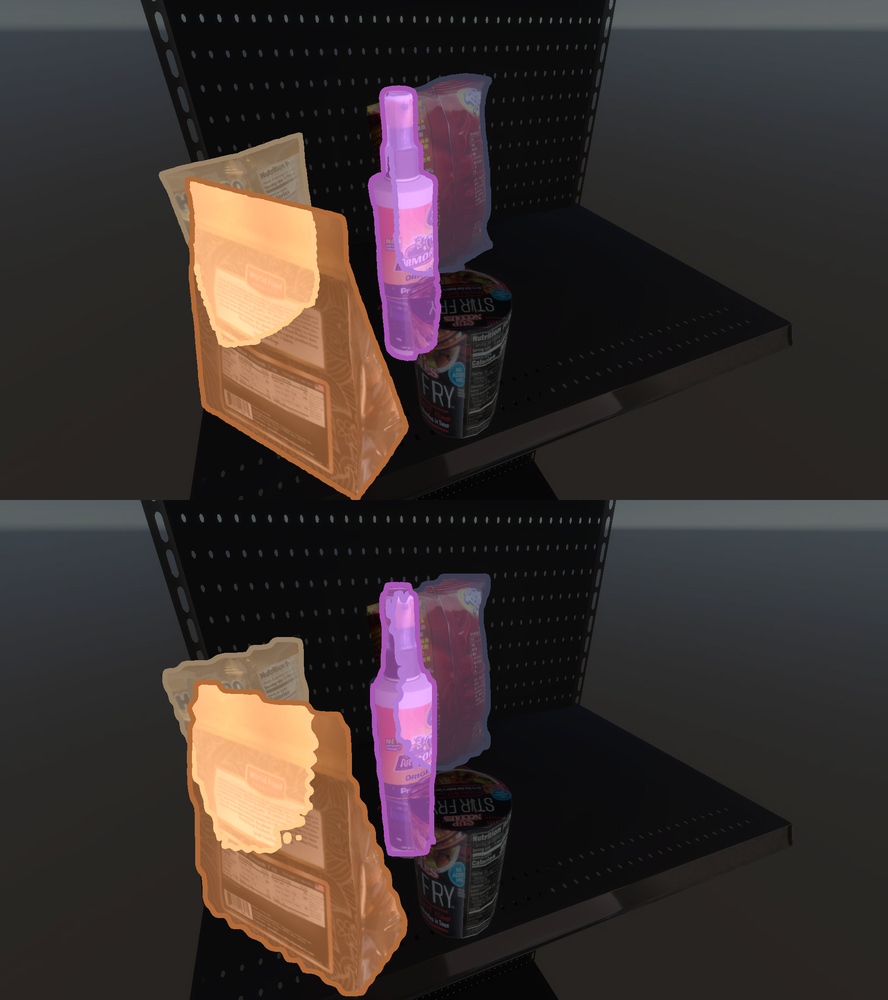}}
        \resizebox{!}{3.7cm}{\includegraphics{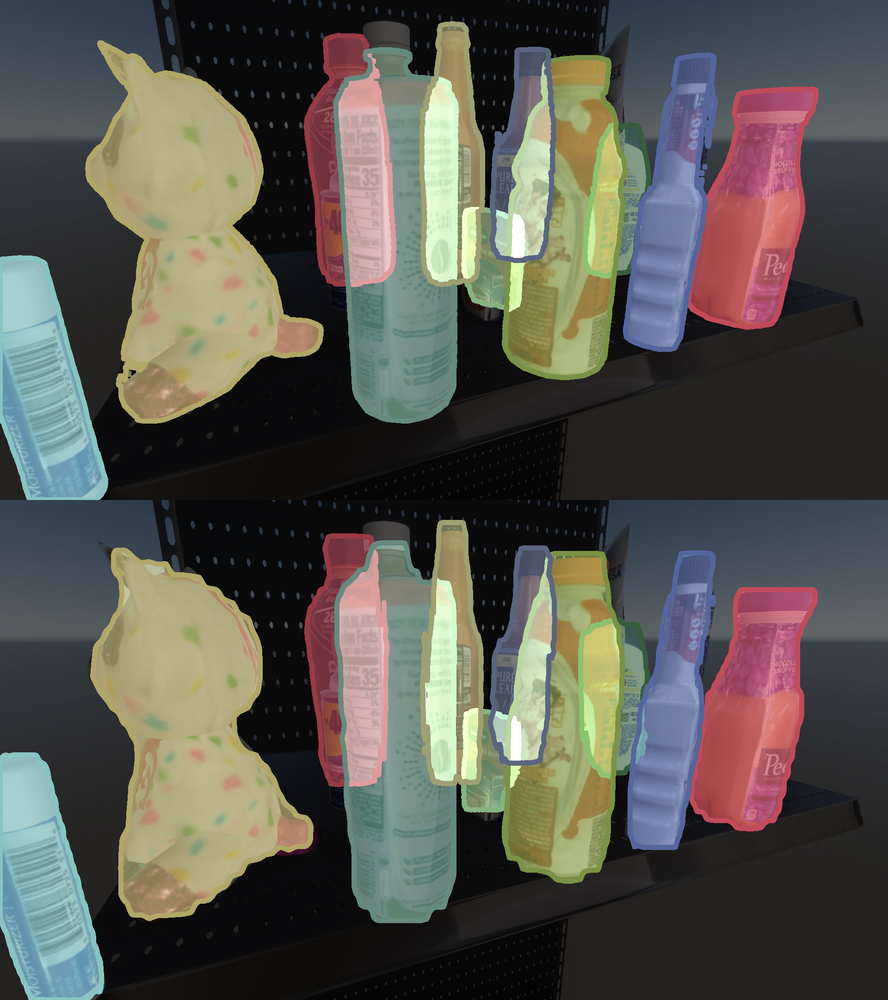}}
        \resizebox{!}{3.7cm}{\includegraphics{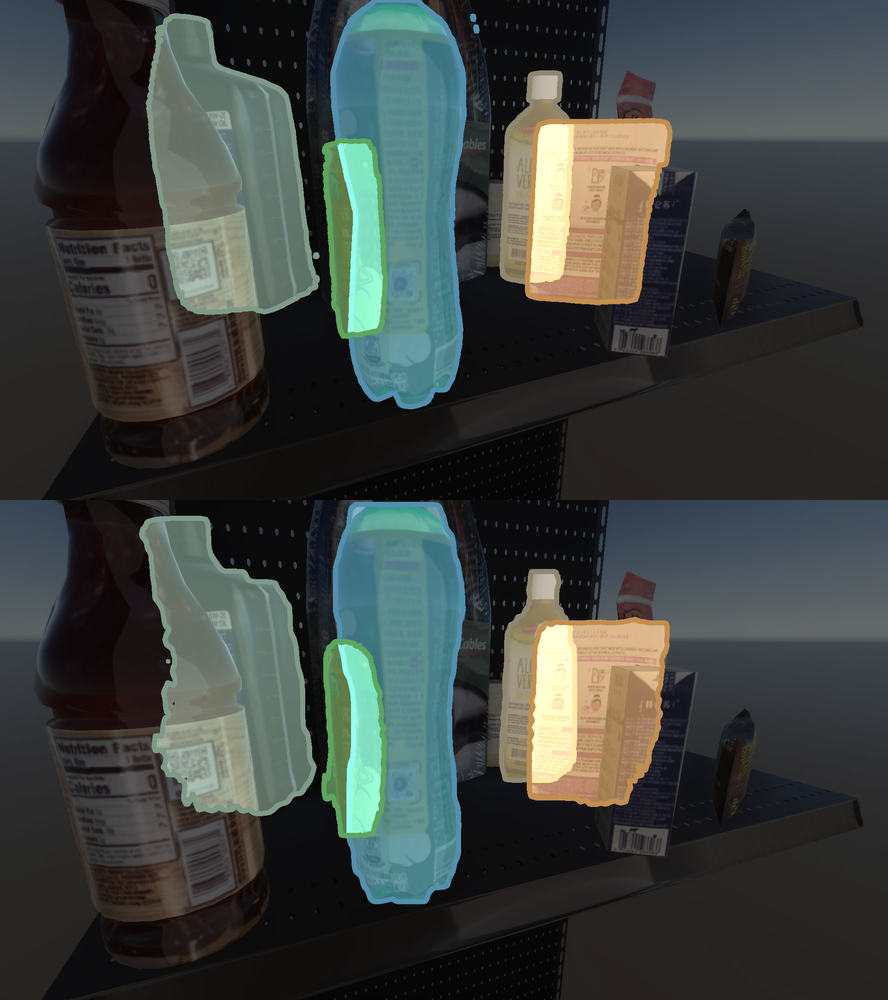}}
        \resizebox{!}{3.7cm}{\includegraphics{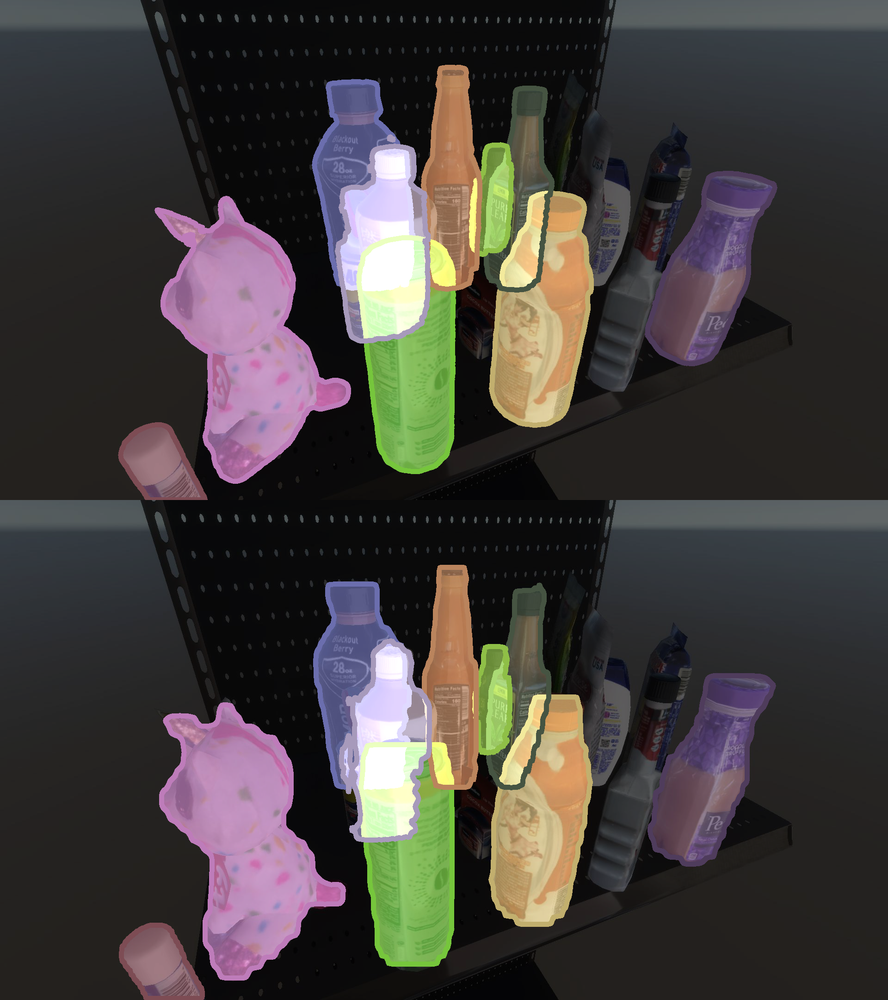}}
    \end{minipage}
    
    \vspace{1pt}
    
    \begin{minipage}{0.02\textwidth}
        \vspace{0.4cm}
        \rotatebox{90}{\textbf{Ours}} \\[0.5cm]
        \rotatebox{90}{\textbf{AISFormer}}
    \end{minipage}%
    \begin{minipage}{0.98\textwidth}
        \centering
        \resizebox{!}{3.7cm}{\includegraphics{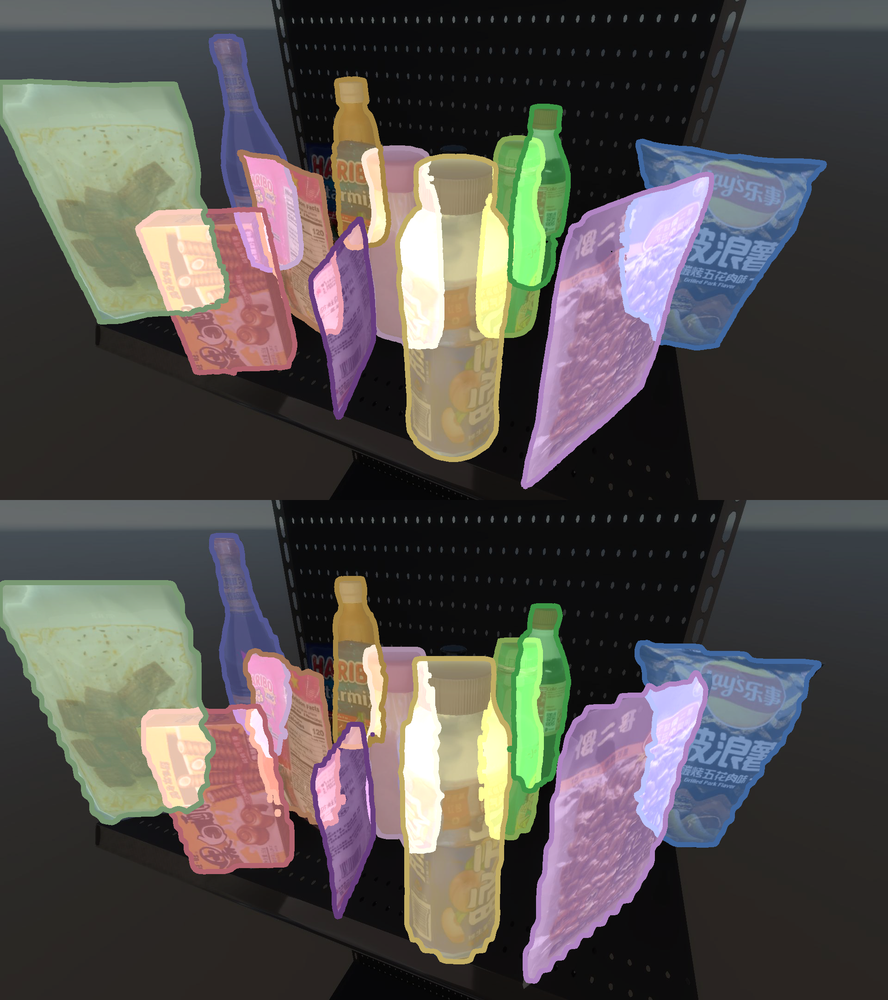}}
        \resizebox{!}{3.7cm}{\includegraphics{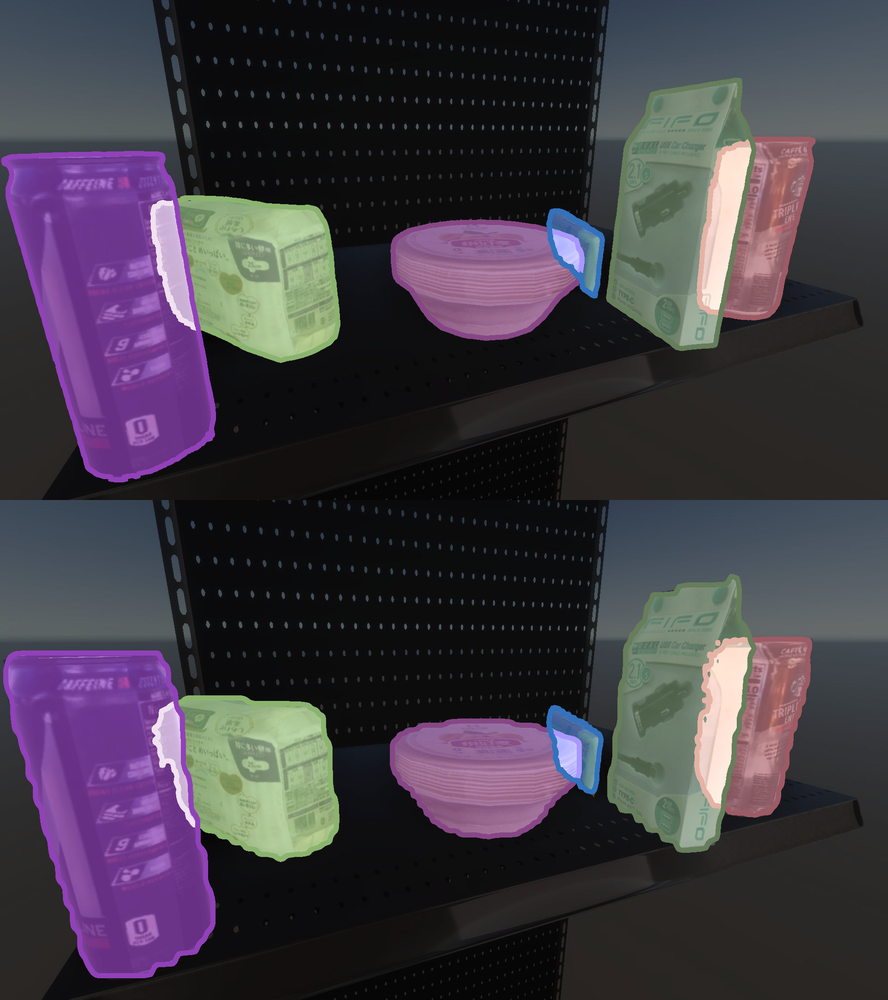}}
        \resizebox{!}{3.7cm}{\includegraphics{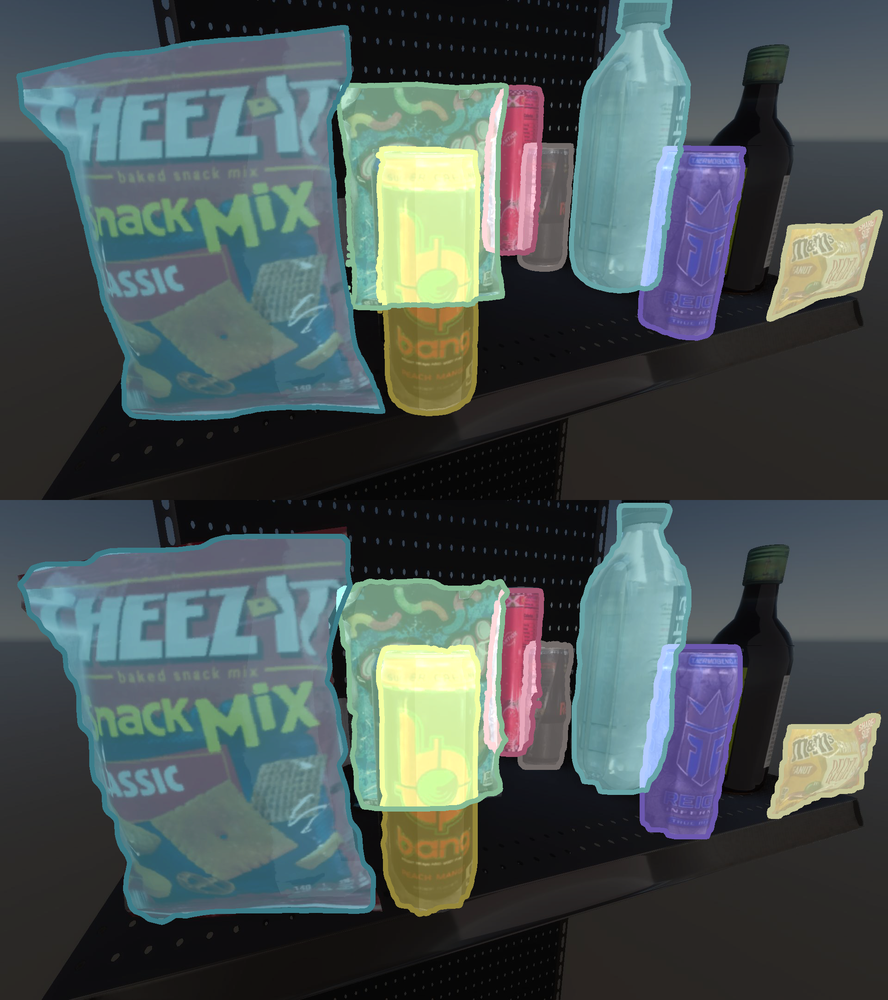}}
        \resizebox{!}{3.7cm}{\includegraphics{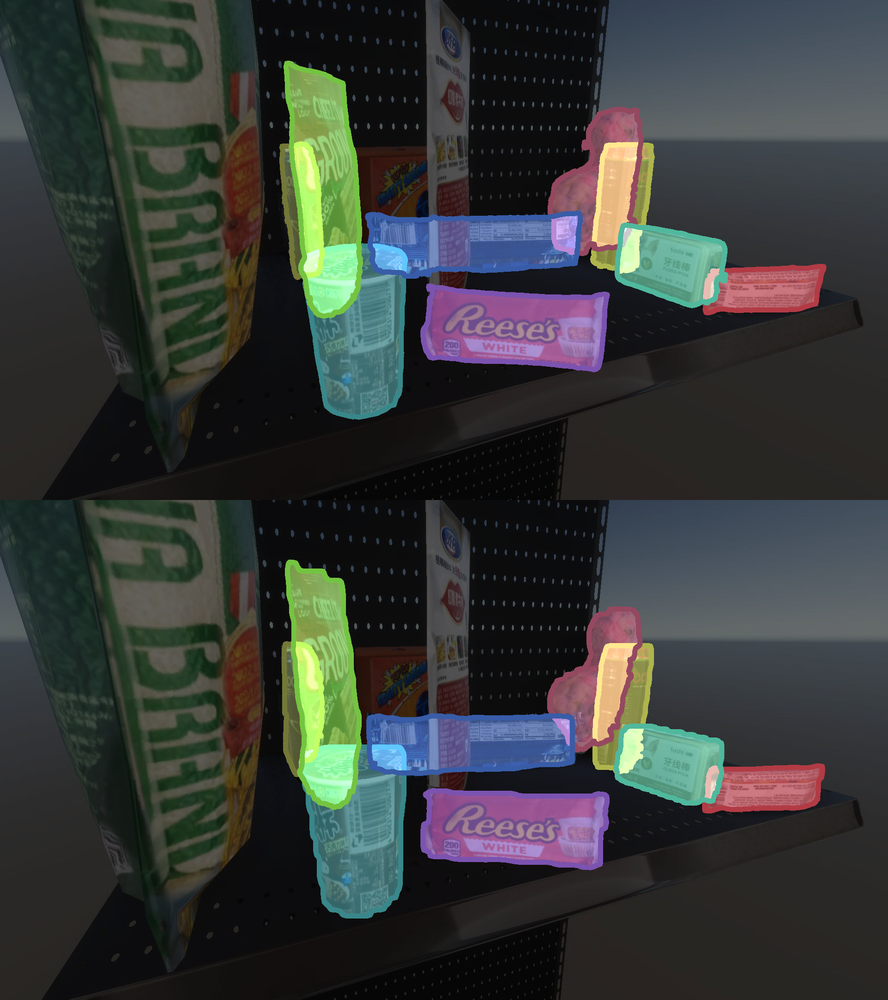}}
        \resizebox{!}{3.7cm}{\includegraphics{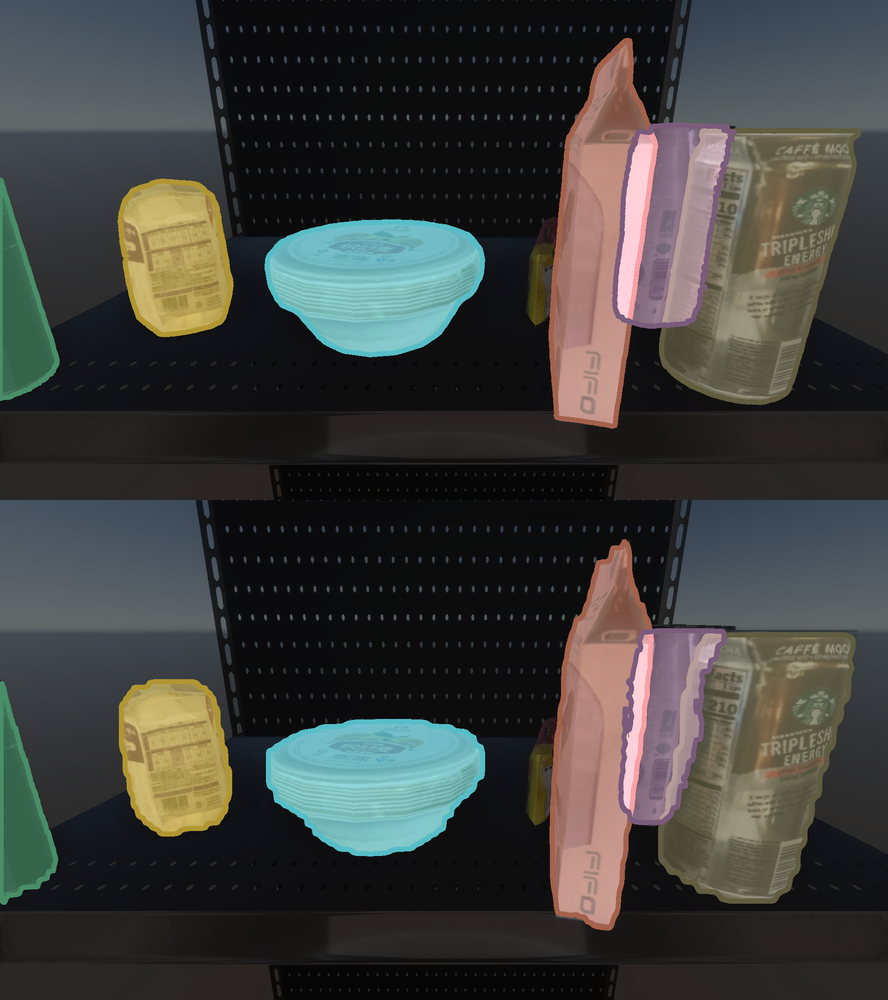}}
    \end{minipage}

    \caption{Qualitative comparison of amodal mask predictions. For each row: SAMEO's amodal prediction (top) with AISFormer box prompts, and AISFormer's prediction (bottom). Our method demonstrates superior mask quality, exhibiting more precise boundary delineation and robust handling of complex occlusion scenarios. Original images used for evaluation are available in supplementary materials.}
    \label{fig:aisformer_compare}
\end{figure*}

\paragraph{Quantitative Results.}
We evaluate our proposed SAMEO on three widely-used datasets: COCOA-cls, D2SA, and MUVA. For each dataset, we train our model on their respective training sets and evaluate on the corresponding test sets (\Cref{tab:nonzeroshot}). To demonstrate SAMEO's effectiveness and versatility, we attach it to various pre-trained front-end models, where modal front-ends are trained with modal annotations and amodal front-ends are trained with amodal annotations of these datasets. Our experimental results show that SAMEO significantly outperforms the current state-of-the-art method, AISFormer, achieving higher AP and AR across all datasets. Notably, our model exhibits robust performance in mask refinement independent of the front-end model's original mask type. Regardless of whether the front-end models produce modal or amodal mask predictions, SAMEO successfully refines them to achieve comparable high performance, demonstrating its strong capability to utilize both types of prompts.
\paragraph{Qualitative Results.}
For qualitative evaluation, we compare our model's predictions against AISFormer on both COCOA-cls and MUVA datasets (\Cref{fig:aisformer_compare}). Our method exhibits superior performance across various challenging scenarios, including complex still life arrangements with multiple overlapping objects (\eg, bottles and containers), scenes with intricate occlusions (\eg, people behind barriers), and diverse object categories and poses. Results show that our model generates significantly more precise amodal masks with sharper boundaries while providing more reasonable predictions for occluded parts. The qualitative comparison clearly demonstrates our method's improvements over the baseline method in both mask quality and occlusion reasoning capabilities, validating the effectiveness of SAMEO for real-world amodal segmentation tasks.

\subsection{Zero-shot Performance}
To evaluate SAMEO's zero-shot generalization capability, we train our model on our dataset collection and the proposed Amodal-LVIS dataset, excluding COCOA-cls and D2SA. During training, for each batch, a dataset is sampled with probability proportional to the logarithm of its size divided by the sum of log sizes across all datasets. We then test these two held-out datasets to demonstrate zero-shot performance (\Cref{tab:ap_zeroshot}). For comparison, we include AISFormer and RTMDet (both trained on target datasets) combined with the original EfficientSAM, showing that our model successfully adapts EfficientSAM for amodal segmentation while preserving its zero-shot capability. Furthermore, we experiment with various pre-trained modal front-end detectors to demonstrate SAMEO's robust zero-shot performance regardless of the front-end choice. 

The results demonstrate SAMEO's superior performance, achieving up to 13.8 AP improvement over AISFormer on COCOA-cls with RTMDet and 8.7 AP over on D2SA with CO-DETR, reaching state-of-the-art results and validating its strength as a robust zero-shot amodal segmentation solution.

\subsection{Ablation Study}
\begin{table}
\centering
\begin{tabular}{c@{\hskip 0.3in}cccc}
\toprule
\multirow{2}{*}{\textbf{IoU Prediction}} & \multicolumn{3}{c}{\textbf{COCOA-cls}} \\
& \textbf{AP} & \textbf{AP$_{50}$} & \textbf{AP$_{75}$} \\
\midrule
$\times$ & 52.4 & 73.2 & 57.8 \\
$\checkmark$ & \textbf{54.3} & \textbf{74.0} & \textbf{59.7} \\

\bottomrule
\end{tabular}
\caption{Ablation study of IoU prediction refinement on COCOA-cls dataset, using AISFormer as front-end. $\times$ and $\checkmark$ indicate without and with IoU prediction refinement, respectively.}
\label{tab:ablation_iou}
\end{table}

\paragraph{Effect of IoU Prediction.}
To validate the effectiveness of the IoU prediction branch in our model, we have conducted experiments comparing the performance metrics before and after confidence score refinement using SAMEO's predicted IoU. Using AISFormer as our front-end model and evaluating this setting on the COCOA-cls dataset, the experimental results demonstrate that SAMEO's precise IoU prediction significantly contributes to improving the ranking of segmentation results (\Cref{tab:ablation_iou}). Specifically, we observe that incorporating the predicted IoU for confidence score refinement leads to notable improvements in AP metric, confirming that the IoU prediction branch plays a crucial role in enhancing the overall performance of our model.

\begin{table}
\centering
\begin{tabular}{c@{\hskip 0.3in}cccc}
\toprule
\multirow{2}{*}{\textbf{Box Prompt}} & \multicolumn{4}{c}{\textbf{COCOA-cls}} \\
& \textbf{AP} & \textbf{AP$_{50}$} & \textbf{AP$_{75}$} & \textbf{AR}  \\
\midrule
amodal & 53.0 & 72.9 & 58.0 & 71.1 \\
modal & 53.7 & 73.3 & 59.3 & 71.2 \\
random & \textbf{54.2} & \textbf{73.5} & \textbf{59.5} & \textbf{71.6} \\

\bottomrule
\end{tabular}
\caption{Comparison of different prompt types during training. We evaluate three variants on the COCOA-cls dataset using both modal and amodal front-ends, reporting the averaged performance over both types. The results show that training with random box prompts has optimal performance across diverse front-end models.}
\label{tab:ablation_prompt}
\end{table}

\paragraph{Impact of Training Prompt Types.}
We investigate the optimal prompt-type strategy for training SAMEO to achieve balanced performance across both modal and amodal front-end prompts. We train three variants of SAMEO using ground truth amodal boxes, modal boxes, and a random mixture of both with equal probability. For evaluation, we integrate each trained model variant with both amodal and modal front-end detectors and evaluate their performance separately. The final performance metric is calculated by averaging the AP and AR scores across both front-end scenarios (\Cref{tab:ablation_prompt}). Our findings reveal that training with random boxes with equal probability yields the most balanced performance when handling various front-end prompt types, demonstrating the model's ability to generalize across different input scenarios.

\begin{figure}
    \centering
    \begin{tabular}{@{\hspace{0.4cm}}c@{\hspace{0.7cm}}c@{\hspace{0.4cm}}c@{\hspace{0.3cm}}c@{}}
    \small original & \small EfficientSAM & \small SAMEO(mix) & \small SAMEO(occ) \\
    \end{tabular}
    \includegraphics[width=0.115\textwidth]{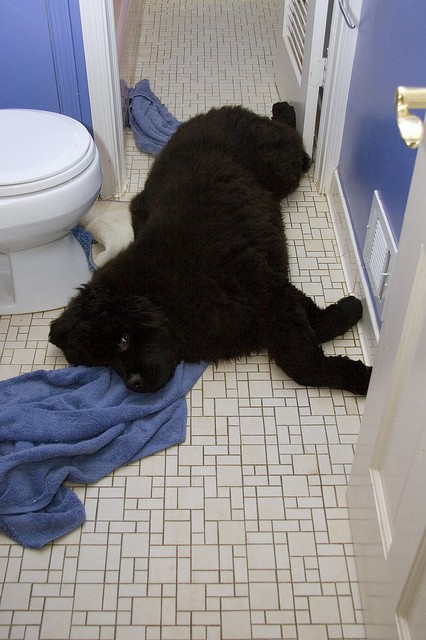}
    \includegraphics[width=0.115\textwidth]{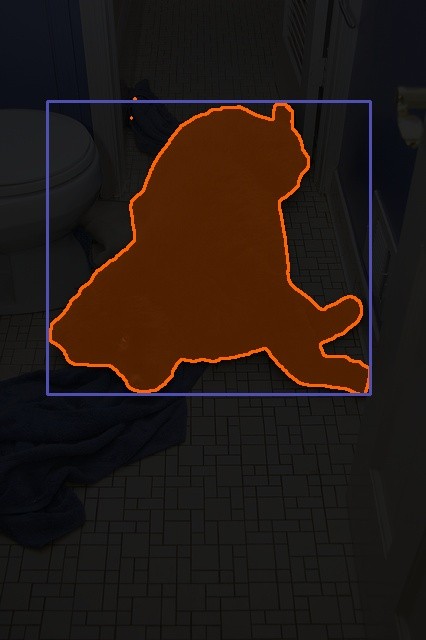}
    \includegraphics[width=0.115\textwidth]{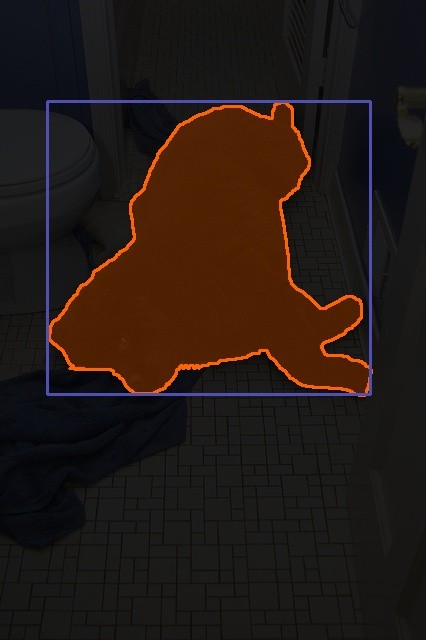}
    \includegraphics[width=0.115\textwidth]{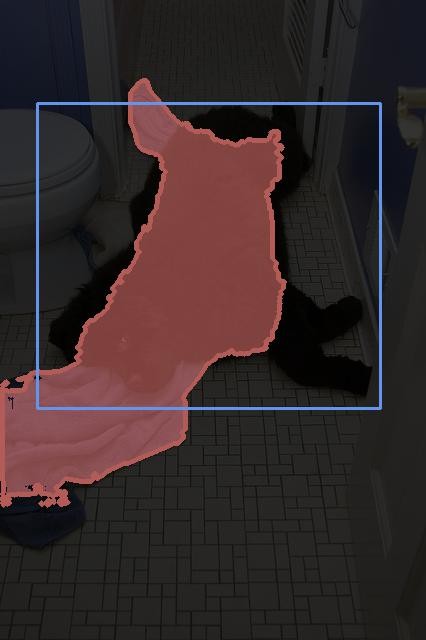}
    
    \caption{Visualization of over-prediction. Given a detector box clearly intended for the foreground dog, SAMEO(occ), trained solely on occluded instances, mistakenly predicts the mask of the background towel. After rebalancing our dataset to include both occluded and non-occluded instances, SAMEO(mix) produces results that closely match the EfficientSAM baseline.}
    \label{fig:amodal_bad_ex}
\end{figure}

\paragraph{Dataset Composition Analysis.}
To understand the importance of diverse instance types in training data, we have conducted an experiment training SAMEO exclusively on datasets containing only occluded instances (\eg, pix2gestalt). Visualization results reveal a significant limitation: the model exhibits over-prediction of background instances, even when the input box prompt clearly indicates a foreground object (\Cref{fig:amodal_bad_ex}). This observation motivates our design choice for the Amodal-LVIS dataset, which maintains an equal distribution of occluded and non-occluded annotations. This balanced composition prevents bias and ensures robustness across various scenarios.

\section{Conclusion}
We present a flexible approach to amodal instance segmentation by adapting foundation segmentation models to handle both visible and occluded portions of objects. Our framework successfully leverages pre-trained modal detectors while maintaining strong amodal segmentation capabilities. The introduction of Amodal-LVIS, containing 300K carefully curated images, along with our comprehensive collection of 1M images and 2M instance annotations, addresses critical limitations in existing datasets and provides the necessary scale for robust model development.

Our extensive experiments demonstrate that SAMEO consistently outperforms state-of-the-art methods on COCOA-cls, D2SA, and MUVA datasets. Most notably, when trained on our dataset collection, including Amodal-LVIS, SAMEO achieves strong zero-shot performance on unseen datasets. The model's robust generalization abilities persist across various front-end detectors, validating our approach of adapting foundation models for amodal segmentation without compromising performance. We further address the limitations and possible future work of SAMEO in the appendix.

{
    \normalem
    \small
    \bibliographystyle{ieeenat_fullname}
    \bibliography{main}
}
\clearpage
\renewcommand\thepage{\Roman{page}}
\setcounter{page}{1}
\maketitlesupplementary
\renewcommand\thesection{\Alph{section}}
\setcounter{section}{0}
\renewcommand\thefigure{\Alph{figure}}
\setcounter{figure}{0}
\renewcommand\thetable{\Alph{table}}
\setcounter{table}{0}

\section*{Appendix}
\label{sec:appendix}
This supplementary document provides additional experimental results and visualizations supporting our main paper. 
\begin{itemize}
\item \Cref{sec:amodal_dataset_vis} presents visual examples from our collected amodal datasets.
\item \Cref{sec:comp_ais} illustrates qualitative comparisons between SAMEO and AISFormer~\cite{Tran0YFKL22}.
\item \Cref{sec:comp_effsam} shows the adaptation from modal to amodal segmentation compared to EfficientSAM~\cite{XiongVWXXZDWSIK24}.
\item \Cref{sec:ap_with_cls} extends our quantitative evaluation with class-specific metrics.
\item \Cref{sec:failure} highlights the limitations of SAMEO and suggests potential directions for future research.
\end{itemize}

\section{Amodal Dataset Visualization}
\label{sec:amodal_dataset_vis}
\begin{figure*}
\centering

\begin{tabular}{c@{\hspace{4pt}}cccc}
& \textbf{COCOA} & \textbf{COCOA-cls} & \textbf{DYCE} & \textbf{KINS} \\
\rotatebox{90}{\makebox[0.094\textheight][c]{\small Image}} &
\includegraphics[height=0.094\textheight]{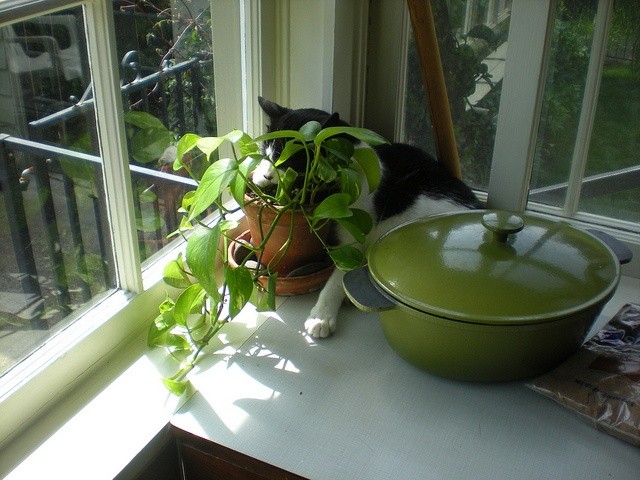} & 
\includegraphics[height=0.094\textheight]{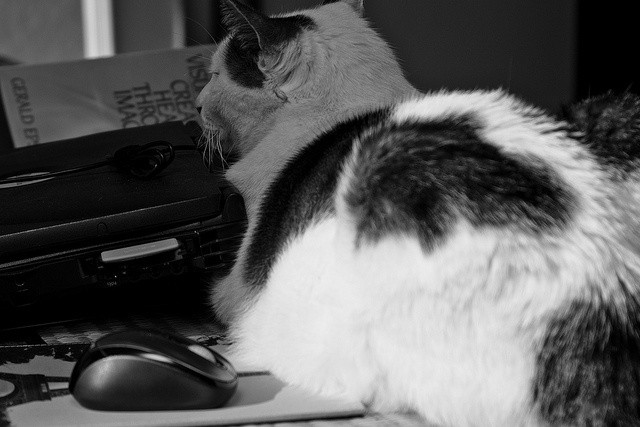} &
\includegraphics[height=0.094\textheight]{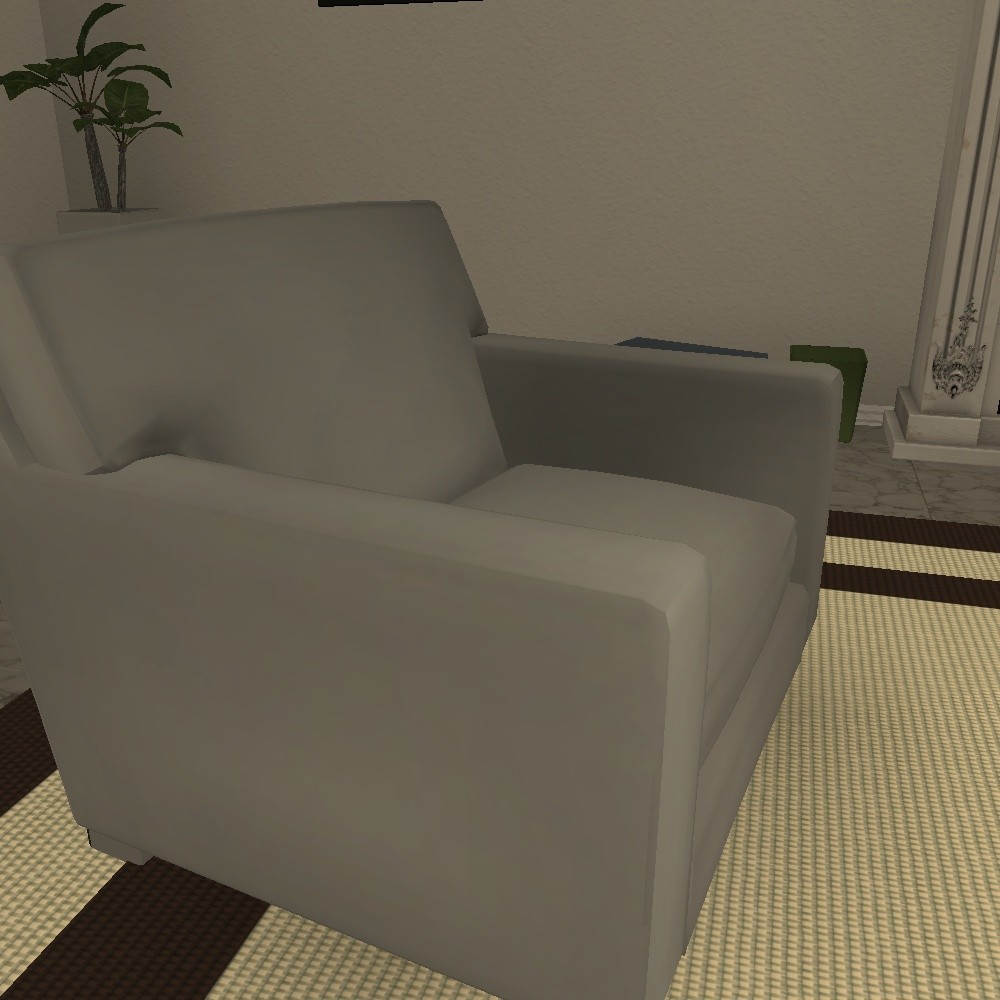} & 
\includegraphics[height=0.094\textheight]{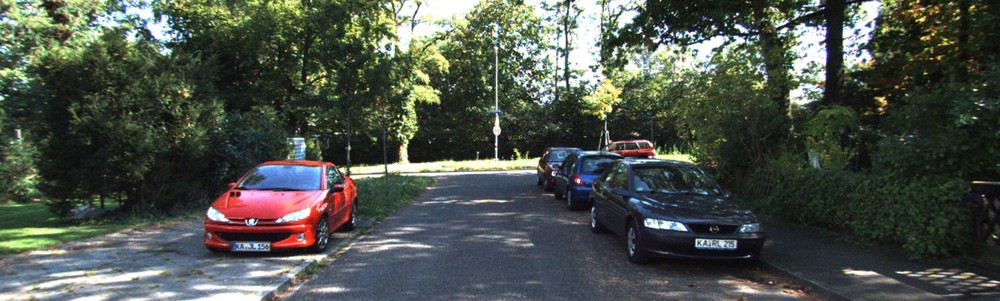} \\
\rotatebox{90}{\makebox[0.094\textheight][c]{\small Modal mask}} &
\includegraphics[height=0.094\textheight]{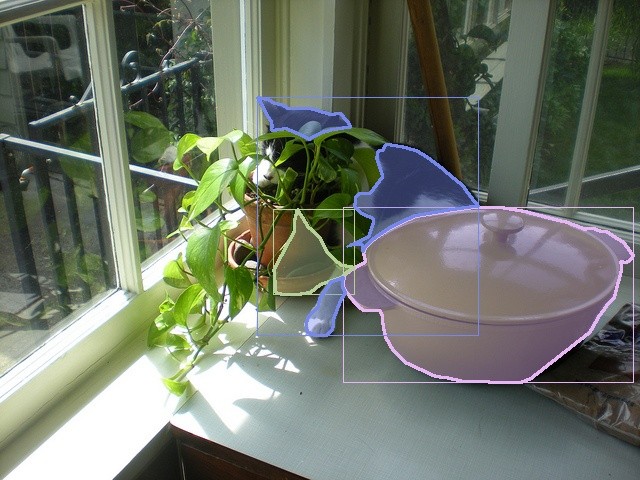} & 
\includegraphics[height=0.094\textheight]{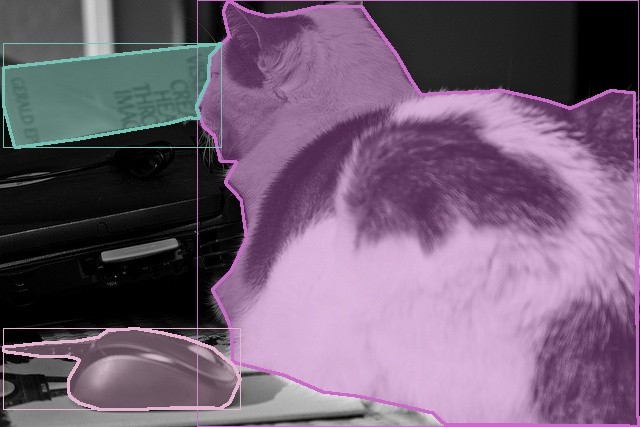} &
\includegraphics[height=0.094\textheight]{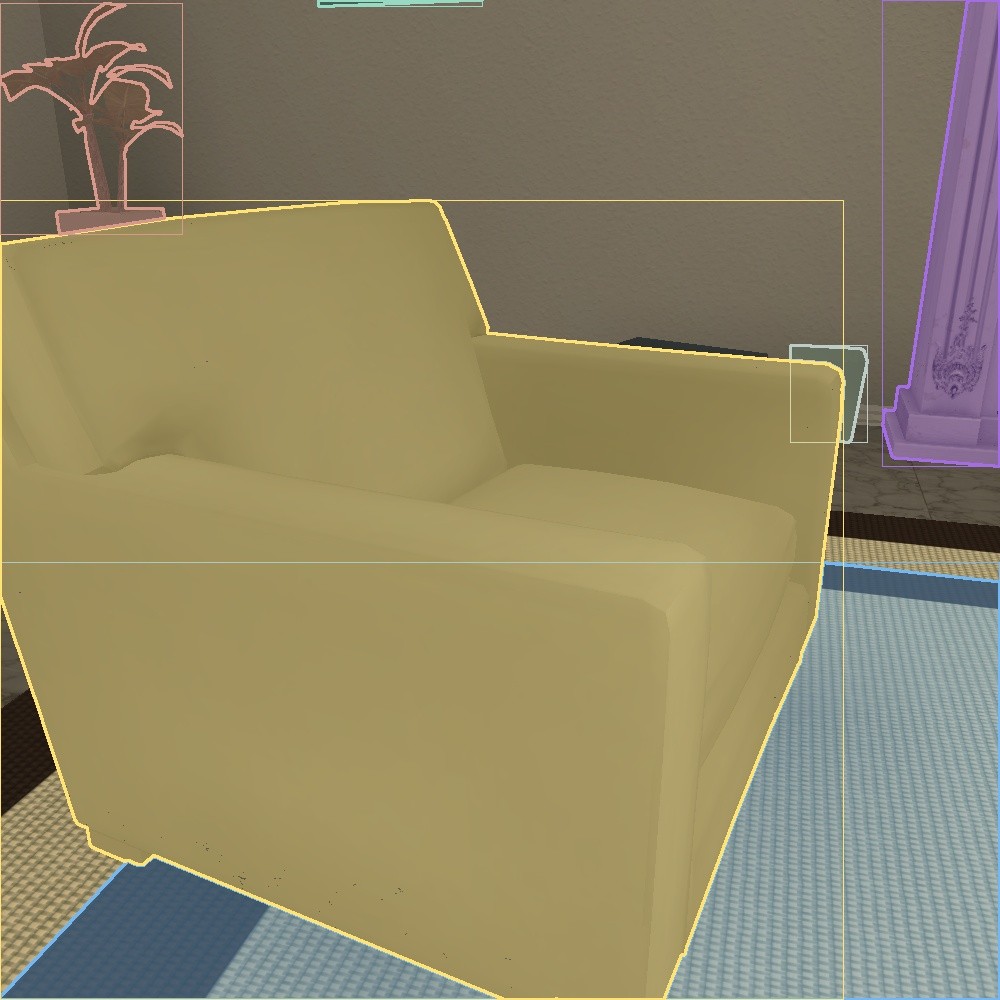} &
\includegraphics[height=0.094\textheight]{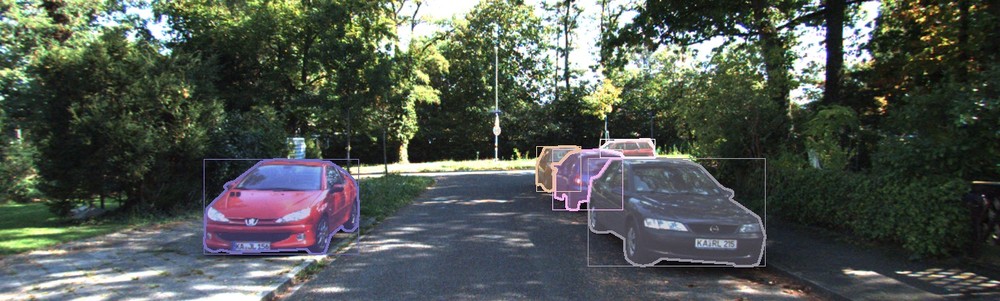} \\
\rotatebox{90}{\makebox[0.094\textheight][c]{\small Amodal mask}} &
\includegraphics[height=0.094\textheight]{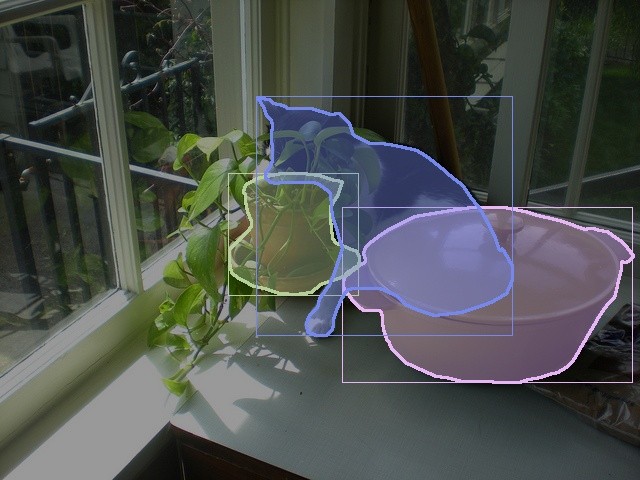} & 
\includegraphics[height=0.094\textheight]{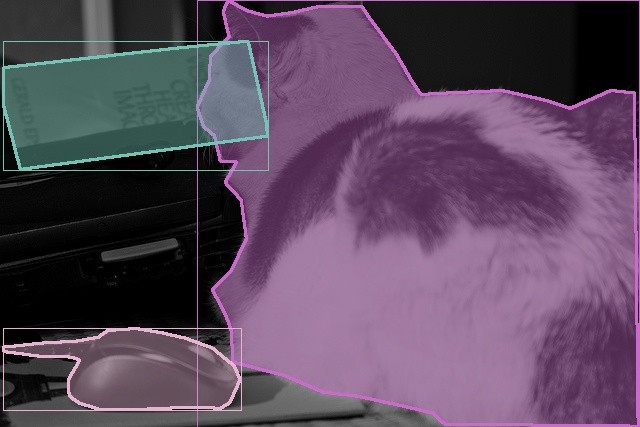} &
\includegraphics[height=0.094\textheight]{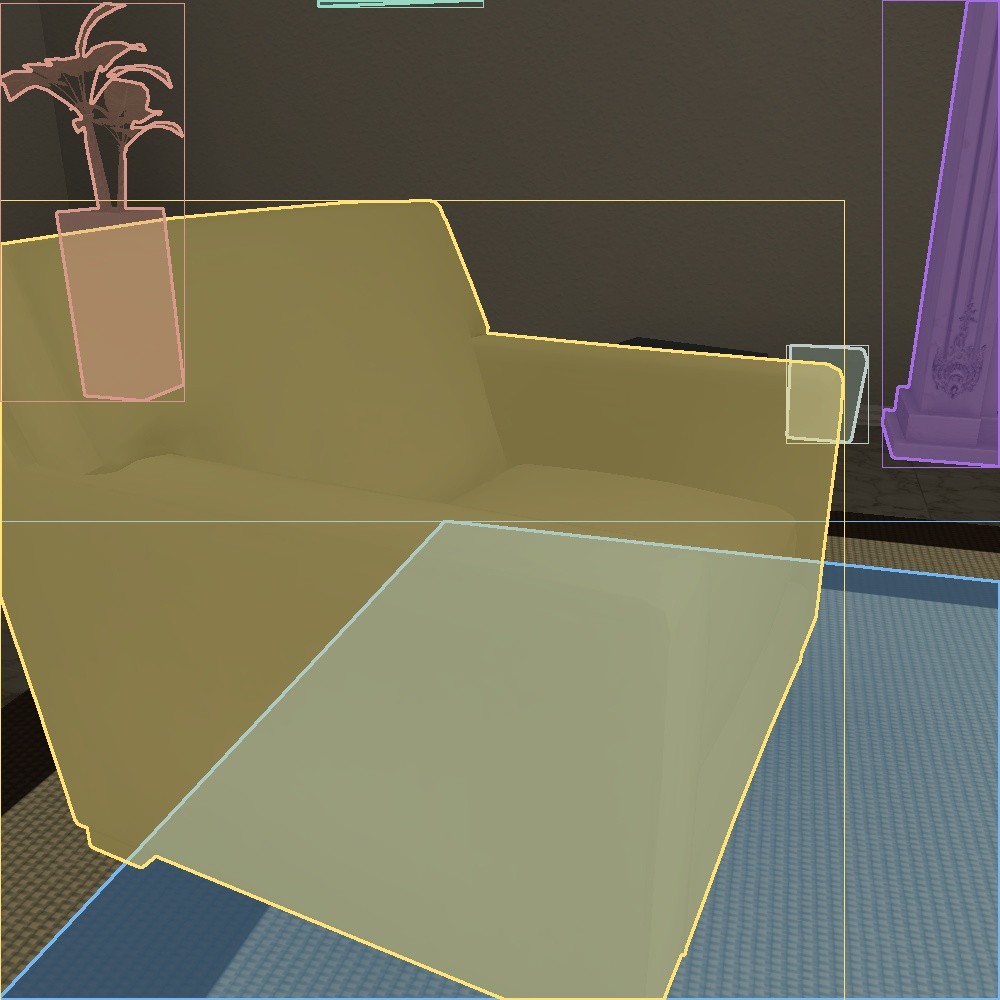} &
\includegraphics[height=0.094\textheight]{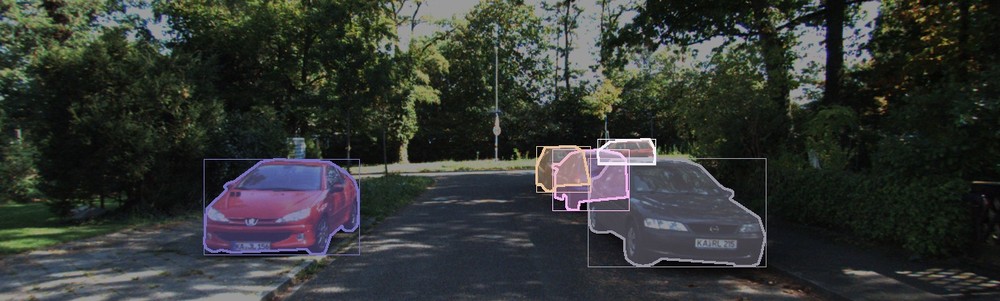}
\end{tabular}
\begin{tabular}{@{}c@{\hspace{4pt}}ccc@{}}
& \textbf{MUVA} & \textbf{D2SA} & \textbf{\small KITTI-360-APS} \\
\rotatebox{90}{\makebox[0.1\textheight][c]{Image}} &
\includegraphics[height=0.1\textheight]{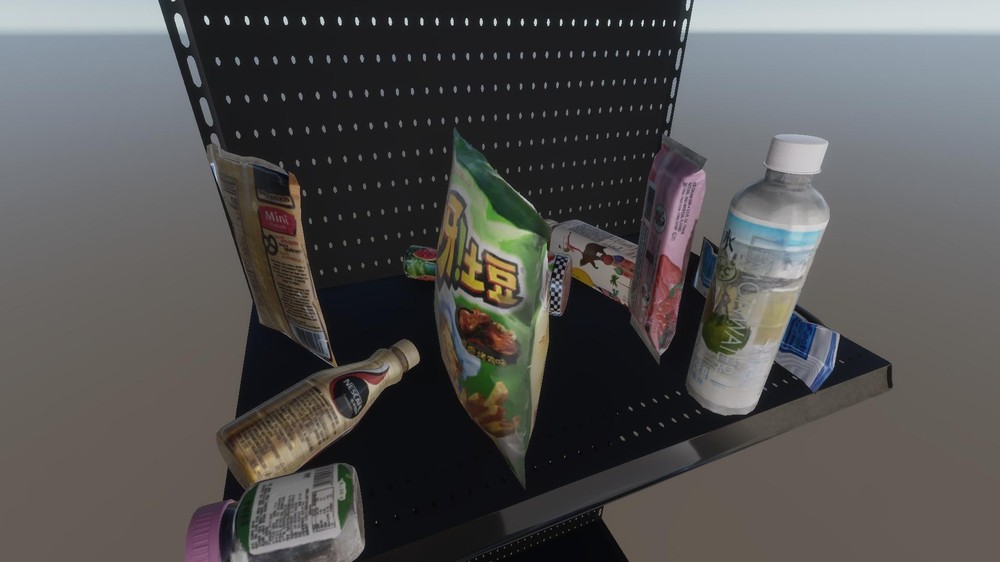} &
\includegraphics[height=0.1\textheight]{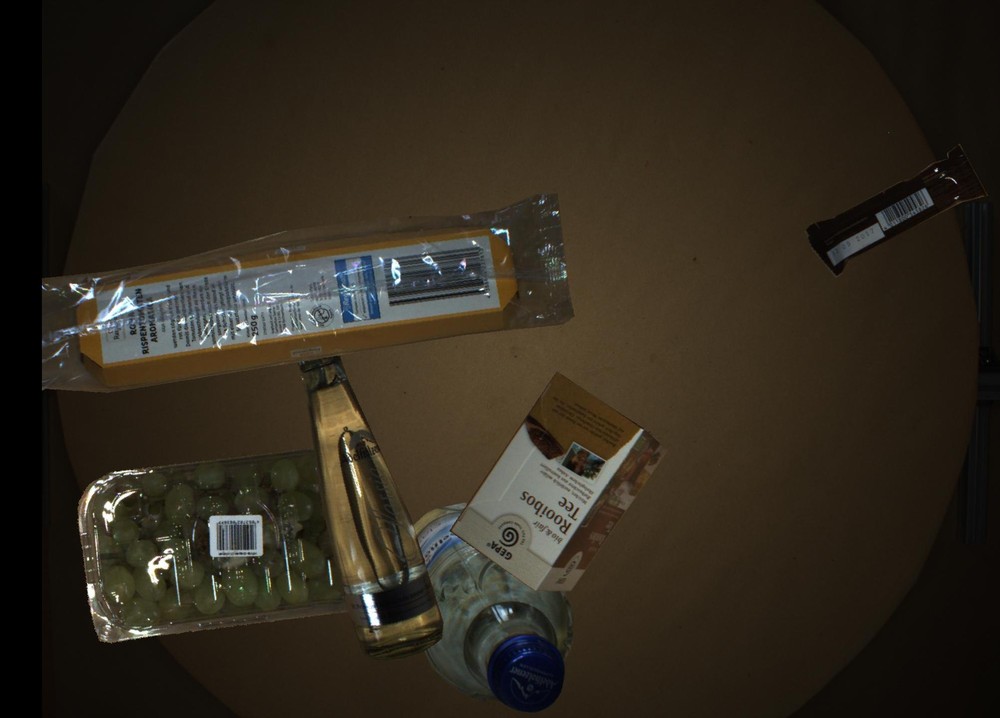} &
\includegraphics[height=0.1\textheight]{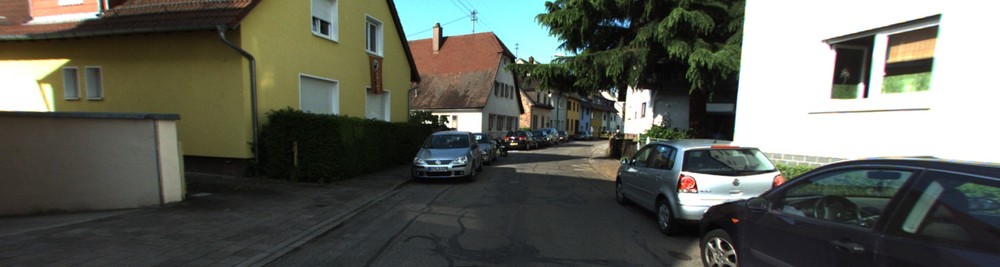} \\
\rotatebox{90}{\makebox[0.1\textheight][c]{\small Modal mask}} &
\includegraphics[height=0.1\textheight]{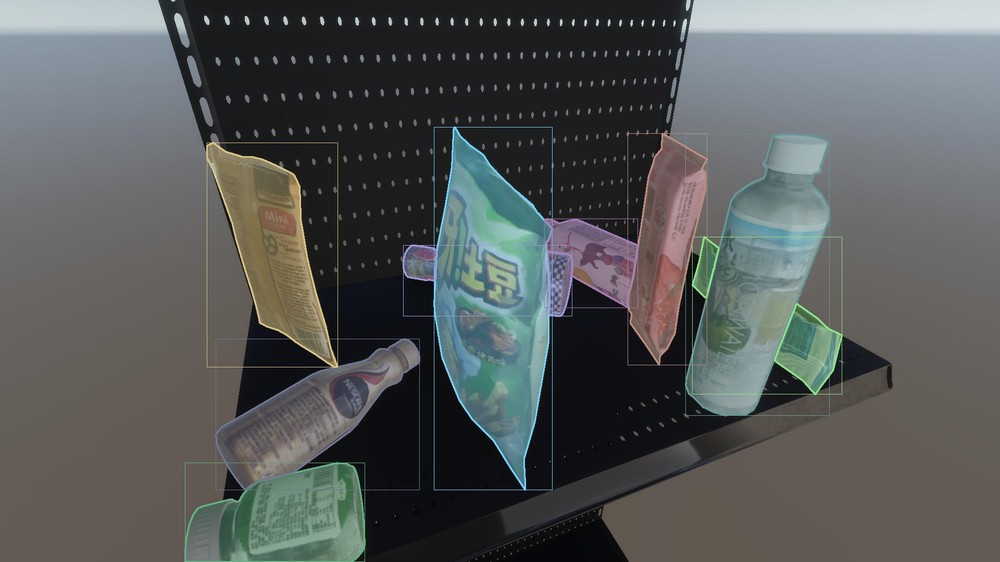} &
\includegraphics[height=0.1\textheight]{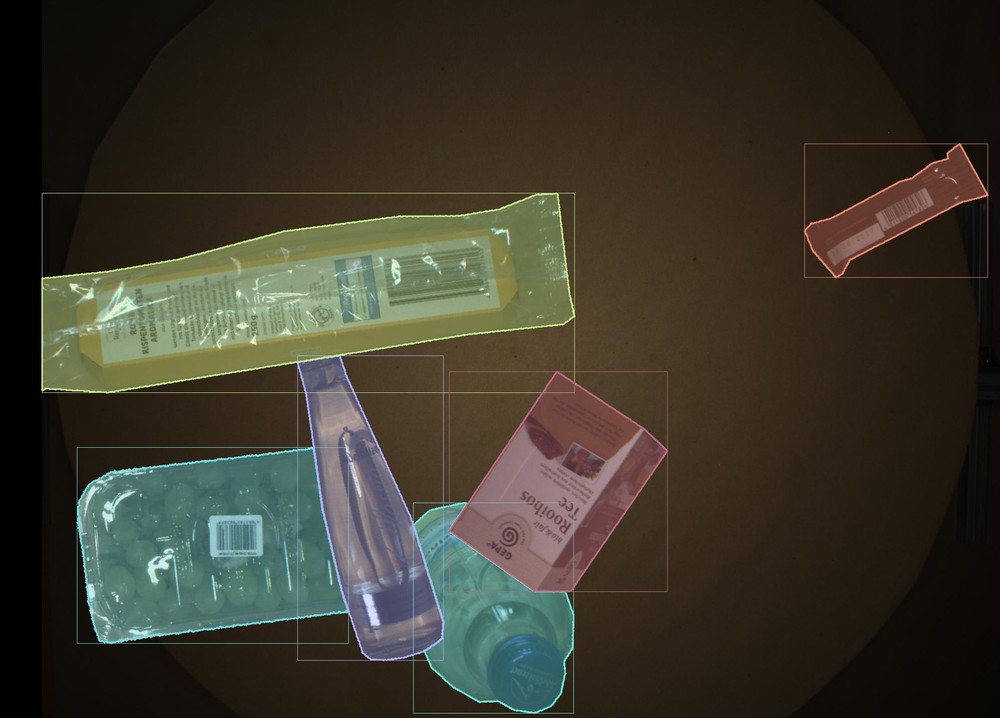} &
\includegraphics[height=0.1\textheight]{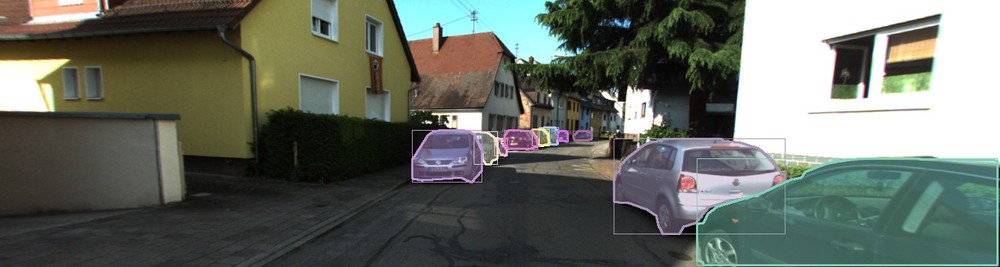} \\
\rotatebox{90}{\makebox[0.1\textheight][c]{\small Amodal mask}} &
\includegraphics[height=0.1\textheight]{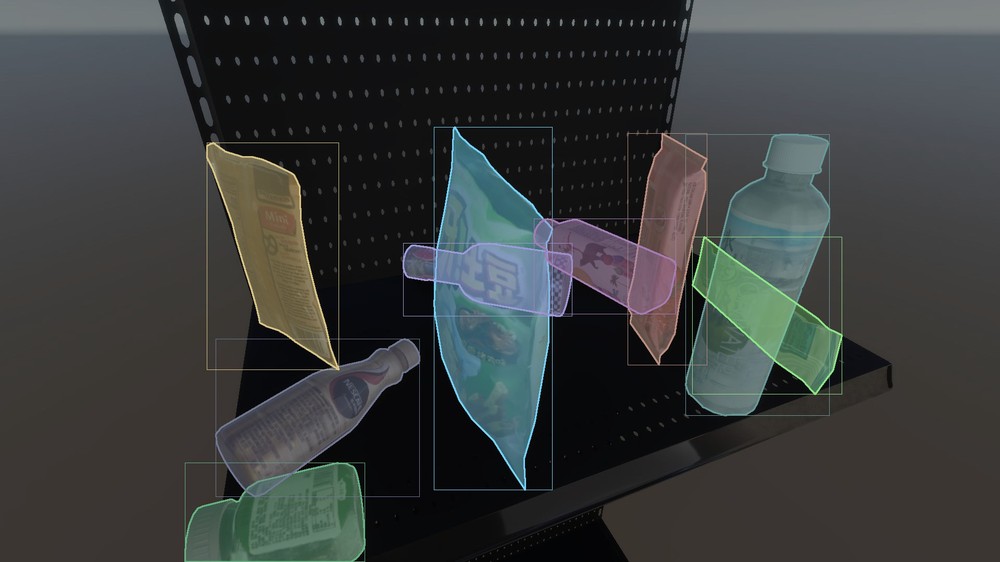} &
\includegraphics[height=0.1\textheight]{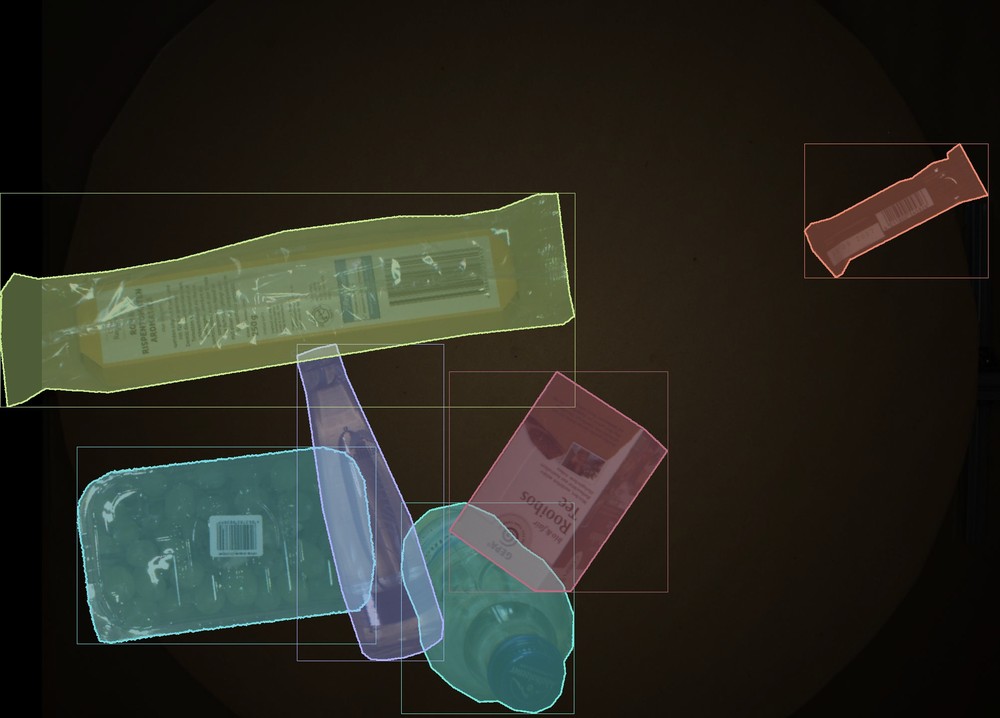} & 
\includegraphics[height=0.1\textheight]{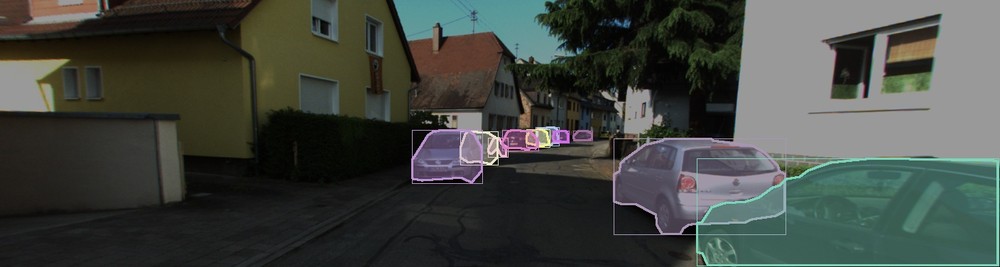}\\
\end{tabular}

\begin{tabular}{c@{\hspace{4pt}}ccccc}
& \textbf{MP3D-amodal} & \textbf{WALT} & \textbf{pix2gestalt} & \multicolumn{2}{c}{\textbf{Amodal-LVIS}}\\
\rotatebox{90}{\makebox[0.099\textheight][c]{\small Image}} &
\includegraphics[height=0.099\textheight]{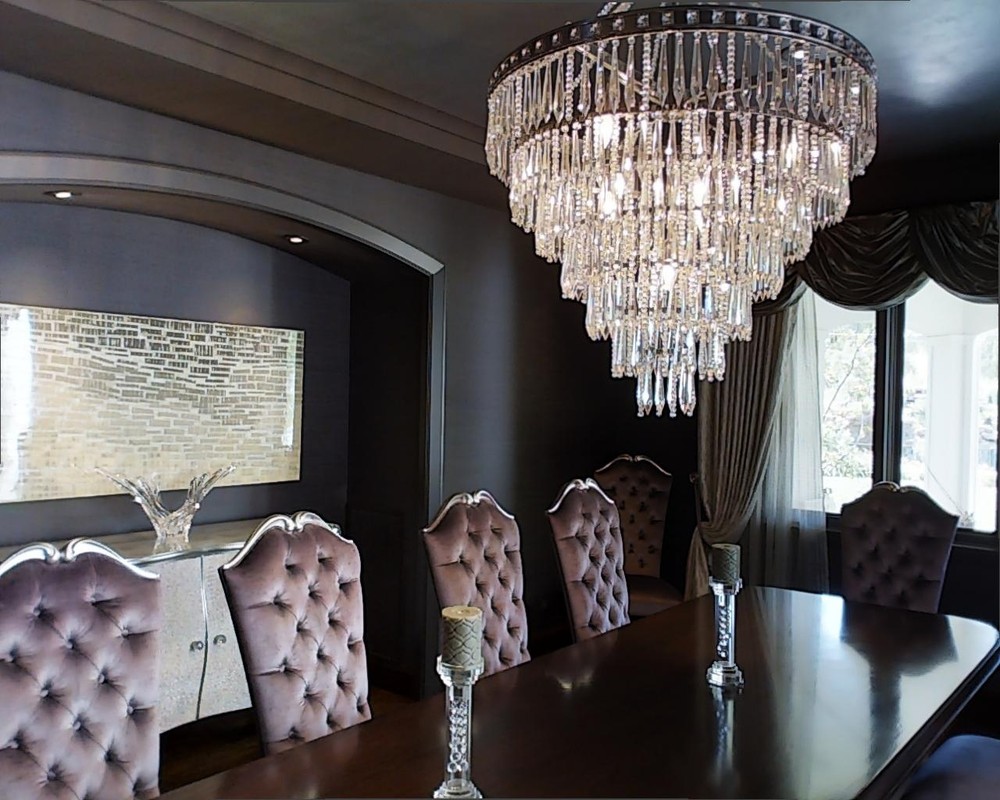} &
\includegraphics[height=0.099\textheight]{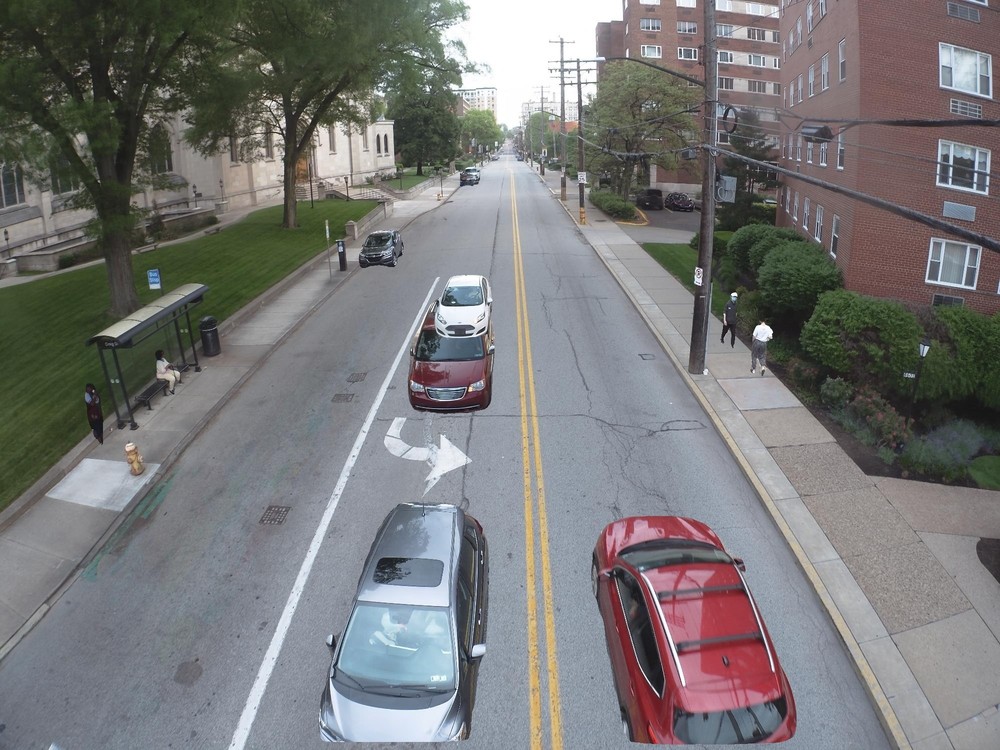} & \includegraphics[height=0.099\textheight]{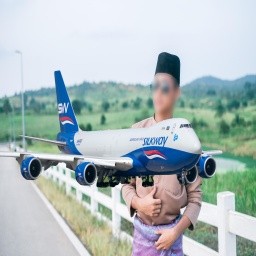} &
\includegraphics[height=0.099\textheight]{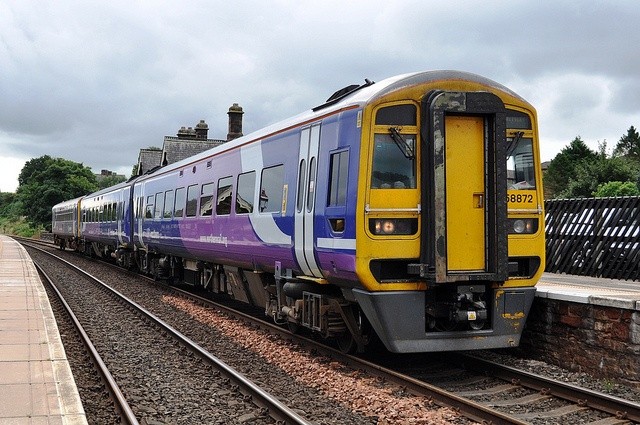} &
\includegraphics[height=0.099\textheight]{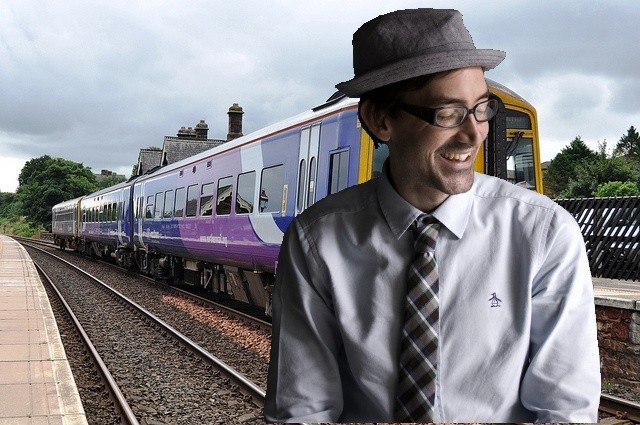}\\
\rotatebox{90}{\makebox[0.099\textheight][c]{\small Modal mask}} &
\includegraphics[height=0.099\textheight]{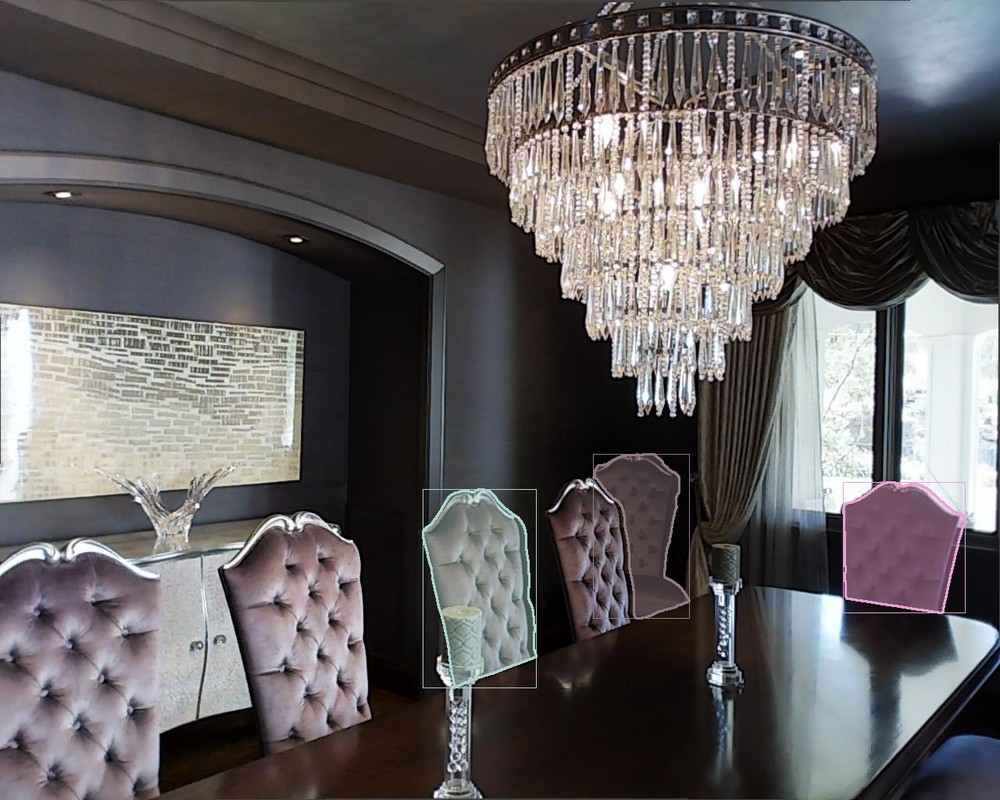} &
\includegraphics[height=0.099\textheight]{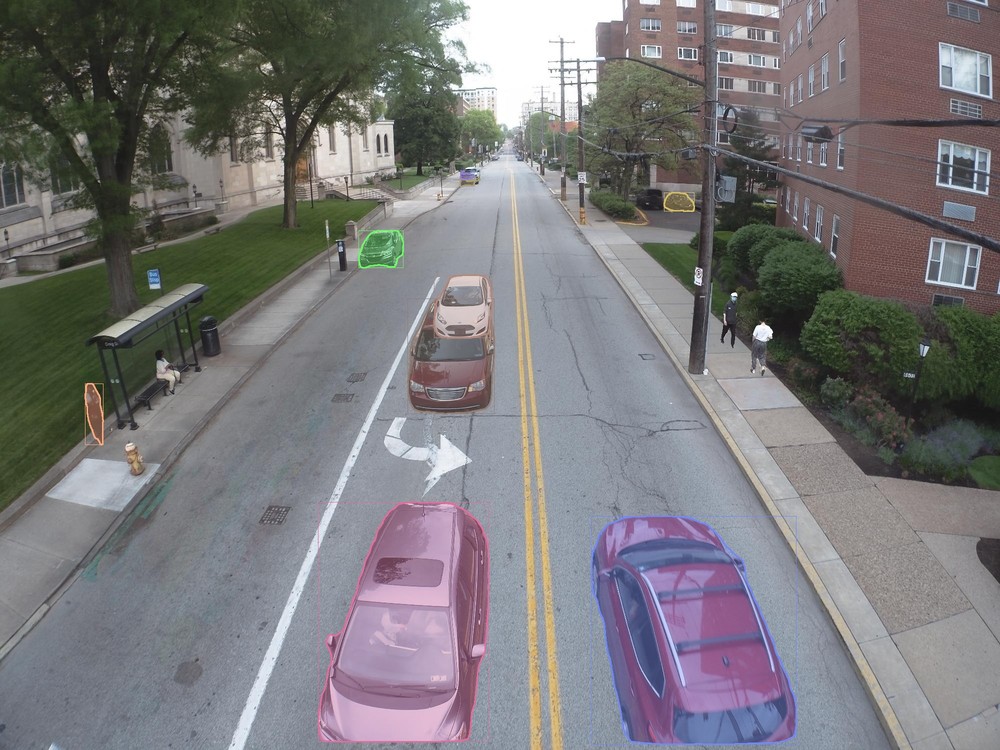} &
\includegraphics[height=0.099\textheight]{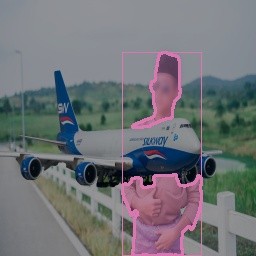} &
\includegraphics[height=0.099\textheight]{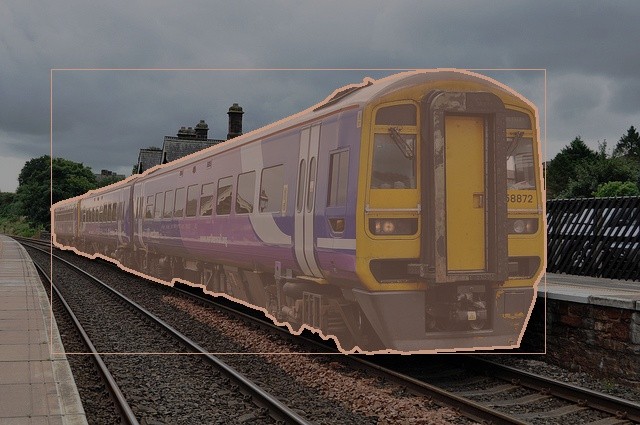} &
\includegraphics[height=0.099\textheight]{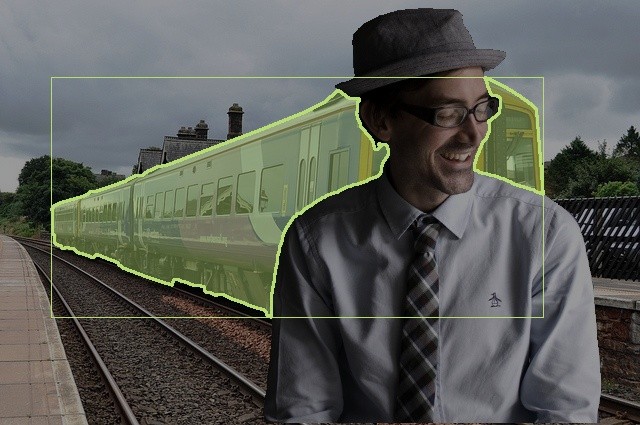}\\
\rotatebox{90}{\makebox[0.099\textheight][c]{\small Amodal mask}} &
\includegraphics[height=0.099\textheight]{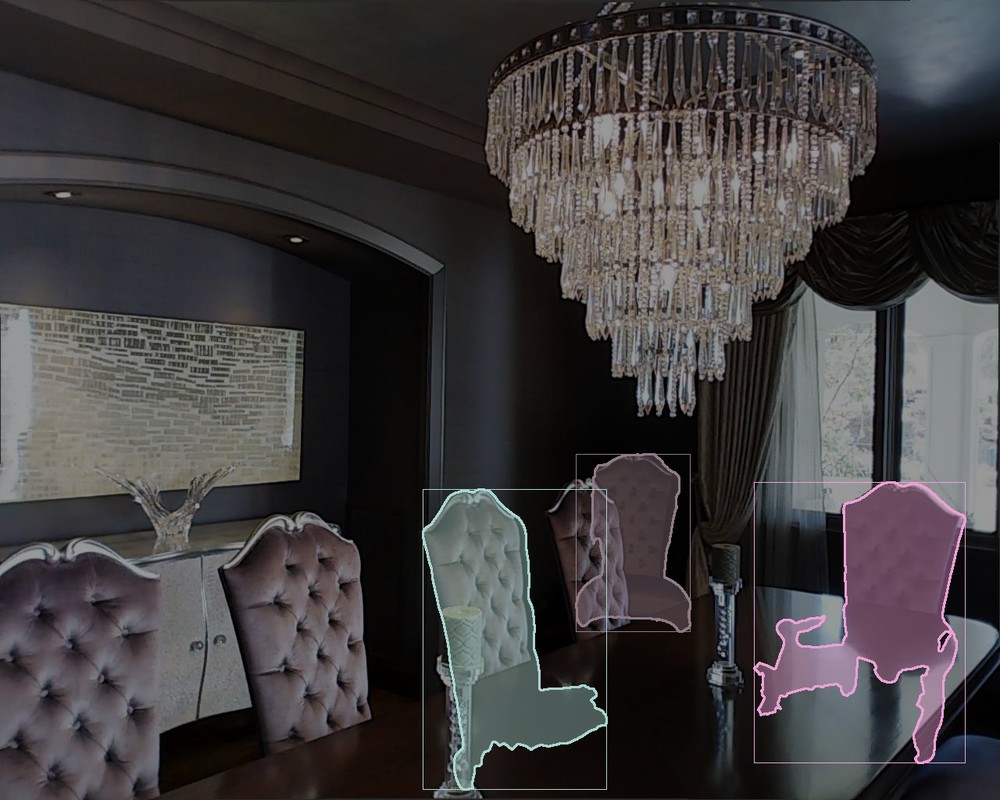} &
\includegraphics[height=0.099\textheight]{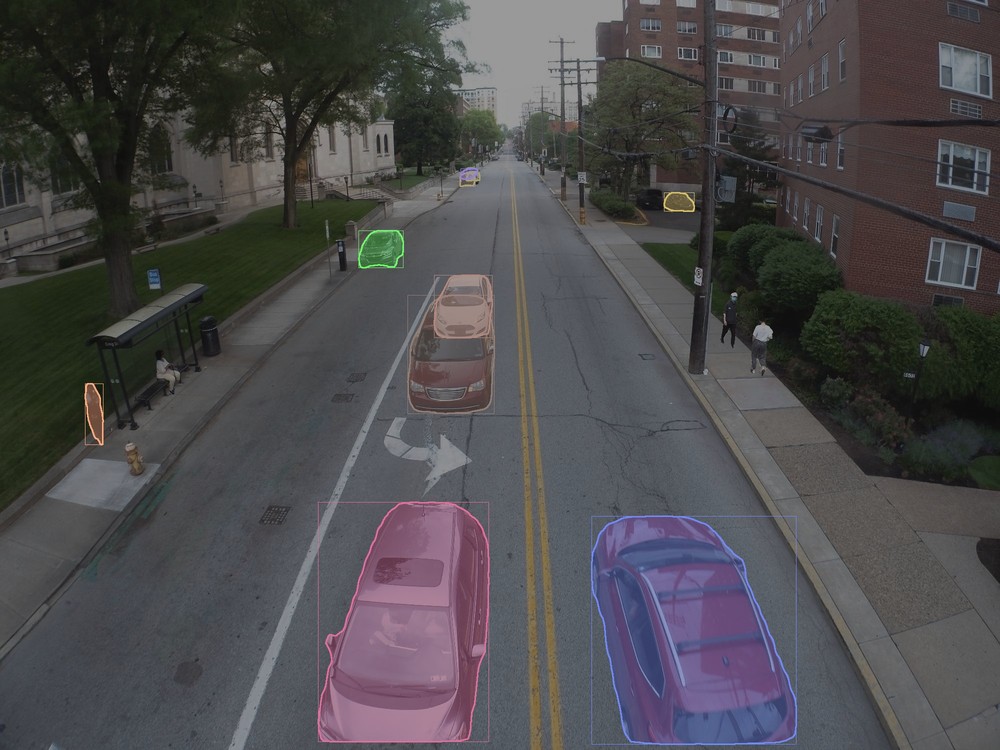} & \includegraphics[height=0.099\textheight]{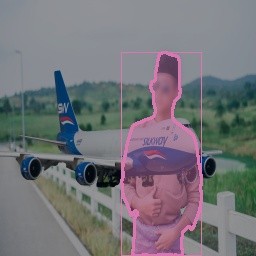} &
\includegraphics[height=0.099\textheight]{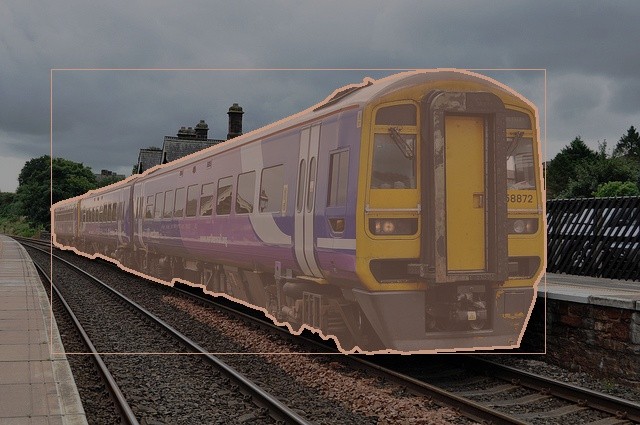} &
\includegraphics[height=0.099\textheight]{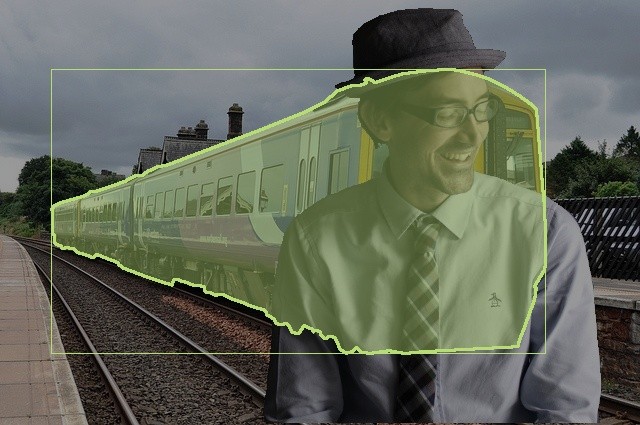}\\
\end{tabular}

\caption{Visualization of collected amodal datasets. For Amodal-LVIS, each instance has unoccluded (left) and occluded (right) versions.}
\label{fig:datavis}
\end{figure*}

Our collected amodal datasets, shown in \Cref{fig:datavis}, serve as essential training data for zero-shot amodal instance segmentation. Across ten diverse examples (COCOA~\cite{ZhuTMD17}, COCOA-cls~\cite{FollmannKHKB19}, DYCE~\cite{EhsaniMF18}, KINS~\cite{QiJ0SJ19}, MUVA~\cite{LiYTBZJ023}, D2SA~\cite{FollmannKHKB19}, KITTI-360-APS~\cite{MohanV22}, MP3D-amodal~\cite{ZhanZXZ24}, WALT~\cite{ReddyTN22}, and pix2gestalt~\cite{OzgurogluLS0DTV24}), we display modal and amodal mask pairs. Our proposed Amodal-LVIS dataset features dual annotations of both occluded and unoccluded versions of each instance. This curated collection provides rich training signals that enable our model to learn generalizable amodal segmentation capabilities across different domains and object categories.

\section{Qualitative Comparison}
\label{sec:comp_ais}
\begin{figure*}[t]
    \centering
    \begin{minipage}{0.02\textwidth}
        \vspace{0.5cm}
        \rotatebox{90}{\textbf{Input}} \\[0.8cm]
        \rotatebox{90}{\textbf{SAMEO}} \\[0.8cm]
        \rotatebox{90}{\textbf{AISFormer}}
    \end{minipage}%
    \begin{minipage}{0.98\textwidth}
        \centering
        \resizebox{!}{6.5cm}{\includegraphics{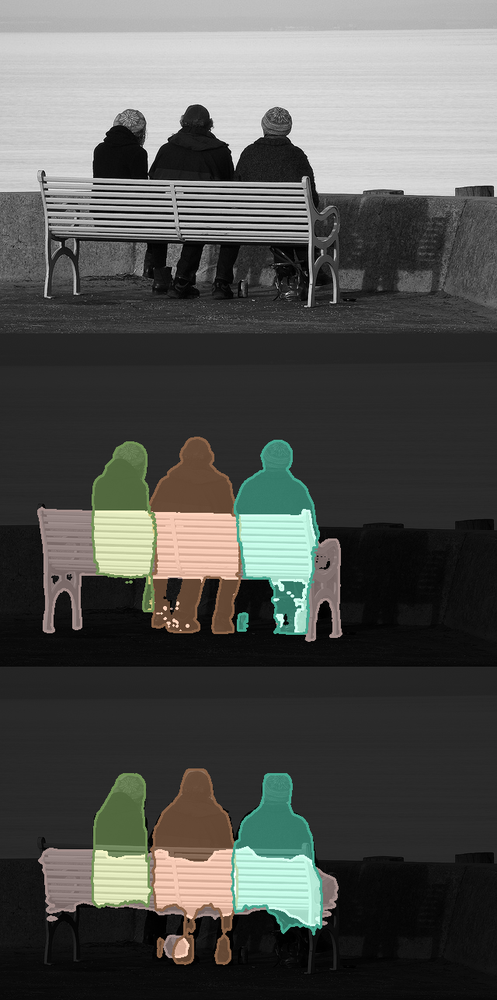}}
        \resizebox{!}{6.5cm}{\includegraphics{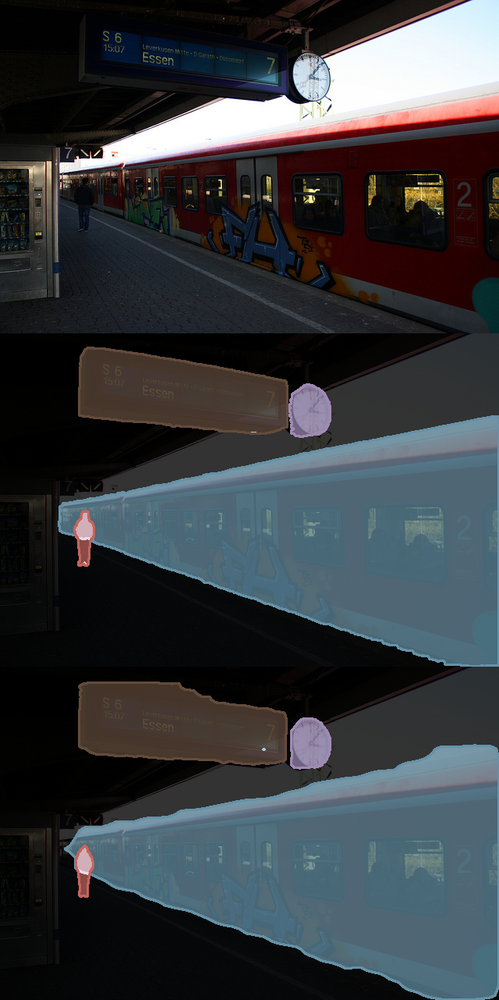}}
        \resizebox{!}{6.5cm}{\includegraphics{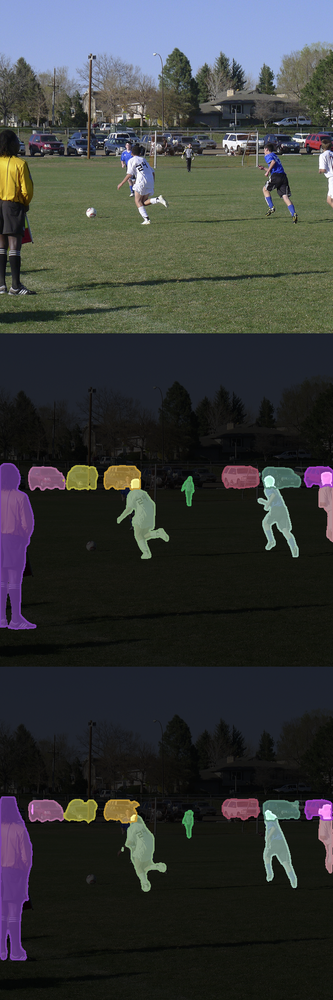}}
        \resizebox{!}{6.5cm}{\includegraphics{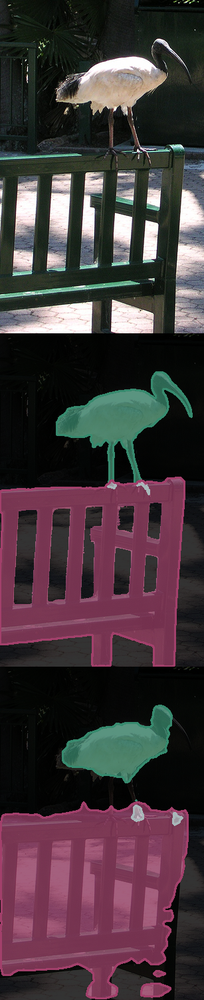}}
        \resizebox{!}{6.5cm}{\includegraphics{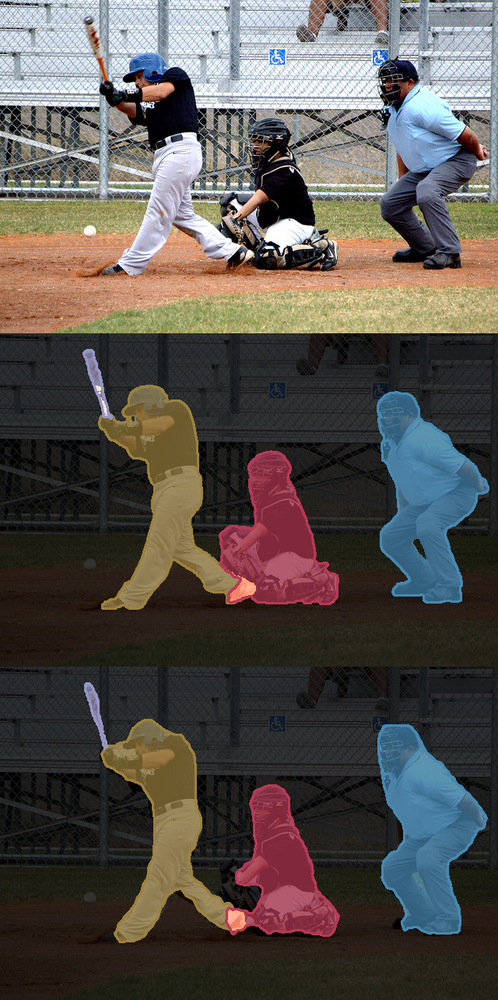}}
        \resizebox{!}{6.5cm}{\includegraphics{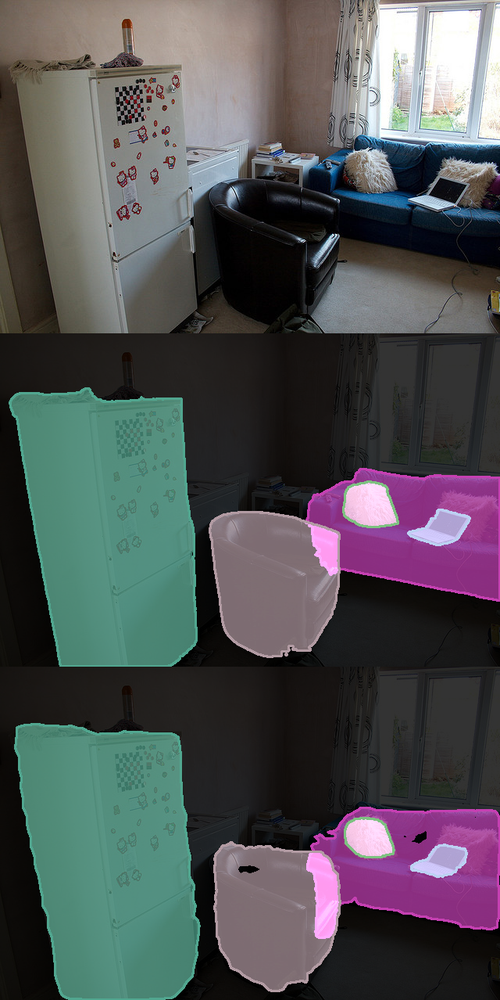}}
    \end{minipage}
    
    \vspace{1pt}
    \begin{minipage}{0.02\textwidth}
        \vspace{0.5cm}
        \rotatebox{90}{\textbf{Input}} \\[0.8cm]
        \rotatebox{90}{\textbf{SAMEO}} \\[0.6cm]
        \rotatebox{90}{\textbf{AISFormer}}
    \end{minipage}%
    \begin{minipage}{0.98\textwidth}
        \centering
        \resizebox{!}{6cm}{\includegraphics{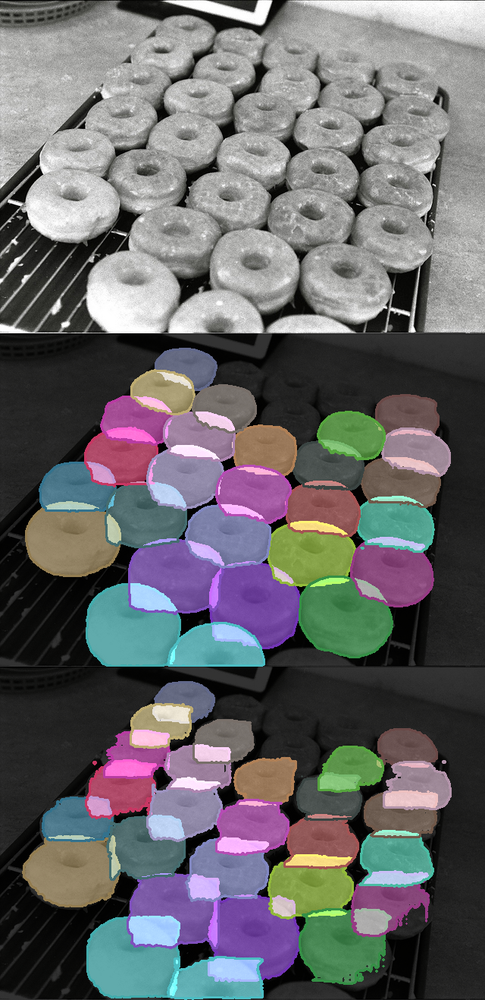}}
        \resizebox{!}{6cm}{\includegraphics{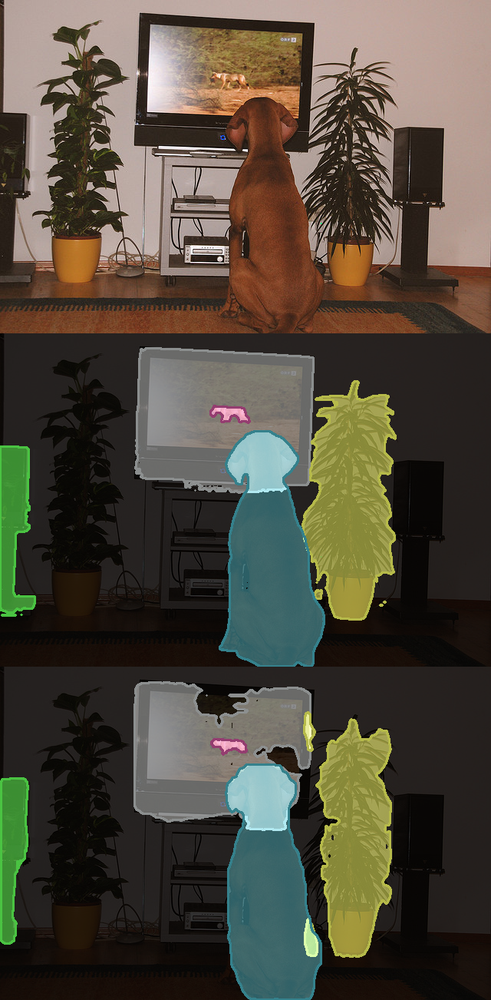}}
        \resizebox{!}{6cm}{\includegraphics{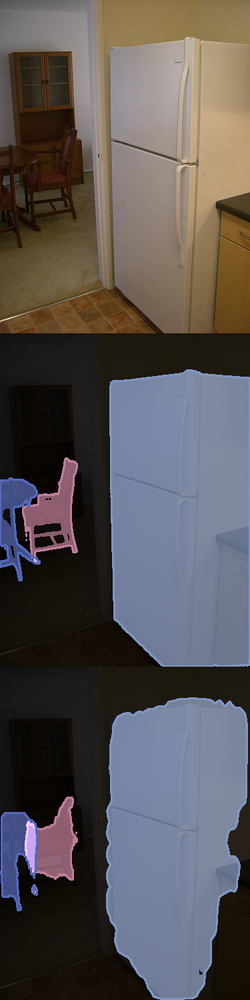}}
        \resizebox{!}{6cm}{\includegraphics{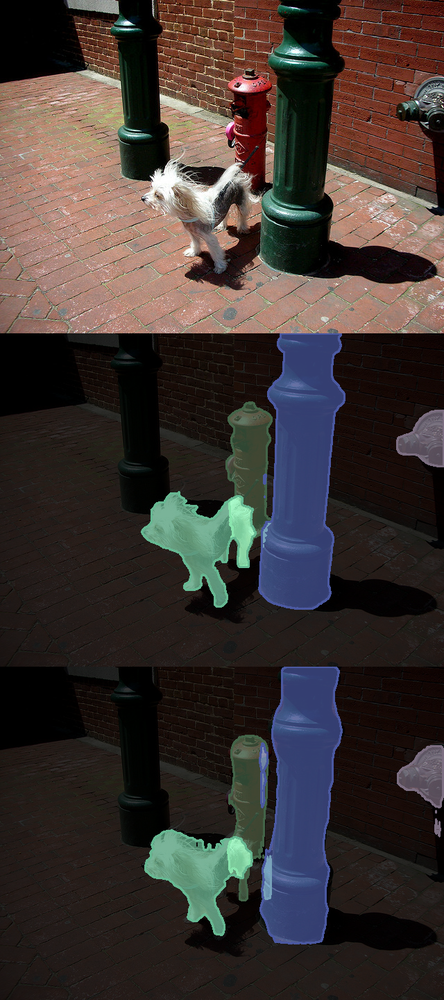}}
        \resizebox{!}{6cm}{\includegraphics{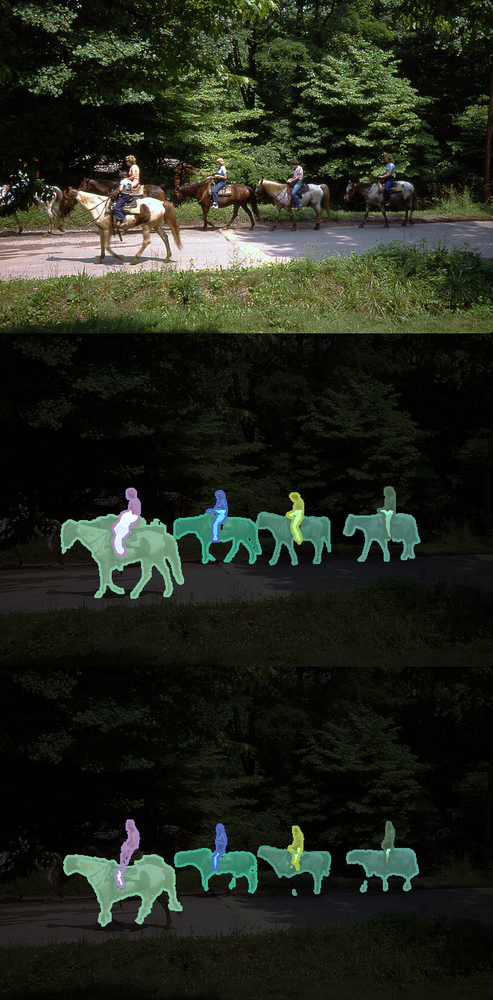}}
        \resizebox{!}{6cm}{\includegraphics{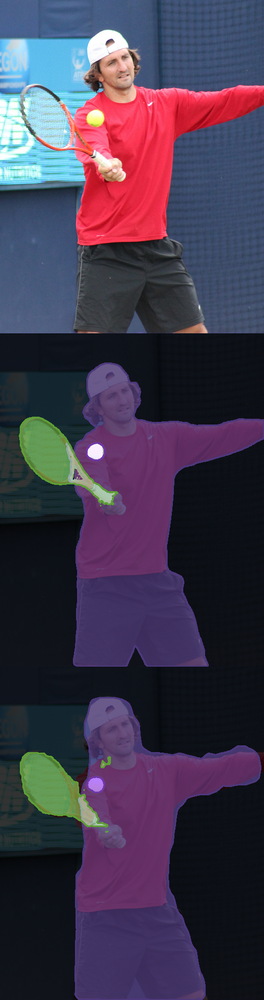}}
        \resizebox{!}{6cm}{\includegraphics{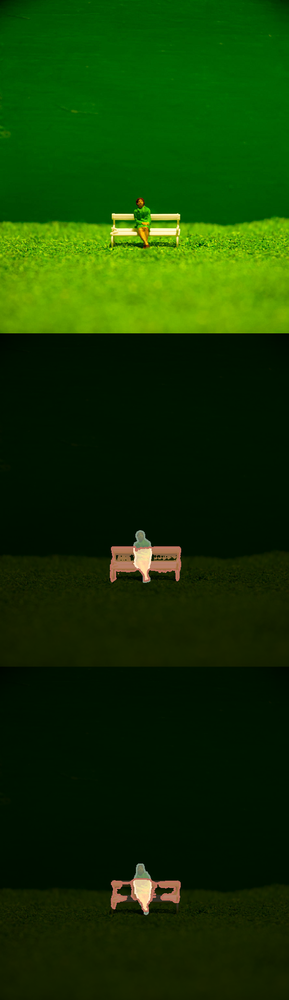}}
    \end{minipage}
    
    \vspace{1pt}
    \begin{minipage}{0.02\textwidth}
        \vspace{0.5cm}
        \rotatebox{90}{\textbf{Input}} \\[0.8cm]
        \rotatebox{90}{\textbf{SAMEO}} \\[0.7cm]
        \rotatebox{90}{\textbf{AISFormer}}
    \end{minipage}%
    \begin{minipage}{0.98\textwidth}
        \centering
        \resizebox{!}{6.51cm}{\includegraphics{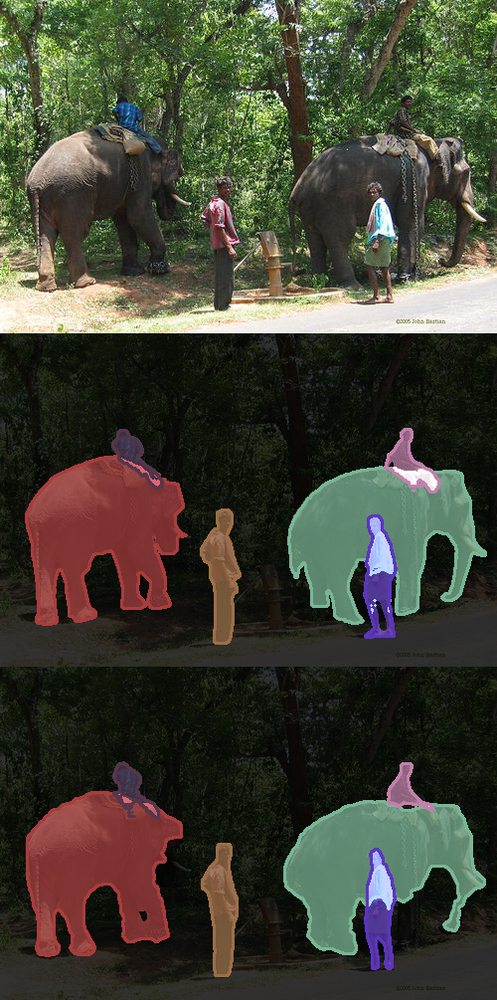}}
        \resizebox{!}{6.51cm}{\includegraphics{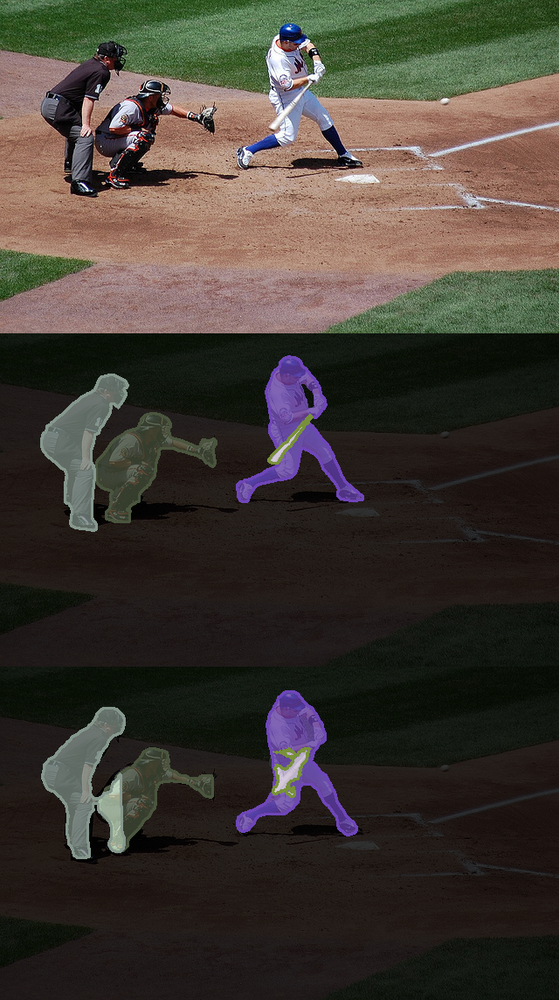}}
        \resizebox{!}{6.51cm}{\includegraphics{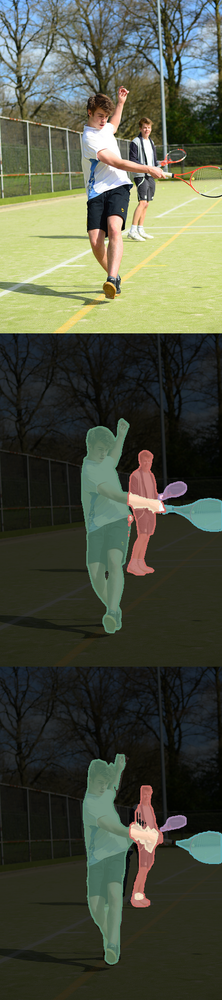}}
        \resizebox{!}{6.51cm}{\includegraphics{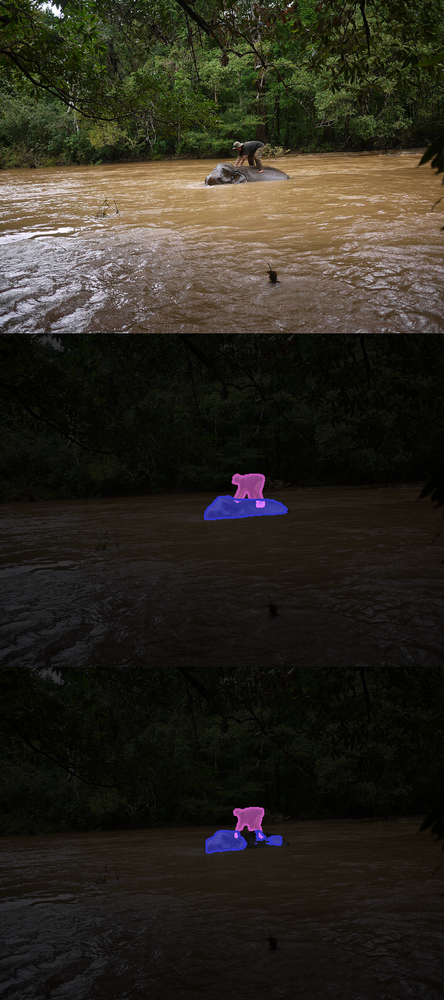}}
        \resizebox{!}{6.51cm}{\includegraphics{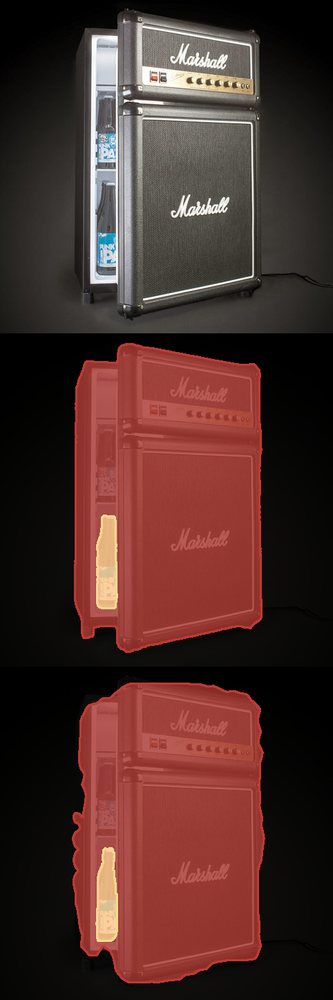}}
        \resizebox{!}{6.51cm}{\includegraphics{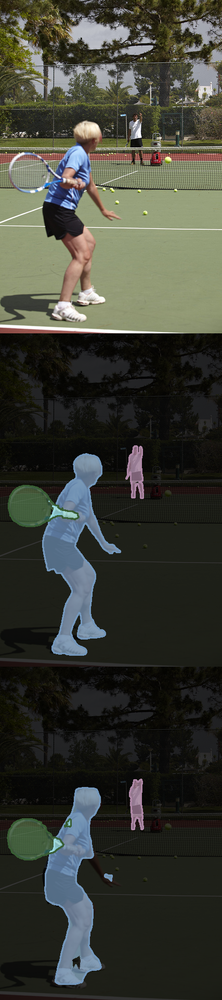}}
        \resizebox{!}{6.51cm}{\includegraphics{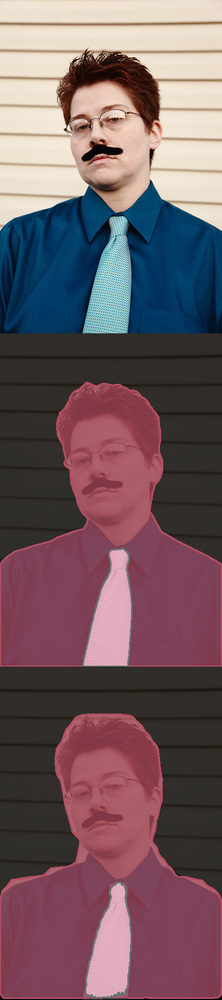}}
    \end{minipage}
    \caption{Qualitative comparison of amodal instance segmentation on COCOA-cls dataset. Each row shows: \emph{i})~input RGB image, \emph{ii})~SAMEO's amodal prediction using AISFormer boxes as prompts, and \emph{iii})~AISFormer's prediction. SAMEO demonstrates superior mask boundary delineation and more accurate occluded region estimation compared to the baseline.}
    \label{fig:supp_cocoa_vis}
\end{figure*}

\begin{figure*}[t]
    \centering
    \begin{minipage}{0.02\textwidth}
        \vspace{0.5cm}
        \rotatebox{90}{\textbf{Input}} \\[0.6cm]
        \rotatebox{90}{\textbf{SAMEO}} \\[0.4cm]
        \rotatebox{90}{\textbf{AISFormer}}
    \end{minipage}%
    \begin{minipage}{0.98\textwidth}
        \centering
        \resizebox{!}{5.6cm}{\includegraphics{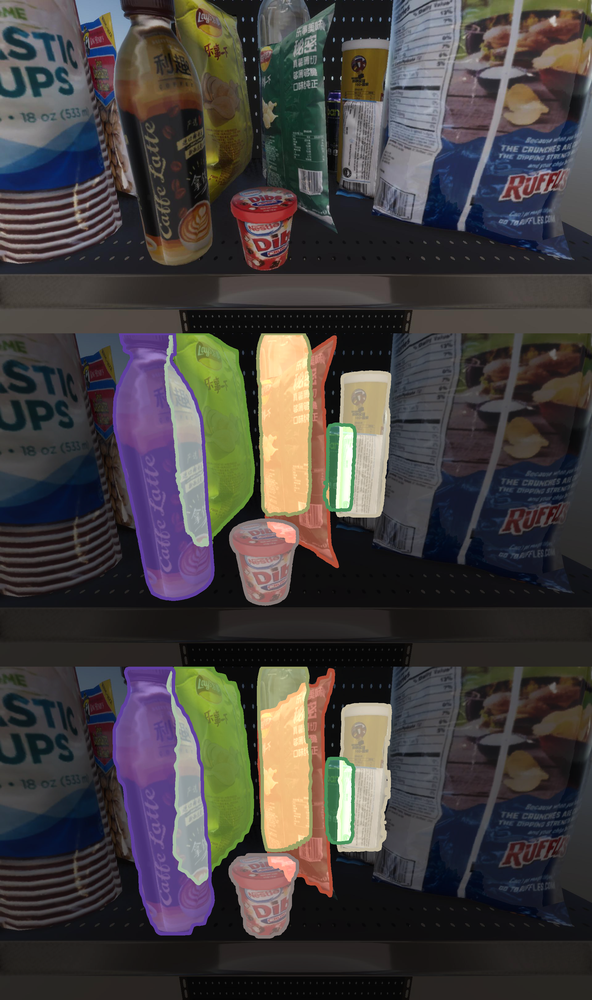}}
        \resizebox{!}{5.6cm}{\includegraphics{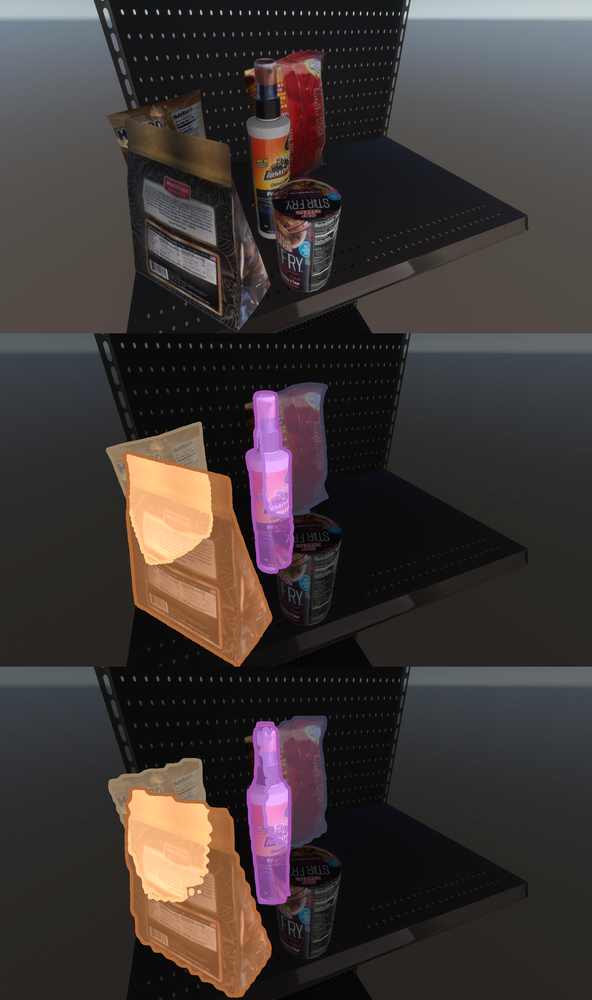}}
        \resizebox{!}{5.6cm}{\includegraphics{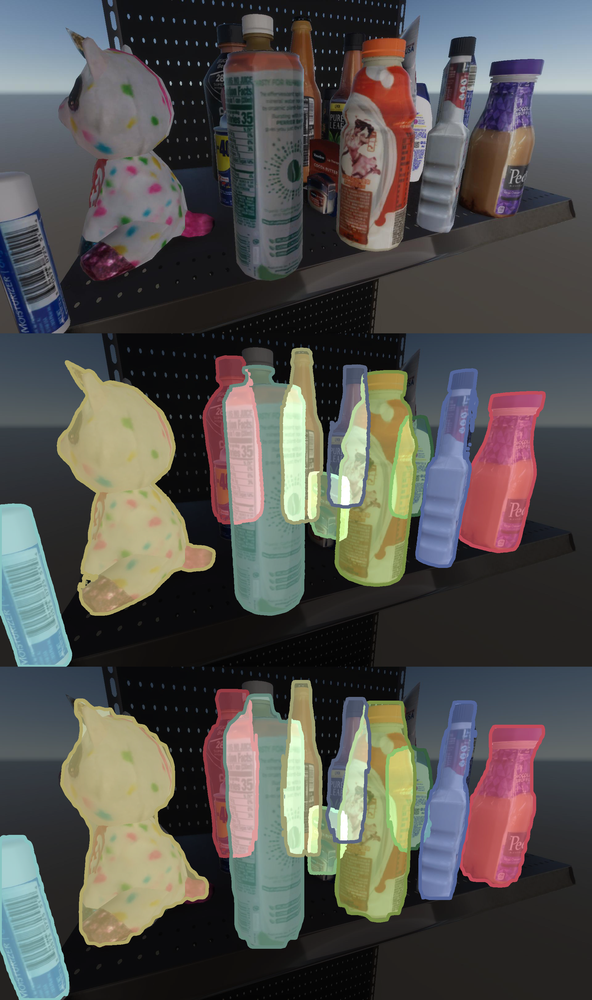}}
        \resizebox{!}{5.6cm}{\includegraphics{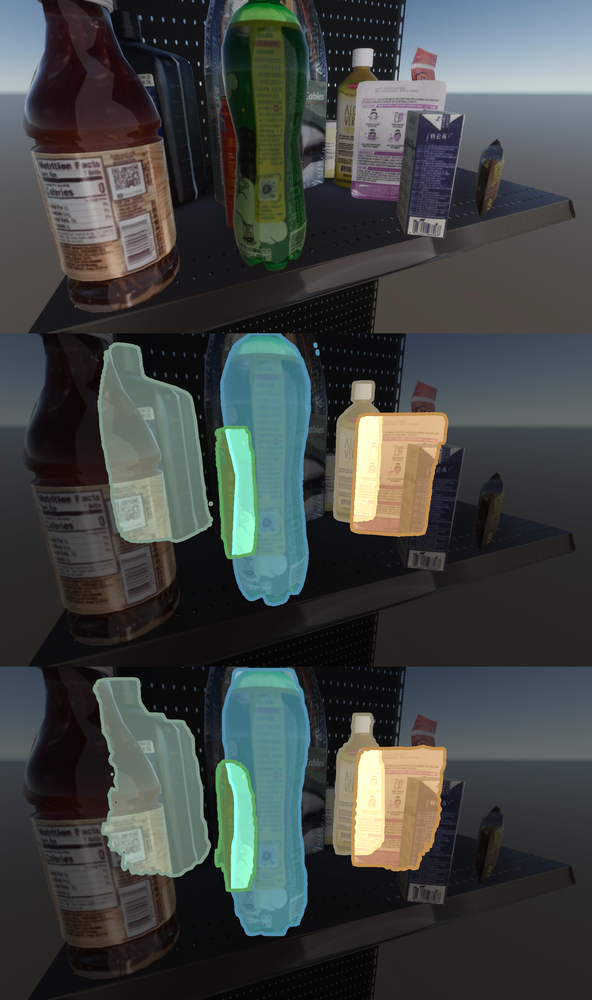}}
        \resizebox{!}{5.6cm}{\includegraphics{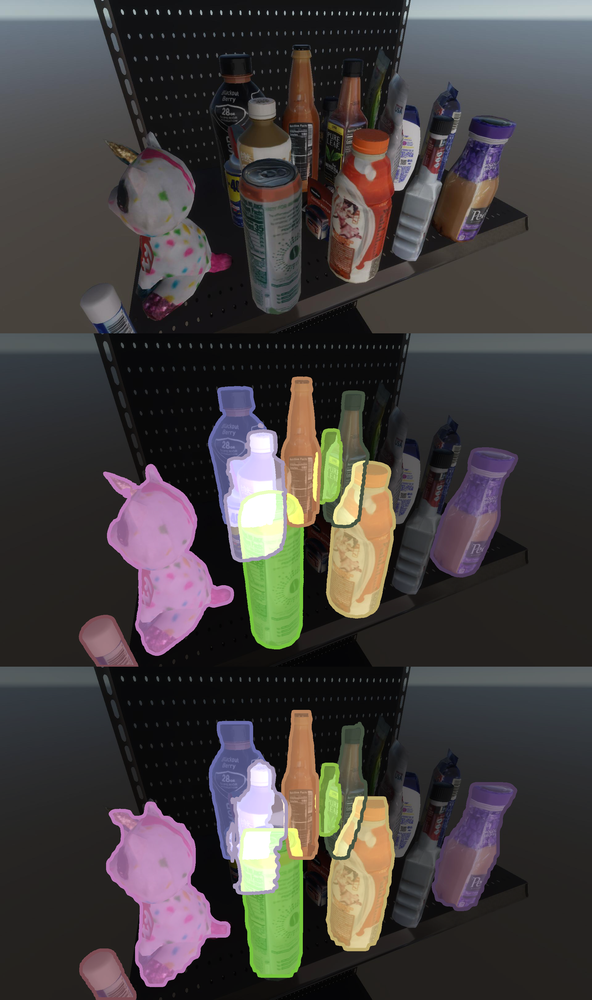}}
    \end{minipage}
    \vspace{1pt}
    \begin{minipage}{0.02\textwidth}
        \vspace{0.5cm}
        \rotatebox{90}{\textbf{Input}} \\[0.6cm]
        \rotatebox{90}{\textbf{SAMEO}} \\[0.4cm]
        \rotatebox{90}{\textbf{AISFormer}}
    \end{minipage}%
    \begin{minipage}{0.98\textwidth}
        \centering
        \resizebox{!}{5.6cm}{\includegraphics{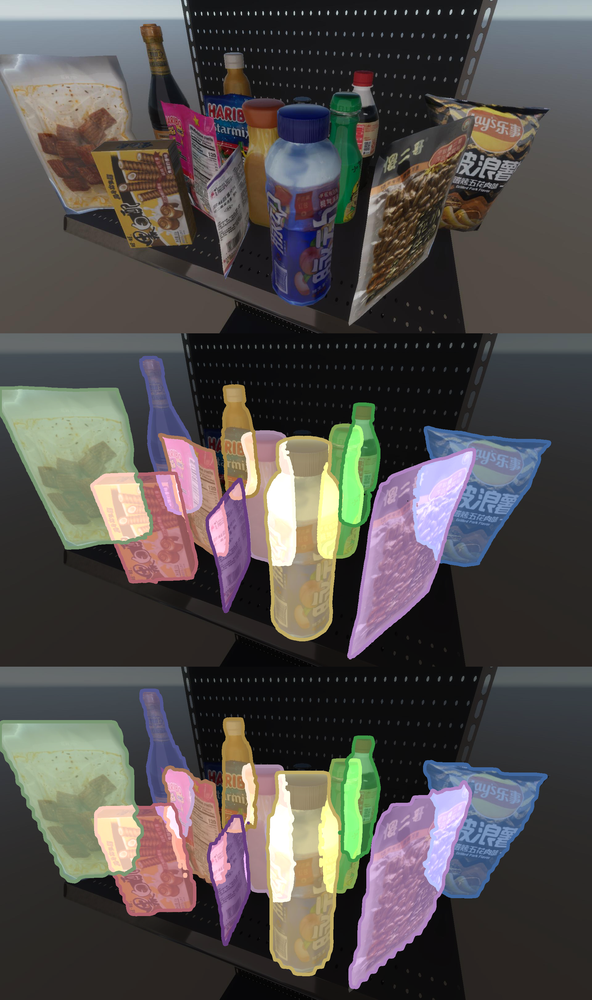}}
        \resizebox{!}{5.6cm}{\includegraphics{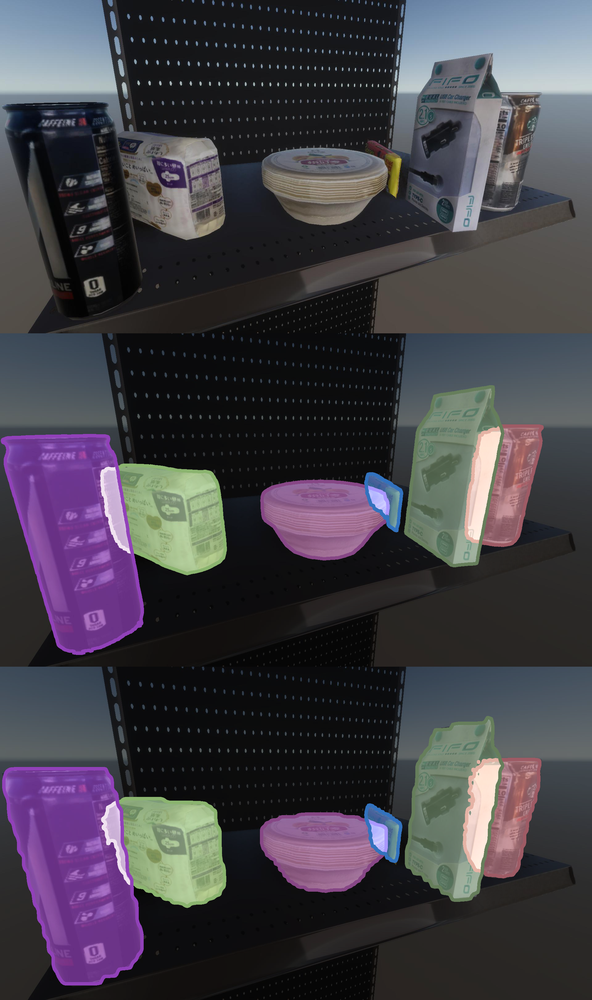}}
        \resizebox{!}{5.6cm}{\includegraphics{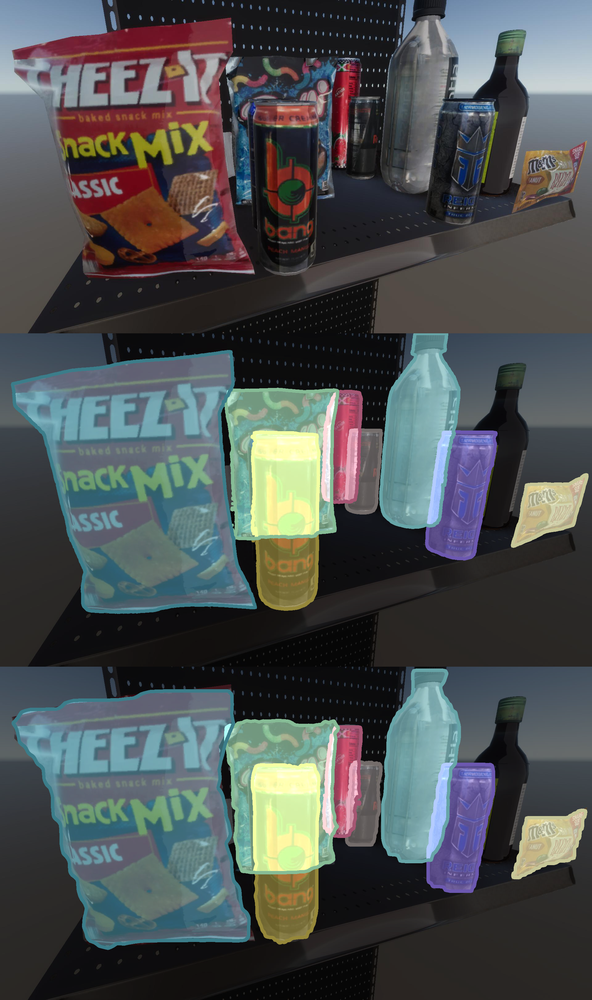}}
        \resizebox{!}{5.6cm}{\includegraphics{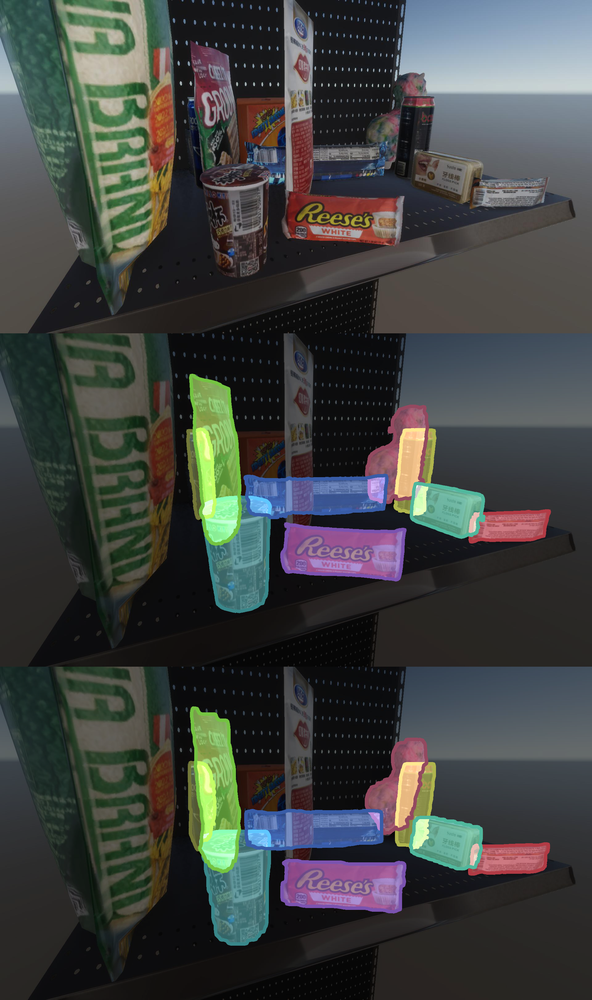}}
        \resizebox{!}{5.6cm}{\includegraphics{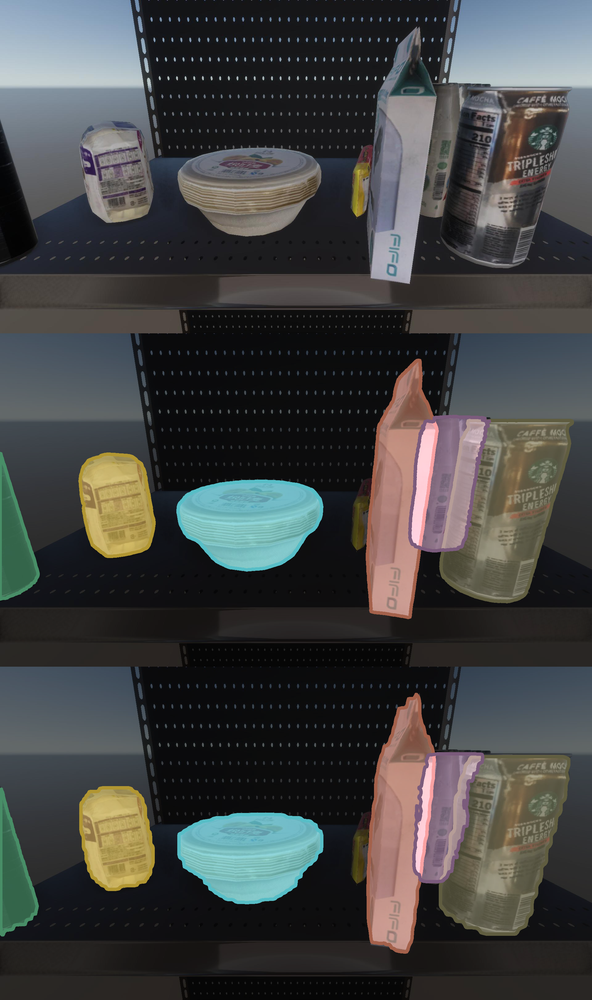}}
    \end{minipage}
    \vspace{1pt}
    \begin{minipage}{0.02\textwidth}
        \vspace{0.5cm}
        \rotatebox{90}{\textbf{Input}} \\[0.6cm]
        \rotatebox{90}{\textbf{SAMEO}} \\[0.4cm]
        \rotatebox{90}{\textbf{AISFormer}}
    \end{minipage}%
    \begin{minipage}{0.98\textwidth}
        \centering
        \resizebox{!}{5.6cm}{\includegraphics{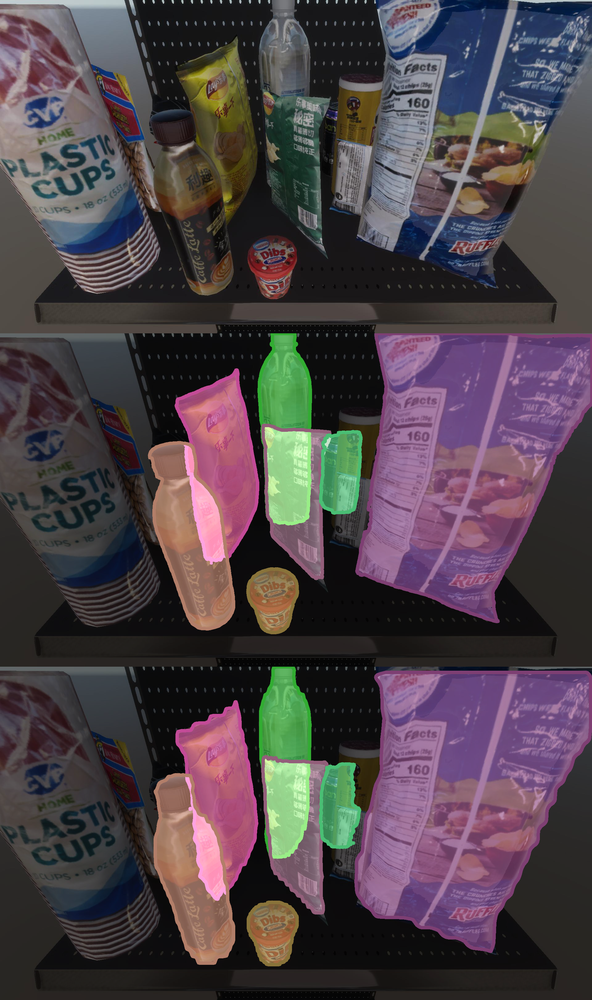}}
        \resizebox{!}{5.6cm}{\includegraphics{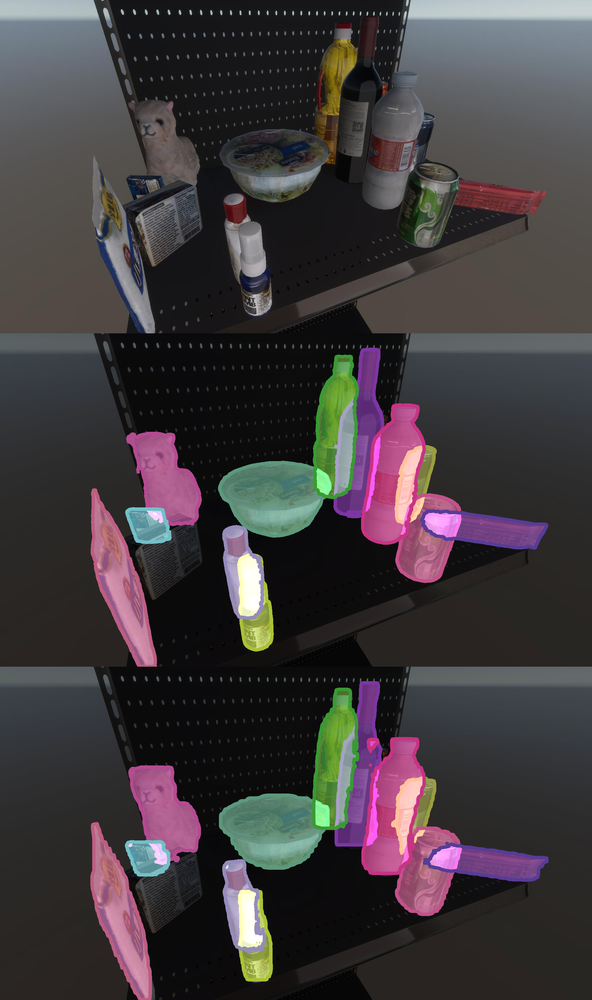}}
        \resizebox{!}{5.6cm}{\includegraphics{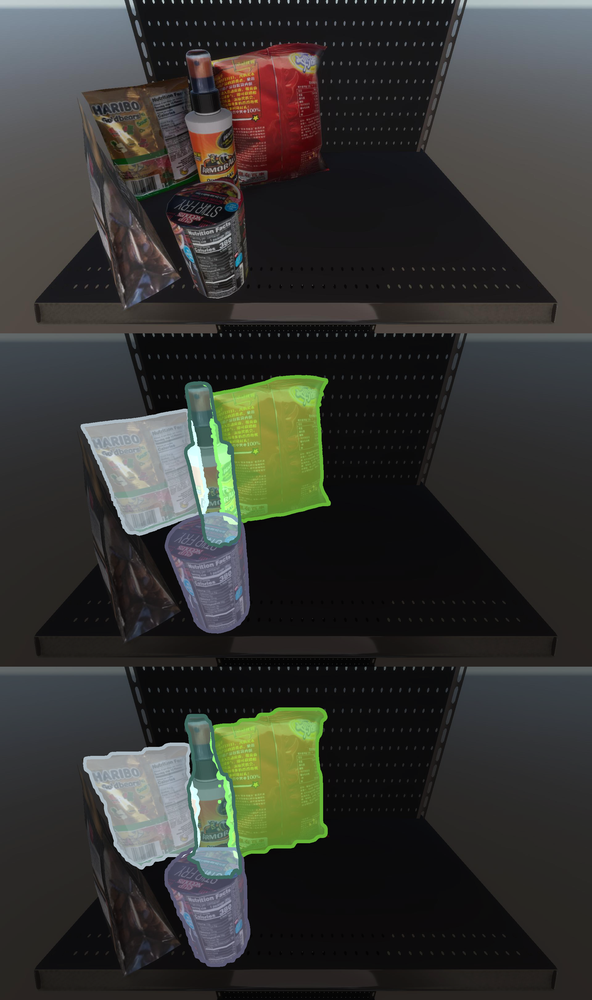}}
        \resizebox{!}{5.6cm}{\includegraphics{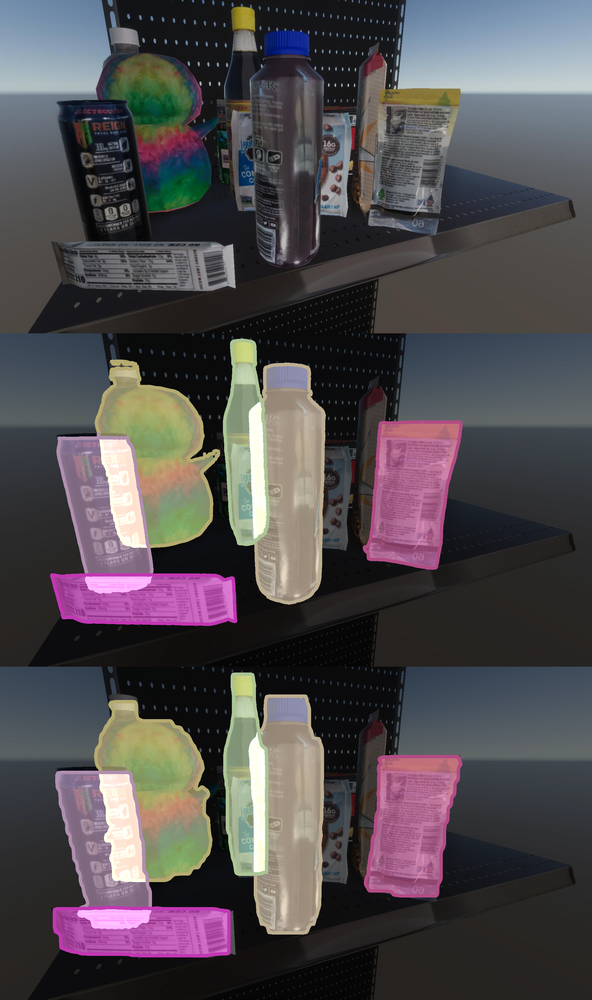}}
        \resizebox{!}{5.6cm}{\includegraphics{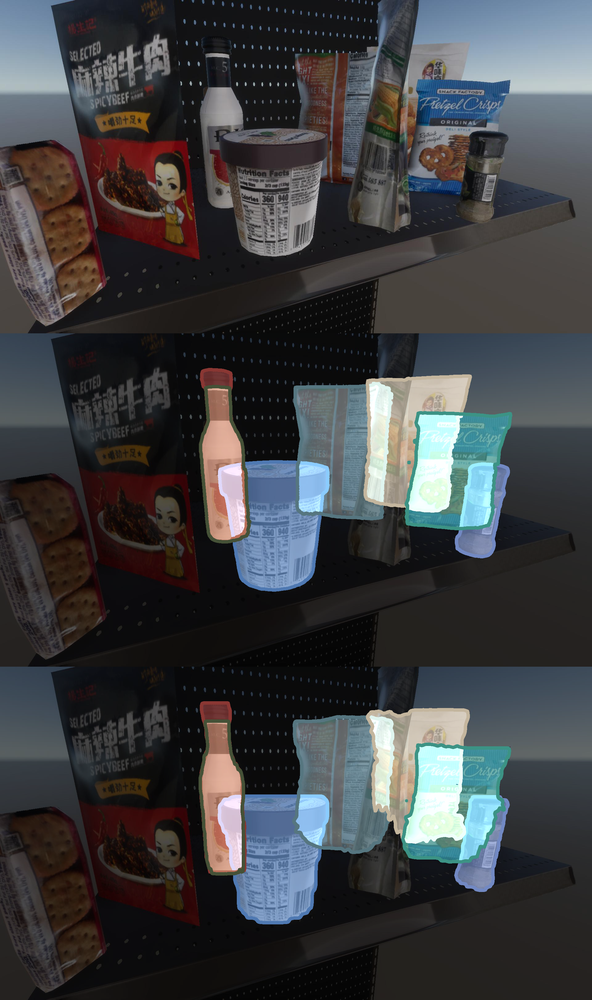}}
    \end{minipage}
    
    \caption{Qualitative comparison of amodal instance segmentation on MUVA dataset. Each row displays: \emph{i})~input RGB image, \emph{ii})~amodal masks predicted by SAMEO with AISFormer box prompts, and \emph{iii})~AISFormer predictions. Our approach yields more precise boundaries and better handles occlusion estimation.}
    \label{fig:supp_muva_vis}
\end{figure*}

We compare SAMEO's amodal instance segmentation capabilities with state-of-the-art AISFormer on COCOA-cls (\Cref{fig:supp_cocoa_vis}) and MUVA (\Cref{fig:supp_muva_vis}) datasets. Using AISFormer's box predictions as prompts, SAMEO generates amodal masks for detected instances. The qualitative results demonstrate SAMEO's superior performance in mask boundary precision and occlusion estimation, particularly for complex shapes and instances with multiple overlaps. Our method significantly outperforms AISFormer in terms of overall mask quality.
\section{Amodal Mask Adaptation}
\label{sec:comp_effsam}
\begin{figure*}
    \centering
    \rotatebox{90}{\makebox[1ex]{\textbf{\hspace{8cm} GT \hspace{0.9cm} SAMEO \hspace{0.2cm} EfficientSAM \hspace{0.5cm} Input}}}
    \includegraphics[width=0.115\textwidth]{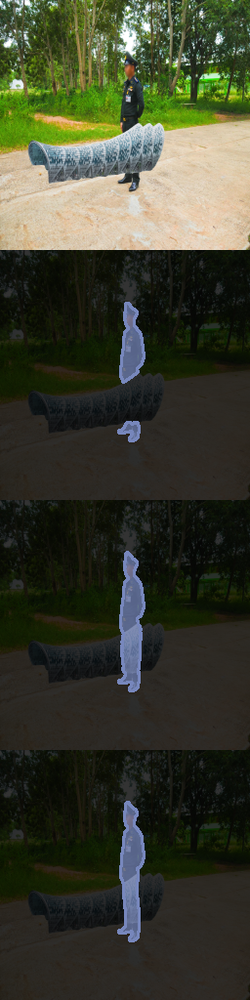}
    \includegraphics[width=0.115\textwidth]{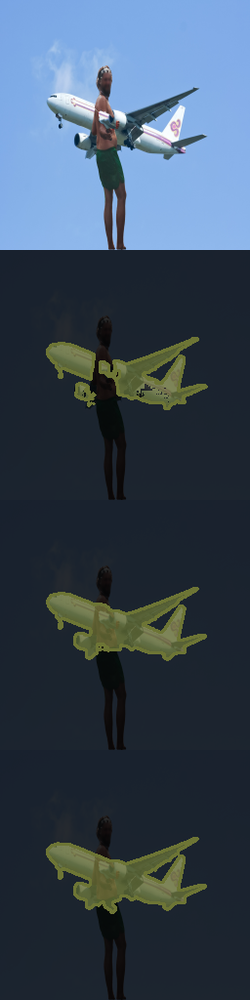}
    \includegraphics[width=0.115\textwidth]{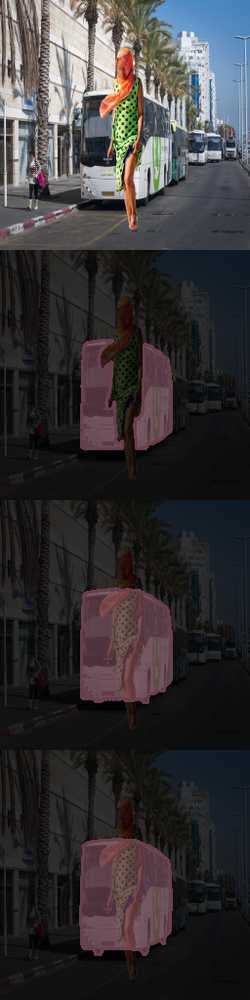}
    \includegraphics[width=0.115\textwidth]{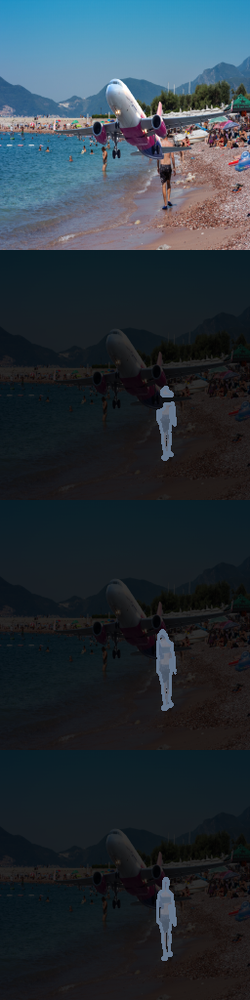}
    \includegraphics[width=0.115\textwidth]{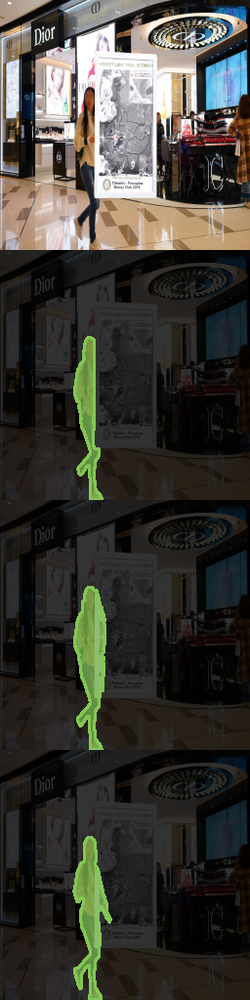}
    \includegraphics[width=0.115\textwidth]{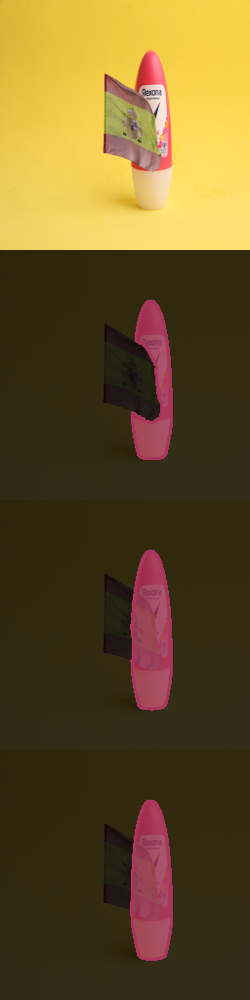}
    \includegraphics[width=0.115\textwidth]{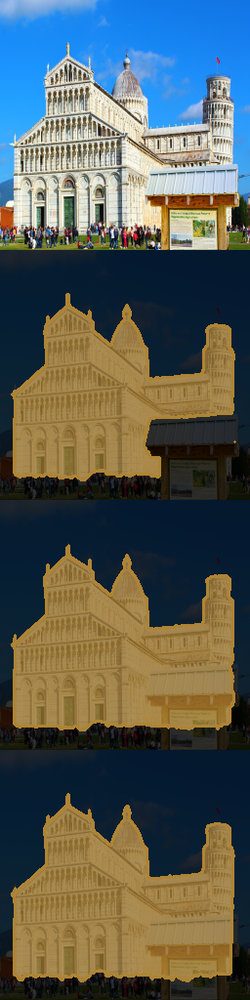}
    \includegraphics[width=0.115\textwidth]{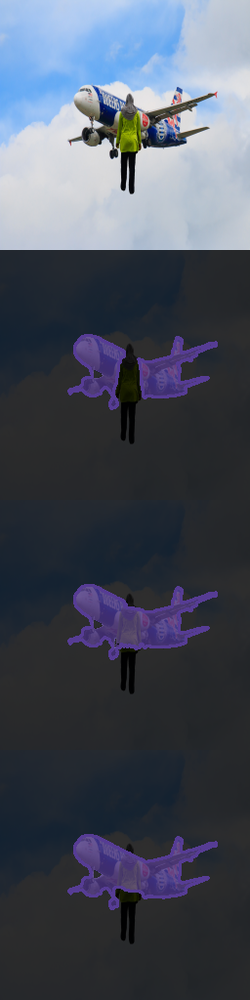}

    \rotatebox{90}{\makebox[1ex]{\textbf{\hspace{8cm} GT \hspace{0.8cm} SAMEO \hspace{0.2cm} EfficientSAM \hspace{0.5cm} Input}}}
    \includegraphics[width=0.115\textwidth]{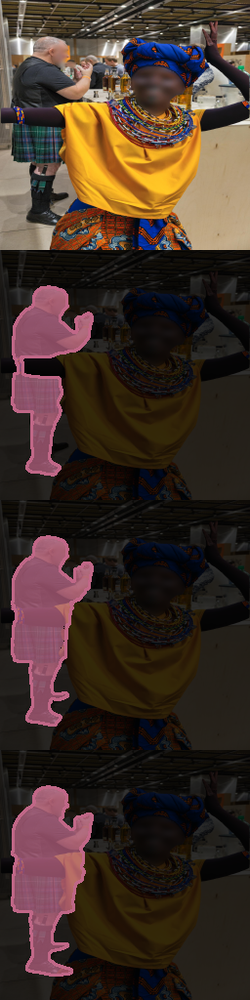}
    \includegraphics[width=0.115\textwidth]{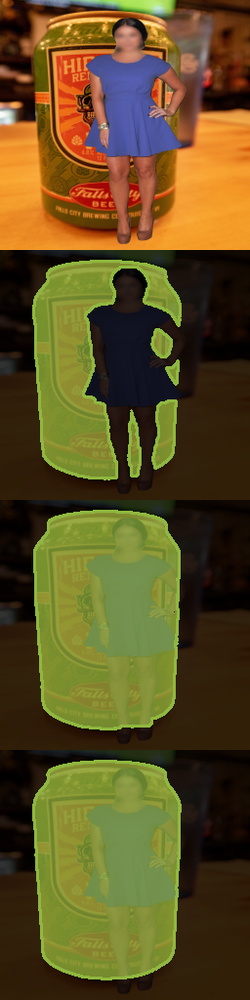}
    \includegraphics[width=0.115\textwidth]{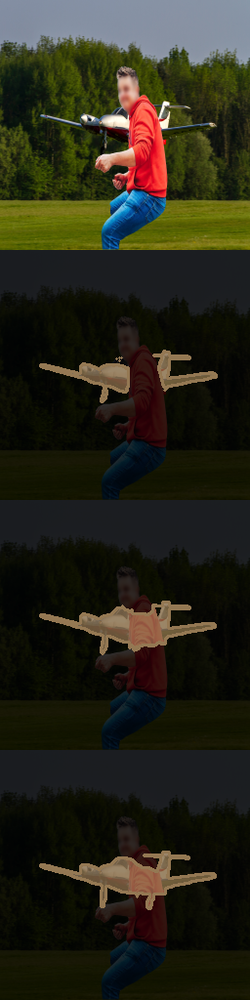}
    \includegraphics[width=0.115\textwidth]{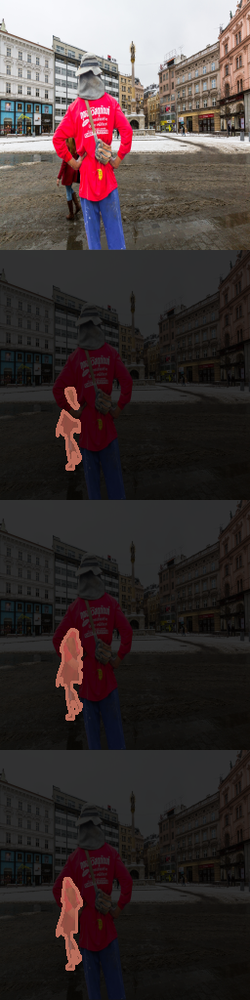}
    \includegraphics[width=0.115\textwidth]{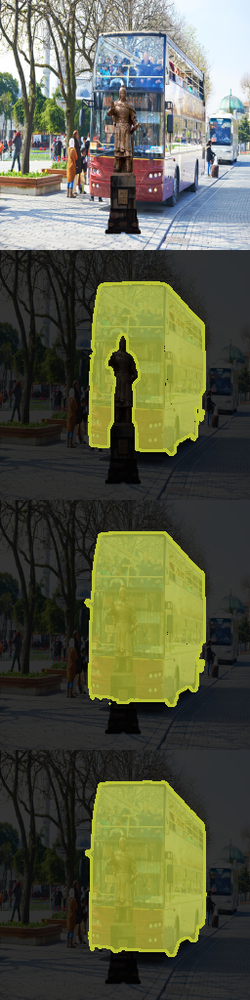}
    \includegraphics[width=0.115\textwidth]{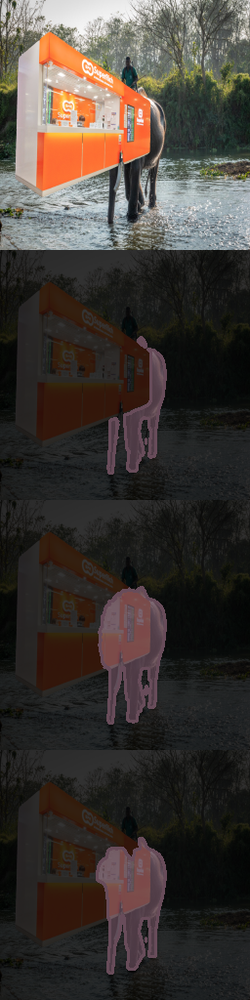}
    \includegraphics[width=0.115\textwidth]{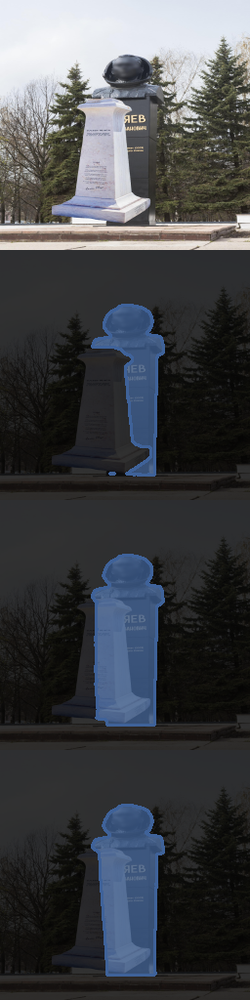}
    \includegraphics[width=0.115\textwidth]{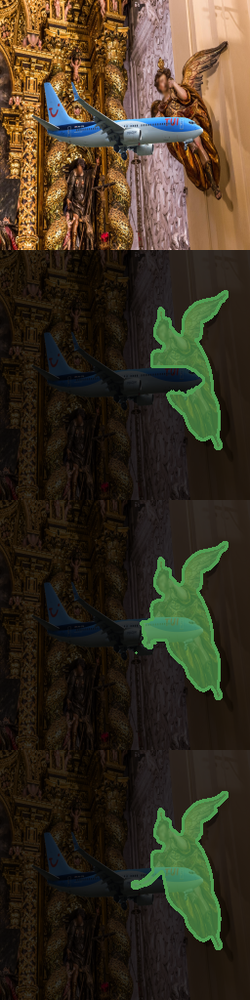}
    \caption{From modal to amodal segmentation on pix2gestalt dataset. Each row demonstrates: \emph{i})~input RGB image, \emph{ii})~modal mask prediction from the original EfficientSAM, \emph{iii})~amodal mask prediction from our SAMEO, \emph{iv})~ground truth amodal mask. The results showcase SAMEO's successful adaptation to amodal segmentation while maintaining zero-shot capabilities.}
    \label{fig:supp_pix_vis}
\end{figure*}

We demonstrate SAMEO's adaptation from modal to amodal segmentation through visualization experiments on the pix2gestalt dataset (\Cref{fig:supp_pix_vis}). Comparing the modal mask predictions from the original EfficientSAM with SAMEO's amodal predictions and ground truth masks reveals successful adaptation to amodal segmentation. Our specialized training enables SAMEO to effectively estimate occluded regions while preserving the high-quality mask prediction and zero-shot capabilities inherent to the original model.
\section{Class-specific Results}
\label{sec:ap_with_cls}
\Cref{tab:clsnonzeroshot} and \Cref{tab:clszeroshot} present the class-specific AP/AR evaluations as a complement to class-agnostic results, following identical experimental settings from \Cref{sec:experiments}. In both standard and zero-shot settings, SAMEO consistently improves the baseline models' performance. In standard evaluation, using RTMDet~\cite{abs-2212-07784} as the front-end detector with SAMEO achieves the best performance on COCOA-cls, while using ConvNeXt-V2~\cite{WooDHC0KX23} as the front-end detector with SAMEO leads on D2SA. For zero-shot settings, using CO-DETR~\cite{ZongS023} as the front-end detector with SAMEO shows strong results on both COCOA-cls and D2SA, indicating SAMEO's effectiveness generalizes well across both class-specific and class-agnostic scenarios.

\section{Limitation and Future Work}
\label{sec:failure}

Although SAMEO notably outperforms SOTA methods in both scores and quality, it still faces challenges with difficult cases, as shown in \Cref{fig:failure}: incomplete amodal masks (a), rough edges (b), and unexpected modal outputs (top of (c)). When multiple objects overlap, using box prompts alone can cause model confusion. Additional experiments with using both box and point prompts show promising results in enhancing target region predictions (bottom of (c)). We believe exploring different prompt types is a direction for future work.
\input{sec/X_fig_future_work}
\begin{table*}
\centering

\begin{tabular}{l@{\hskip 0.25in}c@{\hskip 0.1in}c@{\hskip 0.1in}c@{\hskip 0.1in}c@{\hskip 0.25in}c@{\hskip 0.1in}c@{\hskip 0.1in}c@{\hskip 0.1in}c}
\toprule
\multirow{2}{*}{\textbf{Model}} & \multicolumn{4}{c@{\hskip 0.25in}}{\textbf{COCOA-cls}} & \multicolumn{4}{c@{\hskip 0.25in}}{\textbf{D2SA}} \\
& \textbf{AP} & \textbf{AP$_{50}$} & \textbf{AP$_{75}$} & \textbf{AR} & \textbf{AP} & \textbf{AP$_{50}$} & \textbf{AP$_{75}$} & \textbf{AR}\\
\midrule[\heavyrulewidth]
AISFormer~\cite{Tran0YFKL22} & 35.5 & 58.0 & 37.6 & 49.6 & 62.9 & 83.4 & 68.3 & 72.0\\
RTMDet$^{*}$~\cite{abs-2212-07784} & 50.4 & 68.1 & 55.4 & 69.9 &  53.9 & 71.9 & 57.3 & 75.8\\
ConvNeXt-V2$^{*}$~\cite{WooDHC0KX23} &  46.9 & 64.1 & 51.3 & 70.7 & 60.7 & 81.3 & 63.5 & 74.3  \\
\midrule
AISFormer$+$SAMEO & 46.4 & 62.1 & 50.5 & 62.8 & 72.2 & 84.3 & \textbf{76.6} & 79.2\\
RTMDet$^{*}$$+$SAMEO & \textbf{54.4} & \textbf{71.4} & \textbf{59.6} & \textbf{73.5} & 62.1 & 72.7 & 65.8 & 75.7\\
ConvNeXt-V2$^{*}$$+$SAMEO &  53.7 & 69.8 & 58.8 & 72.8  & \textbf{72.9} & \textbf{84.9} & 76.5 & \textbf{83.3} \\

\bottomrule
\end{tabular}

\caption{Class-specific performance on COCOA-cls and D2SA datasets. * denotes modal object detectors that provide modal bounding boxes as prompts. Bold numbers indicate the best performance.}
\label{tab:clsnonzeroshot}
\end{table*}

\begin{table*}
\centering

\begin{tabular}{l@{\hskip 0.25in}c@{\hskip 0.1in}c@{\hskip 0.1in}c@{\hskip 0.1in}c@{\hskip 0.25in}c@{\hskip 0.1in}c@{\hskip 0.1in}c@{\hskip 0.1in}c}
\toprule
\multirow{2}{*}{\textbf{Model}} & \multicolumn{4}{c@{\hskip 0.25in}}{\textbf{COCOA-cls}} & \multicolumn{4}{c@{\hskip 0.25in}}{\textbf{D2SA}} \\
& \textbf{AP} & \textbf{AP$_{50}$} & \textbf{AP$_{75}$} & \textbf{AR} & \textbf{AP} & \textbf{AP$_{50}$} & \textbf{AP$_{75}$} & \textbf{AR}\\
\midrule[\heavyrulewidth]
AISFormer & 35.5 & 58.0 & 37.6 & 49.6 & 62.9 & 83.4 & 68.3 & 72.0  \\
AISFormer$+$EfficientSAM$^\dagger$ & 42.0 & 59.2 & 45.3 & 59.3 &  62.7 & 80.5 & 64.9 & 72.4  \\
RTMDet$^{*}$$+$EfficientSAM$^\dagger$ & 48.7 & 67.5 & 53.3 & 65.8 &  55.9 & 72.4 & 57.4 & 77.3  \\
\midrule
AISFormer$+$SAMEO$^\dagger$ & 45.2 & 61.5 & 49.8 & 61.4 & 66.9 & 81.7 & 70.5 & 74.6\\
RTMDet$^{*}$$+$SAMEO$^\dagger$ & 53.3 & 70.6 & 59.2 & 72.5 & 60.2 & 74.1 & 62.6 & \textbf{81.0}\\
CO-DETR$^{*}$~\cite{ZongS023}$+$SAMEO$^\dagger$ & \textbf{53.6} & \textbf{70.6} & \textbf{59.5} & \textbf{73.3}  & \textbf{72.2} & \textbf{87.7} & \textbf{74.6} & 79.4 \\

\bottomrule
\end{tabular}

\caption{Zero-shot class-specific performance on COCOA-cls and D2SA datasets. † indicates zero-shot evaluation without training on the test dataset. * denotes modal object detectors that provide modal bounding boxes as prompts. Bold numbers indicate the best performance.}
\label{tab:clszeroshot}
\end{table*}

\end{document}